\documentclass{article}

\usepackage[backend=biber,style=numeric,sorting=none]{biblatex}
\usepackage{geometry,booktabs,xcolor,float}
\usepackage{hyperref,graphicx,tabularx, comments}
\usepackage[export]{adjustbox}
\usepackage{authblk}
\usepackage{latexsym,amsfonts,amssymb,amsthm,amsmath}
\usepackage[noabbrev]{cleveref}
\crefname{figure}{Fig.}{}
\Crefname{figure}{Figure}{}
\crefname{table}{Table}{}
\Crefname{table}{Table}{}
\crefname{section}{Sec.}{}
\Crefname{section}{Section}{}
\crefname{equation}{}{}
\Crefname{equation}{}{}
\usepackage[version=4]{mhchem} 
\usepackage{nomencl}
\makenomenclature
\usepackage[english]{babel}
\usepackage{csquotes}
\usepackage[ruled,vlined]{algorithm2e}

\usepackage[round-mode=places,round-precision=2,group-separator={,},output-decimal-marker={.}]{siunitx}

\usepackage{dcolumn,tikz}
\usetikzlibrary{shapes.geometric, arrows}
\usetikzlibrary{decorations.pathreplacing,calligraphy}
\tikzstyle{model} = [rectangle, rounded corners, 
  minimum width=3cm, 
  minimum height=1.5cm,
  text centered, 
  draw=black, 
  fill=red!30]

\tikzstyle{loss} = [rectangle, rounded corners,
  minimum width=3cm, 
  minimum height=1.5cm, text centered, 
  draw=black, fill=orange!30]

\tikzstyle{values} = [rectangle, rounded corners,
  minimum width=3cm, 
  minimum height=1.5cm, 
  text centered, 
  draw=black, 
  fill=pink!30]

\tikzstyle{simulation} = [rectangle, rounded corners,
  minimum width=3cm, 
  minimum height=1.5cm, 
  text centered, 
  text width=3cm, 
  draw=black, 
  fill=purple!30]
\tikzstyle{arrow} = [thick,->,>=stealth]
\usepackage{listofitems} 
\usetikzlibrary{arrows.meta} 
\usepackage[outline]{contour} 
\contourlength{1.4pt}

\tikzset{>=latex} 

\colorlet{myred}{red!30!black}
\colorlet{myblue}{blue!30!black}
\colorlet{mygreen}{green!30!black}
\colorlet{mydarkblue}{black!40!black}
\tikzstyle{node}=[thick,circle,draw=myblue,minimum size=22,inner sep=0.5,outer sep=0.6]
\tikzstyle{node in}=[node,green!20!black,draw=mygreen!30!black,fill=mygreen!25]
\tikzstyle{node hidden}=[node,blue!20!black,draw=myblue!30!black,fill=myblue!20]
\tikzstyle{node out}=[node,red!20!black,draw=myred!30!black,fill=myred!20]
\tikzstyle{connect}=[thick,mydarkblue] 
\tikzset{ 
  node 1/.style={node in},
  node 2/.style={node hidden},
  node 3/.style={node out},
}
\def\nstyle{int(\lay<\Nnodlen?min(2,\lay):3)} 

\makeatletter
\newcolumntype{B}[1]{>{\boldmath\DC@{.}{.}{#1}}c<{\DC@end}}
\makeatother

\newcommand\hdr[2]{\multicolumn{1}{#1}{#2}}
\newcolumntype{d}{D{.}{.}{-1}}
\newcolumntype{,}{D{,}{,}{-1}}
\newcommand\bd[1]{\multicolumn{1}{B{-1}}{#1}}

\newcommand{\Rey}{\mathrm{Re}}
\newcommand{\Ma}{\mathrm{Ma}}

\title{Physics-based parameterized neural ordinary differential equations: prediction of laser ignition in a rocket combustor}
\author[1]{Yizhou Qian}
\author[2]{Jonathan Wang}
\author[3]{Quentin Douasbin}
\author[4]{Eric Darve}
\affil[1]{Institute for Computational and Mathematical Engineering, Stanford University, USA}
\affil[2]{Center for Turbulence Research, Stanford University, USA}
\affil[3]{CERFACS, Toulouse, France}
\affil[4]{Department of Mechanical Engineering, Stanford University, USA}
\date{}

\begin{document}

\maketitle

\nomenclature{\(m\)}{total mass of all species in the reactor}
\nomenclature{\(\dot{m}_{\text{in}}\)}{mass inflow of an inlet}
\nomenclature{\(\dot{m}_{\text{out}}\)}{mass outflow of an outlet}
\nomenclature{\(\dot{m}_{k,\text{gen}}\)}{the rate of creation of species $k$}
\nomenclature{\(W_k\)}{molecular weight of species $k$}
\nomenclature{\(\dot{\omega}_{k}\)}{net reaction rate of species $k$}
\nomenclature{\(Y_k\)}{mass fraction of species $k$}
\nomenclature{\(X_k\)}{molar concentration of species $k$}
\nomenclature{\(Y_{k,\text{in}}\)}{mass fraction of species $k$ in mass inflow}
\nomenclature{\(Y_{k,\text{out}}\)}{mass fraction of species $k$ in mass outflow}
\nomenclature{\(\mathcal{M}_k\)}{symbol for species $k$}
\nomenclature{\(\nu'_{kj}\)/\(\nu''_{kj}\)}{molar stoichiometric coefficients of species $k$ in reaction $j$}
\nomenclature{\(\mathcal{Q}_j\)}{rate of progress of reaction $j$}
\nomenclature{\(K_{fj}\)}{forward rates of reaction $j$}
\nomenclature{\(K_{rj}\)}{reverse rates of reaction $j$}
\nomenclature{\(A_{fj}\)}{pre-exponential factor of reaction $j$}
\nomenclature{\(\beta_j\)}{temperature exponent of reaction $j$}
\nomenclature{\(\Delta S^0_j\)}{entropy changes of reaction $j$}
\nomenclature{\(\Delta H^0_j\)}{enthalpy changes of reaction $j$}
\nomenclature{\(E_j\)}{activation energy of reaction $j$}
\nomenclature{\(c_v\)}{specific heat at constant volume}
\nomenclature{\(\dot{Q}\)}{heat source}
\nomenclature{\(h\)}{specific total enthalpy}
\nomenclature{\(h_{\text{in}}\)}{specific total enthalpy of the mass inflow}
\nomenclature{\(U\)}{total internal energy}
\nomenclature{\(u_k\)}{internal energy of species $k$}
\nomenclature{\(T\)}{temperature}
\nomenclature{\(P\)}{pressure}
\nomenclature{\(p_a\)}{atmospheric pressure}

\printnomenclature

\section{Abstract}

This paper introduces a novel framework for reduced-order modeling of laser ignition in a model rocket combustor. It utilizes a physics-based data-driven approach based on parameterized neural ordinary differential equations (PNODE). Deep neural networks are embedded as functions of high-dimensional parameters of laser ignition to predict various terms in a 0D flow model including the heat source function, pre-exponential factors, and activation energy. By using the governing equations of a 0D flow model, physics-based PNODE requires only a limited number of training samples to predict trajectories of various quantities such as temperature, pressure, and mass fractions of species, while also satisfying physical constraints. We validate our physics-based PNODE on solution snapshots of high-fidelity Computational Fluid Dynamics (CFD) simulations of laser-induced ignition in a prototype rocket combustor. We compare the performance of our physics-based PNODE with that of kernel ridge regression and fully connected neural networks. Our results show that our physics-based PNODE provides solutions with lower mean relative errors of average temperature over time, thus improving the prediction of successful laser ignition with high-dimensional parameters. 

\section{Introduction}
\subsection{Background and related work} 

Rocket engine system design and optimization necessitate accurate predictions of their behavior, which is often characterized by complex, non-linear interactions among various components including heat transfer, combustion, flow resistance, and pump operations. Hence, repeated experiments or simulations are usually required for parametric studies of engine systems. Due to the complex combustion chemistry and wide range of spatiotemporal scales, Computational Fluid Dynamics (CFD) simulations are computationally expensive and therefore prohibitive for such systems. One possible approach is the use of flow networks, which consist of groups of nodes and flow branches that simulate interactions among components such as pumps, valves, nozzles, and orifices \cite{binder1995transient, majumdar2015generalized}. 

Many studies have employed flow network analysis for efficient rocket engine modeling and software tool development. \citeauthor{binder1993rl10a} utilized flow networks to model the RL10A rocket engine, the main propulsion system for the Centaur upper-stage vehicle, for both steady-state and transient analyses \cite{binder1993rl10a}. \citeauthor{di2011steady} developed a steady-state library based on flow networks for the iterative design of liquid rocket engines \cite{di2011steady}. \citeauthor{yamanishi2004transient} simulated the transient behavior of the LE-7A rocket engine by dividing the system into volumes and junctions, where scalar quantities, such as temperature and pressure, are considered at the center of the volume \cite{yamanishi2004transient}. However, only low-order models, such as zero or one-dimensional models, are used to represent various components including cooling jackets, turbines, and combustion chambers in flow network models. This limits their ability to simulate complex combustion dynamics inside rocket engine systems accurately.

Another promising alternative is data-driven surrogate models, which involve constructing fast solvers based on simulation data or experiments. Machine learning-based models are particularly attractive due to their ability to learn highly non-linear and complex mappings from combustion dynamics \cite{zhou2022machine, ihme2022combustion}. Several studies have attempted to learn known chemical schemes in combustion engines by leveraging artificial neural networks (ANN) \cite{blasco1998modelling, sharma2020deep}. Scalar quantities in chemically reacting flows, such as species composition, temperature, and pressure, are fed into neural networks as input to predict either the chemical source term or physical states at the next time step. Temporal evolutions of the reacting flows are later obtained by numerical integration or repeated evaluation of ANNs. The performance of ANNs demonstrated encouraging savings in computational time and memory \cite{blasco1998modelling}. 

Having learned detailed chemical mechanisms, neural networks can then be coupled with DNS or CFD solvers to provide scalable and accurate surrogate models for multi-dimensional combustion processes. \citeauthor{wan2020chemistry} used deep neural networks, which take species mass fractions and temperature as input, to predict the chemical source term from DNS of a turbulent non-premixed syngas oxy-flame interacting with a cooled wall \cite{wan2020chemistry}. \citeauthor{emami2012laminar} replaced numerical integration with artificial neural networks in a laminar flamelet model to predict mean reactive scalars in a non-premixed methane/hydrogen/nitrogen flame \cite{emami2012laminar}. Their neural networks were shown to achieve computational speedups of at least one order of magnitude, while also providing accurate predictions of species concentration and temperature. 

Other studies have also attempted to train deep learning models to learn the evolution of high-dimensional field variables directly. \citeauthor{an2020deep} replaced a conventional numerical solver with convolutional neural networks (CNNs) to predict flow fields from hydrogen-fueled turbulent combustion simulations, achieving two orders of magnitude of acceleration \cite{an2020deep}. Furthermore, to facilitate the experimental design of rocket engines, various engine parameters can be used as inputs to neural networks to predict flow fields of crucial quantities. \citeauthor{zapata2022data} trained fully-connected neural networks and U-nets as surrogate models to predict the global quantities, average, and root mean square fields inside shear-coaxial large edge simulations inside an injector rocket combustor \cite{zapata2022data}. Three design parameters, namely chamber diameter, recess length, and oxidizer-fuel ratio, are varied and fed into neural networks. While there are large errors in subdomains with high gradients, the distribution of error fields from deep learning models shows good agreement in general with high-fidelity simulations.

Nonetheless, the lack of physical constraints in data-driven models means that their solutions will not typically satisfy governing equations of physical laws, and thus can result in non-physical solutions. Furthermore, in those previous studies, deep learning models are trained based on paired samples with information on the current and next states. Therefore, when coupled with numerical integration or other iterative schemes, these deep learning models often diverge from the true solutions due to the accumulation of errors in the early stages of the algorithm.

\subsection{Neural ordinary differential equations} 

Deep neural networks have been known for their capability to learn highly complex and non-linear functions. With sufficient training data and a suitable model structure, neural networks can approximate arbitrary functions with arbitrary precision \cite{hornik1989multilayer}. For dynamical systems, a novel class of deep learning models known as neural ordinary differential equations has become popular recently, where deep neural networks are used to predict the derivative of quantities of interest given the current state variables \cite{chen2018neural,dupont2019augmented, chalvidal2020neural}. More precisely, a neural ODE is a parameterized ODE system of the following form:
\begin{equation}
    \frac{du}{dt} = F(u(t), t;\theta),
    \label{eq:neuralode}
\end{equation}
 where $u$ is the state variable of the system, $t$ is time, $F$ is a deep neural network, and $\theta$ are model parameters such as weights and biases. Gradients of the loss function with respect to model parameters are computed through automatic differentiation or the adjoint sensitivity method, which allows the neural ODE to be trained with high computational efficiency. 

Instead of computing the loss function based on paired samples of consecutive states, neural ODEs calculate the loss function directly based on solutions from the ODE, thus avoiding accumulating errors when performing numerical integration separately. Previous studies in combustion systems have attempted to use neural ODEs to learn chemical reactions from homogeneous reactors \cite{owoyele2022chemnode, dikeman2022stiffness}. \citeauthor{owoyele2022chemnode} proposed a deep learning framework based on neural ODEs to learn the homogeneous auto-ignition of hydrogen-air mixture with varying initial temperature and species composition \cite{owoyele2022chemnode}. \citeauthor{dikeman2022stiffness} combined an auto-encoder neural network with a neural ODE to compute the solution in the latent space instead of the original stiff system of chemical kinetics \cite{dikeman2022stiffness}. While initial temperature and equivalence ratio are varied to generate simulation data for training, the neural ODEs used in previous studies only learned a single dynamical system (i.e., combustion processes with fixed boundary conditions such as heat exchange, mass inflow, and mass outflow of homogeneous reactors) due to its limited expressivity. To enhance model generalizability and learn a parameterized system of ODEs, \citeauthor{lee2021parameterized} changed the original NODE framework and introduced parameterized neural ordinary differential equations (PNODE) that take additional parameters as input to DNNs \cite{lee2021parameterized}. Parameterized Neural Ordinary Differential Equations now describe the following system:
\begin{equation}
    \frac{du}{dt} = F(u(t,\eta), t, \eta;\theta),
    \label{eq:pnode}
\end{equation}
where $\eta$ represents problem-dependent parameters for the dynamical systems. This allows us to learn a parameterized family of ODEs instead of a single ODE. Using a similar optimization procedure (\textit{i.e.}, automatic differentiation) and the same set of model weights $\theta$, PNODE is capable of predicting solutions of dynamical systems under different conditions depending on the input parameter $\eta$.

In previous data-driven approaches based on neural ODEs, the source function $F$ in \cref{eq:neuralode} is a deep neural network $F_{\theta}$ with weights $\theta$ that learns to predict the time derivative of the state variables from input and output training data. However, these purely data-driven neural ODEs require large amounts of combustion data to learn combustion processes and chemical reactions with a wide range of parameters. Obtaining such training data from expensive high-fidelity simulations or physical experiments can be challenging. As a result, previous applications of neural ODEs in combustion studies have been limited to simulations of simplified chemical kinetics in homogeneous or 0D reactors, with data sets varying only two or three parameters, such as mass inflow, initial temperature, and species composition. To overcome these limitations, new approaches are needed to improve the accuracy and scalability of neural ODEs for complex combustion systems.

\section{Our contribution}

This paper introduces a novel physics-based PNODE framework that improves existing methods by directly learning from high-fidelity CFD simulations. Our approach incorporates the parameterization of a neural ODE to account for different experimental conditions, including mass inflow, initial species composition, and the geometry of the combustor chamber. We represent physical knowledge in our 0D flow model by utilizing deep neural networks to predict the heat source term and chemical kinetics, rather than approximating the entire source term $F$ as in \cref{eq:pnode}. By leveraging our framework, we can significantly enhance the accuracy and performance of our combustion system simulations. Our physics-based PNODE framework allows us to build a reduced-order surrogate model for laser ignition in a rocket combustor that
\begin{itemize}
    \item provides solutions satisfying \emph{physics constraints},
    \item learns combustion with \emph{high-dimensional} input parameters,
    \item needs \emph{less training samples} than purely data-driven approaches,
    \item matches the accuracy of \emph{high-fidelity} simulations. 
\end{itemize}

\begin{figure}
\begin{center}
\begin{tikzpicture}[node distance=2cm]
\node (laser_parameter) [values, align=center] {Parameters of \\ Laser Ignition};
\node (experiments) [simulation, right of=laser_parameter, xshift = 2.5cm, align=center] {Experiments or High-\\fidelity Simulations};
\node (pnode) [simulation, below of=experiments] {Physics-based\\ PNODEs};
\node (output) [values, right of=experiments, xshift = 2.5cm, align=center] {Average Temperature,\\ Pressure, and Mass\\ Fractions of Species};
\draw [arrow] (laser_parameter) -- (experiments);
\draw [arrow] (laser_parameter) |- (pnode);
\draw [arrow] (pnode) -| node[anchor = south east, align = center, xshift = -0.4cm]{Lower Cost} (output);
\draw [arrow] (experiments) -- (output);
\draw [arrow] (output) -- (9.0, 2.0) -- node[anchor=south]{Improved Design} (0.0, 2.0) -- (laser_parameter);
\end{tikzpicture}
\end{center}
\caption{Motivation for constructing a physics-based PNODE model. Time traces of volume-averaged temperature, pressure, and mass fractions of species provide crucial information that indicates ignition success or failure. Instead of computing volume-averaged quantities from high-fidelity simulations or physical experiments, we construct a PNODE-based 0D flow model that provides volume-averaged time traces directly with a lower computational cost.}
\label{fig:motivation}
\end{figure}
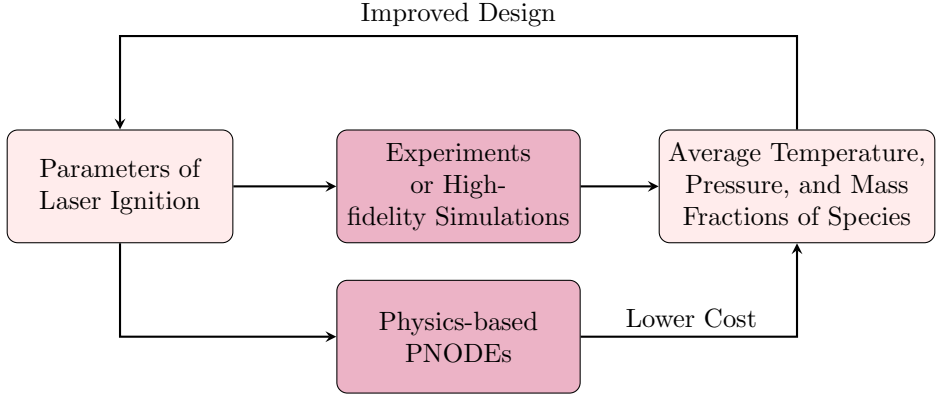

The general workflow is illustrated in \cref{fig:motivation}. We validate our approach on high-fidelity simulation data describing a laser-induced ignition of a prototype rocket combustor. We compare the performance of our PNODE-based 0D model with conventional interpolation methods such as kernel ridge regression and neural networks. 

The remainder of this paper is organized as follows. \Cref{sec:method} describes our overall methodology for physics-based PNODEs. \Cref{sec:0dmodel} describes the overall framework of our PNODE-based 0D model. \Cref{sec:neuralnetworks} reviews the structure of deep neural networks, the choice of hyper-parameters, and optimization pipelines. \Cref{sec:experiments} shows the numerical benchmarks of the PNODE-based models on high-fidelity simulations of laser-induced ignition of a rocket combustor.

\section{Methodology} \label{sec:method}
\subsection{0D flow model} \label{sec:0dmodel}

Given a parameter $\eta$, we wish to use a 0D model to describe the ignition process of the target combustion system by predicting the evolution of volume-averaged quantities such as the temperature, pressure, and mass fractions of species. We formulate our 0D flow model based on the Continuously Stirred Tanked Reactor (CSTR) model with fixed volume from the Cantera library \cite{cantera}. CSTR is a simplified constant volume reactor that accounts for several inlets, several outlets, and the reacting flow in the chamber is considered spatially homogeneous due to the high mixing rate provided by the continuous stirring.

\begin{figure}
\centering
\tikzset{every picture/.style={line width=0.75pt}} 

\begin{tikzpicture}[x=0.75pt,y=0.75pt,yscale=-1,xscale=1]

\draw    (229,35.67) -- (229,80) ;
\draw    (170.48,80) -- (207.5,80) -- (229,80) ;
\draw    (229,188.95) -- (229,237.86) ;
\draw    (229,99) -- (169.75,99) ;
\draw    (229,36) -- (460,36) ;
\draw    (460,36.09) -- (460,189) ;
\draw    (460,237) -- (229,237) ;
\draw    (460,210) -- (460,237) ;
\draw    (460,189) -- (519.43,189) ;
\draw    (460,210.13) -- (519.43,210) ;
\draw    (229,189) -- (229,99) ;
\draw    (290,20.25) -- (400.5,20.25) ;
\draw    (345.25,20.25) -- (345.25,170) ;
\draw    (320.5,170) -- (370.5,170) ;
\draw  [fill={rgb, 255:red, 255; green, 255; blue, 255 }  ,fill opacity=1 ] (288,80.5) .. controls (288,66.69) and (310.39,55.5) .. (338,55.5) -- (338,70.5) .. controls (310.39,70.5) and (288,81.69) .. (288,95.5) ;\draw  [fill={rgb, 255:red, 255; green, 255; blue, 255 }  ,fill opacity=1 ] (288,95.5) .. controls (288,105.75) and (300.34,114.56) .. (318,118.42) -- (318,123.42) -- (338,113) -- (318,98.42) -- (318,103.42) .. controls (300.34,99.56) and (288,90.75) .. (288,80.5)(288,95.5) -- (288,80.5) ;
\draw  [fill={rgb, 255:red, 255; green, 255; blue, 255 }  ,fill opacity=1 ] (402.85,95.07) .. controls (403.36,108.87) and (381.4,120.88) .. (353.81,121.9) -- (353.25,106.91) .. controls (380.85,105.89) and (402.81,93.88) .. (402.3,80.08) ;\draw  [fill={rgb, 255:red, 255; green, 255; blue, 255 }  ,fill opacity=1 ] (402.3,80.08) .. controls (401.92,69.84) and (389.26,61.49) .. (371.48,58.28) -- (371.29,53.29) -- (351.69,64.44) -- (372.21,78.27) -- (372.03,73.27) .. controls (389.82,76.48) and (402.47,84.83) .. (402.85,95.07)(402.3,80.08) -- (402.85,95.07) ;
\draw   (250.5,170.86) .. controls (250.44,163.61) and (266.06,157.6) .. (285.39,157.44) .. controls (304.72,157.27) and (320.44,163) .. (320.5,170.25) .. controls (320.56,177.5) and (304.95,183.51) .. (285.62,183.68) .. controls (266.29,183.84) and (250.57,178.11) .. (250.5,170.86) -- cycle ;
\draw   (370.5,170.25) .. controls (370.5,163) and (386.17,157.13) .. (405.5,157.13) .. controls (424.83,157.13) and (440.5,163) .. (440.5,170.25) .. controls (440.5,177.5) and (424.83,183.37) .. (405.5,183.37) .. controls (386.17,183.37) and (370.5,177.5) .. (370.5,170.25) -- cycle ;
\draw[arrow]    (69.43,88) -- node[anchor=south]{$\dot{m}_{\text{in}}, Y_{k,\text{in}}, T_{\text{in}}$} (156,88) ;
\draw[arrow]   (534.43,200) -- node[anchor=south]{$\dot{m}_{\text{out}},Y_{k, \text{out}}, T_{\text{out}}$} (627,200);
\draw (290,217.31) node [align=left] {\begin{minipage}[lt]{68pt}\setlength\topsep{0pt}
$P,V,T,Y_k$
\end{minipage}};
\end{tikzpicture}
\caption{Illustration of a Continuously Stirred Tanked Reactor (CSTR) with one inlet and one outlet. $P$ is volume-averaged pressure, $V$ is volume, $T$ is volume-averaged temperature, and $Y_k = \frac{m_k}{m}$ is the mass fraction of species $k$ in the reactor. Gases inside the reactor are perfectly mixed. This is a simplified model for high-fidelity simulations, where fuels are injected into the chamber through the inlet at rate $\dot{m}_{\text{in}}$ and gases inside the reactor leave through the outlet at rate $\dot{m}_{\text{out}}$. $Y_{k, \text{in}} = \frac{\dot{m}_{k, \text{in}}}{\dot{m}_{\text{in}}}$ is the mass fraction of species $k$ of the mass inflow. $T_{\text{in}}$ is the temperature of the mass inflow. $Y_{k, \text{out}} = \frac{\dot{m}_{\text{k,out}}}{\dot{m}_{\text{out}}}$ is the mass fraction of species $k$ of the mass outflow. $T_{\text{out}}$ is the temperature of the mass outflow. }
\end{figure}
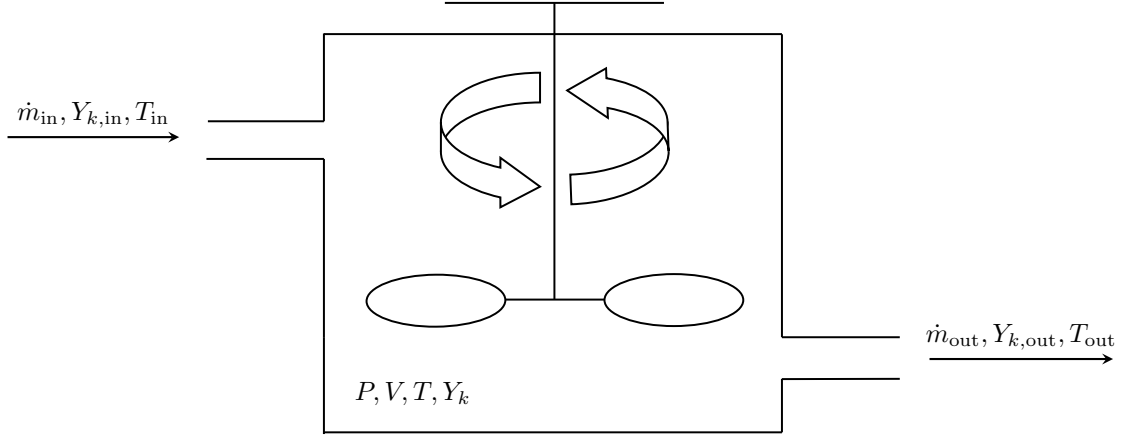

\subsubsection{Mass conservation}

In a CSTR reactor, the total mass of the system is conserved. Therefore, we have
\begin{equation}
    \frac{dm}{dt} = \sum_{\text{in}} \dot{m}_{\text{in}} - \sum_{\text{out}} \dot{m}_{\text{out}} = \sum_{k} \frac{dm_k}{dt} = \sum_{\text{in}} \sum_k \dot{m}_{k, \text{in}} - \sum_{\text{out}} \sum_k \dot{m}_{\text{k,out}},
    \label{eq:mass_conservation}
\end{equation}
where $m$ is the total mass of all species in the reactor, $m_k$ is the total mass of species $k$, $\dot{m}_{\text{in}}$ is the mass inflow of an inlet, $\dot{m}_{k, \text{in}}$ is the mass inflow of species $k$ of an inlet, $\dot{m}_{\text{out}}$ is the mass outflow of an outlet, and $\dot{m}_{k, \text{out}}$ is the mass outflow of species $k$ of an outlet.

\subsubsection{Species conservation}

In reacting flows, species are created and destroyed in time from chemical reactions. The rate of species creation $k$ is given as:
\begin{equation}
    \dot{m}_{k,\text{gen}} = V\dot{\omega}_{k}W_{k},
    \label{eq:creationrate}
\end{equation}
where $W_k$ is the molecular weight of species $k$ and $\dot{\omega}_k$ is the net reaction rate. The rate of change of a species' mass is given by
\begin{equation}
    \frac{d(mY_k)}{dt} = \sum_{\text{in}} \dot{m}_{\text{in}}Y_{k, \text{in}} - \sum_\text{out} \dot{m}_\text{out} Y_{k,\text{out}} + \dot{m}_{k,\text{gen}},
    \label{eq:speciesmass}
\end{equation}
where $Y_k = \frac{m_k}{m}$ is the mass fraction of species $k$ in the reactor, $Y_{k,\text{in}} = \frac{\dot{m}_{k, \text{in}}}{\dot{m}_{\text{in}}}$ is the mass fraction of species $k$ in the inflow, $Y_{k,\text{out}} = \frac{\dot{m}_{\text{k,out}}}{\dot{m}_{\text{out}}}$ is the mass fraction of species $k$ in the outflow. Combining \cref{eq:creationrate,eq:speciesmass}, we obtain
\begin{equation}
    m\frac{dY_k}{dt} = \sum_{\text{in}} \dot{m}_{\text{in}}(Y_{k,\text{in}}-Y_k) - \sum_{\text{out}} \dot{m}_{\text{out}}(Y_k - Y_{k,\text{out}}) +  V \dot{\omega}_{k} W_k.
    \label{eq:species}
\end{equation}

\subsubsection{Chemical reactions}

Suppose that for reaction $j$ we have 
\begin{equation}
    \sum_{k=1}^N \nu'_{kj} \mathcal{M}_k \leftrightharpoons \sum_{k=1}^N \nu''_{kj}\mathcal{M}_k,
    \label{eq:reactions}
\end{equation}
where $\mathcal{M}_k$ is a symbol for species $k$, $\nu'_{kj}$ and $\nu''_{kj}$ are the stoichiometric coefficients of species $k$ in reaction $j$. Let $\nu_{kj} = \nu''_{kj} - \nu'_{kj}$. Then the source term $\dot{\omega}_{k}$ can be written as the sum of the source terms $\dot{\omega}_{k,j}$ from reaction $j$:
\begin{equation}
    \dot{\omega}_k = \sum_j \dot{\omega}_{k, j}
    = W_k \sum_{j=1}^M \nu_{kj}\mathcal{Q}_j,
    \label{eq:omegasum}
\end{equation}
where
\begin{equation}
    \mathcal{Q}_j = K_{fj}\prod_{k=1}^N [X_k]^{\nu'_{kj}}-K_{rj}\prod_{k=1}^{N}[X_k]^{\nu''_{kj}},
    \label{eq:qj}
\end{equation}
and
\begin{equation}
    K_{fj} = A_{fj}T^{\beta_j}\exp{\left(-\frac{E_j}{RT}\right)}, K_{rj} = \frac{K_{fj}}{\left(\frac{p_a}{RT}\right)^{\sum_{k=1}^N\nu_{kj}}\exp{\left( \frac{\Delta S^0_j}{R} - \frac{\Delta H^0_j}{RT} \right)}},
    \label{eq:forwardconstant}
\end{equation}
where $p_a$ is atmospheric pressure, $\mathcal{Q}_j$ is the rate of progress of reaction $j$, $X_k = \frac{W}{W_k}Y_k$ is the molar concentration of species $k$, $K_{fj}$ and $K_{rj}$ are the forward and reverse rates of reaction $j$, $A_{fj}$, $\beta_j$ and $E_j$ are the pre-exponential factor, temperature exponent and activation energy for reaction $j$, respectively. $\Delta S^0_j$ and $\Delta H^0_j$ are entropy and enthalpy changes, respectively, for reaction $j$ \cite{poinsot2005theoretical}. 

\subsubsection{Energy conservation}

For a CSTR reactor with fixed volume, the internal energy can be expressed by writing the first law of thermodynamics for an open system \cite{kee1989chemkin}:
\begin{equation}
    \frac{dU}{dt} = - \dot{Q} + \sum_{\text{in}} \dot{m}_{\text{in}}h_{\text{in}} - \sum_{\text{out}} h\dot{m}_{\text{out}},
    \label{eq:thermo}
\end{equation}
where $\dot{Q}$ is the heat source, $h$ is the specific enthalpy of the homogeneous gas in the reactor, and $h_{\text{in}}$ is the specific enthalpy of the mass inflow. We can describe the evolution of the volume-averaged temperature by expressing the internal energy $U$ in terms of the species mass fractions $Y_k$ and temperature $T$:
\begin{equation}
    U = m\sum_{k}Y_ku_k(T).
    \label{eq:enthalpy}
\end{equation}
so that 
\begin{equation}
        \frac{dU}{dt} = u\frac{dm}{dt} + mc_v\frac{dT}{dt} + m \sum_k u_k \frac{dY_k}{dt}.
        \label{eq:dudt}
\end{equation}
where $c_v$ is the specific heat at constant volume. From \cref{eq:dudt,eq:thermo} we have
\begin{align*}
    mc_v \frac{dT}{dt} & = \frac{dU}{dt} - u \frac{dm}{dt} - m\sum_{k}u_k \frac{dY_k}{dt}, \\
    & = -\dot{Q} + \sum_{\text{in}} \dot{m}_{\text{in}}h_{\text{in}} - \sum_{\text{out}} h\dot{m}_{\text{out}} - u \frac{dm}{dt} - m\sum_{k}u_k \frac{dY_k}{dt}.
\end{align*}
Next, using \cref{eq:mass_conservation,eq:species} we have 

\begin{equation}
\begin{aligned}
    mc_v \frac{dT}{dt} & = -\dot{Q} + \sum_{\text{in}} \dot{m}_{\text{in}}\left(h_{\text{in}}  
     - \sum_{k} u_k Y_{k,\text{in}}\right) \\ 
     & - \sum_{\text{out}} \dot{m}_{\text{out}}\left(h - \sum_{k} u_k Y_k - (Y_k -Y_{k,\text{out}})\right) - \sum_k V\dot{\omega}_{k}W_ku_k
    \label{eq:energy}
\end{aligned}
\end{equation}

\subsection{Physics-based parameterized neural ordinary differential equations} \label{sec:neuralnetworks}

In this work, we consider the use of neural ODE for combustion studies from a different perspective. Instead of learning the source function directly with deep neural networks in NODE, we consider using existing 0D models as a starting point to build the differential equation. More precisely, to incorporate physical knowledge into our PNODE, we construct $F$ in \cref{eq:pnode} based on a 0D flow model $F_{0D}$ and embed deep neural networks into our 0D flow model to model terms such as the heat source and activation energy. Let $\dot{Q}(t, \eta; \theta)$ be the heat source function, $A_f(t,\eta; \theta)$ be the pre-exponential factor, and $E(\eta; \theta)$ be the activation energy function. These are represented by deep neural networks and are used as inputs to the 0D flow model $F$. Then we have
\[\frac{du}{dt} = F\left(u(t,\eta), t, \eta;\theta\right) = F_{0D}\left(u(t, \eta), \dot{Q}(t, \eta; \theta), A_f(t, \eta; \theta), E(\eta; \theta)\right)\]
The prediction $\hat{u}(t_i)$ can be obtained by solving the ODE system using a numerical solver:
\[\hat{u} = (\hat{u}(t_0), \dots, \hat{u}(t_n)) = \textrm{ODESolve}(u(t_0), F_{0D}, t_1,...,t_n)\]
To optimize the parameters in the neural networks that model the heat source, pre-exponential factor, and activation energy, we minimize the root mean squared loss between observations and predictions. This is accomplished using \emph{reverse mode automatic differentiation} applied to the numerical ODE solver. The exact formulation of each component in $F_{0D}$ is described in \cref{sec:method}. Our work represents a novel application of 0D models in conjunction with neural ODEs for combustion systems. Unlike previous approaches that rely solely on neural ODEs, our physics-based PNODE benefits from additional information provided by the 0D model. This enables PNODE to learn parameters from a higher dimensional space directly from high-fidelity simulations, distinguishing it from previous neural ODE approaches. To the best of our knowledge, our study is the first to employ this innovative technique.

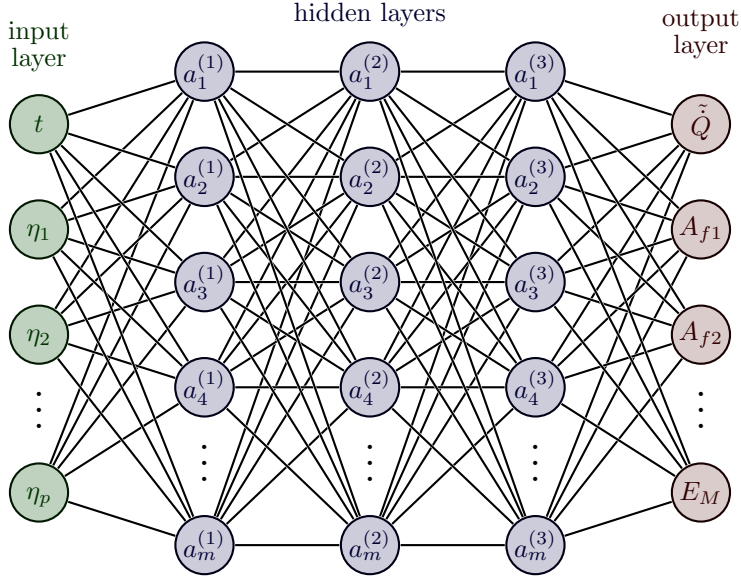
\begin{figure}
\begin{center}
\begin{tikzpicture}[x=2.2cm,y=1.4cm]

  \readlist\Nnod{4,5,5,5,4} 
  \readlist\Nstr{p,m,m,m,j} 
  \readlist\Cstr{\strut \eta,a^{(\prev)},a^{(\prev)},a^{(\prev)},Y} 
  \readlist\Zstr{\strut \Tilde{\dot{Q}}, A_{f1}}
  \def\yshift{0.5} 

  \foreachitem \N \in \Nnod{ 
    \def\lay{\Ncnt} 
    \pgfmathsetmacro\prev{int(\Ncnt-1)} 
    \foreach \i [evaluate={\c=int(\i==\N); \y=\N/2-\i-\c*\yshift;
                 \index=(\i<\N?int(\i):"\Nstr[\lay]");
                 \indexfirst=(\i<\N?int(\i-1):"\Nstr[\lay]");
                 \x=\lay; \n=\nstyle; \s=int(\i-1); \z=int(\lay); \u=(\i<\N?int(\i-2):"\Nstr[\lay]")}] in {1,...,\N}{ 
      \ifnum\lay>1
        \ifnum\lay=5
            \ifnum\i<3
                \node[node \n] (N\lay-\i) at (\x,\y) {$\Zstr[\i]$}; 
            [\else
                \ifnum\i=3
                    \node[node \n] (N\lay-\i) at (\x,\y) {$A_{f2}$};
                [\else
                    \node[node \n] (N\lay-\i) at (\x,\y) {$E_{M}$};
                ]
                \fi
            ]
            \fi
        [\else
           \node[node \n] (N\lay-\i) at (\x,\y) {$\Cstr[\lay]_{\index}$}; 
        ]
        \fi
        [\else
            \ifnum\i=1
                \node[node \n] (N\lay-\i) at (\x,\y) {$t$};
            [\else
                \node[node \n] (N\lay-\i) at (\x,\y) {$\Cstr[\z]_{\indexfirst}$};
            ]
            \fi
        ]
      \fi
      
      \ifnum\lay>1 
        \foreach \j in {1,...,\Nnod[\prev]}{ 
          \draw[connect,white,line width=1.2] (N\prev-\j) -- (N\lay-\i);
          \draw[connect] (N\prev-\j) -- (N\lay-\i);
        }
      \fi 
      
    }
    \path (N\lay-\N) --++ (0,1+\yshift) node[midway,scale=1.5] {$\vdots$};
  }
  
  \node[above=5,align=center,mygreen!60!black] at (N1-1.90) {input\\[-0.2em]layer};
  \node[above=2,align=center,myblue!60!black] at (N3-1.90) {hidden layers};
  \node[above=10,align=center,myred!60!black] at (N\Nnodlen-1.90) {output\\[-0.2em]layer};
\end{tikzpicture}
\end{center}
\caption{Structure of a fully connected neural network $F_{\theta}$ with three hidden layers. Here $a_j^{(i)}$ refers to the $j$th neuron in the $i$th hidden layer of our deep neural network. In each forward pass, our deep neural network takes laser parameters $\eta_1,\dots,\eta_p$, time $t$ as input and predicts the temporal distribution of the heat source function $\Tilde{\dot{Q}}$, pre-exponential factors $A_{f1}, \dots, A_{fM}$, and activation energy $E_1,\dots, E_M$ at time $t$ as output.}
\label{fig:neuralnetwork}
\end{figure}

Using information from the parameters $\eta = (\eta_1, ..., \eta_p)$ of the laser ignition as input, we compute the heat source $\dot{Q}$ in \cref{eq:energy}, the pre-exponential factor $A_{fj}$, and the activation energy $E_j$ in \cref{eq:forwardconstant} with fully connected feed-forward neural networks $F_{\theta}(t, \eta)$ as shown in \cref{fig:neuralnetwork}, where $\theta$ are model parameters. We remark here that our DNN $F_{\theta}$ is not directly learning the heat source function $\dot{Q}$ in \cref{eq:energy}. To ensure the training stability of our physics-based PNODE, our $F_{\theta}$ outputs a different quantity $\tilde{\dot{Q}}$ that learns the temporal distribution of the heat injected into the chamber first. The heat source function $\dot{Q}$ is obtained by normalizing $\tilde{\dot{Q}}$ and multiplying it by another neural network $C_{\xi}$ that learns the total amount of heat injected into the system separately. That is, we define 
\begin{equation}
  \dot{Q}(t, \eta) = C_{\xi}(\eta) 
  \frac{\tilde{\dot{Q}}(t,\eta)}{\int_{t_\text{start}}^{t_\text{end}} 
  \tilde{\dot{Q}}(t,\eta) dt},
  \label{eq:normalization}
\end{equation}
where $C_{\xi}(\eta)$ is another deep neural network with model parameter $\xi$. 

We recall the advantage of this parameterization of the 0D flow model using neural networks. While only scalar parameters are calibrated in conventional 0D models, our physics-based PNODE framework uses deep neural networks that are optimized automatically based on arbitrary loss functions. Additionally, deep neural networks have been shown as a powerful tool for learning complex and nonlinear chemical reactions in combustion studies\cite{christo1996integrated, christo1996artificial}. The temperature $T(i,j)$ and the mass fraction of oxygen $Y_{\ce{O2}}(i,j)$ over time are used to compute the mean squared error as our loss function. That is, our loss function is defined as
\begin{equation}
    L = \sum_{i = 1}^{n} \sum_{j = 1}^{m} (T(i, j) - T_\text{true}(i,j))^2 + \alpha (Y_{\ce{O2}}(i,j) - Y_{\ce{O2}, \text{true}}(i, j))^2,
    \label{eq:loss}
\end{equation}
where $\alpha$ is a hyper-parameter that determines the weighted mean squared errors of the temperature and mass fraction of oxygen, $T(i,j)$ and $Y_{\ce{O2}}(i,j)$ are the predictions of the temperature and mass fraction of oxygen, respectively, at observation time $t_j$ for sample $i$. Note that since only a 1 step chemical mechanism (which is described in \cref{sec:setup}) is used in our 0D flow model and high-fidelity simulations, it is sufficient to minimize the mean squared error for the mass fraction of oxygen in the loss function. 

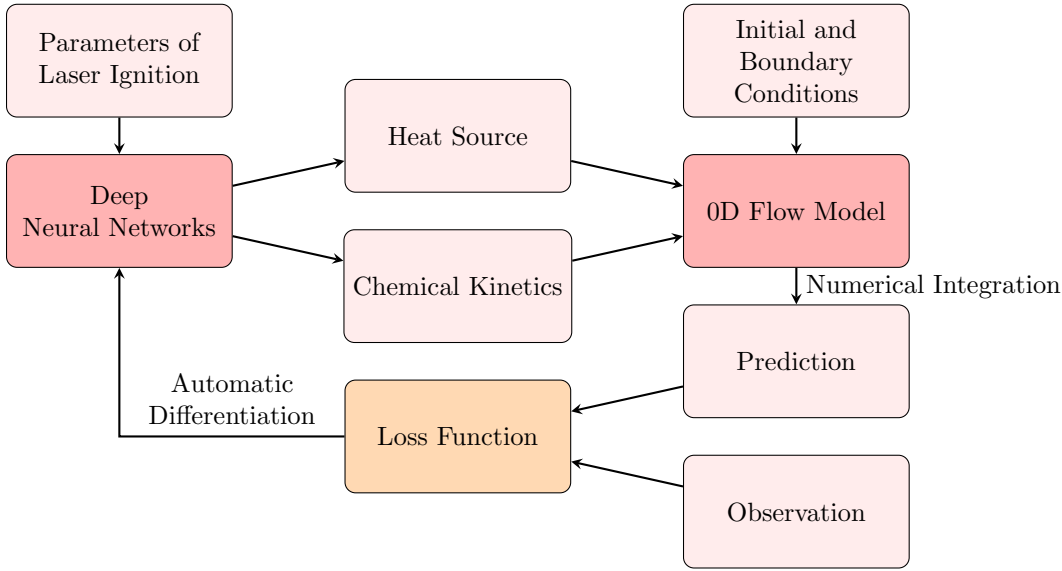
\begin{figure}[htbp]
\begin{center}
\begin{tikzpicture}[node distance=2cm]

\node (laser_parameter) [values, align=center] {Parameters of \\ Laser Ignition};
\node (in1) [model, below of=laser_parameter, align=center] {Deep\\ Neural Networks};
\node (pnode) [values, right of=in1, xshift = 2.5cm, yshift = -1cm] {Chemical Kinetics};
\node (experiments) [values, right of=laser_parameter, xshift = 2.5cm, yshift = -1cm, align=center] {Heat Source};
\node (output) [model, right of=experiments, xshift = 2.5cm, yshift = -1cm, align=center] {0D Flow Model};
\node (condition) [values, above of=output, align=center] {Initial and\\ Boundary\\ Conditions} ;
\node (prediction) [values, below of=output] {Prediction} ;
\node (observation) [values, below of=prediction] {Observation} ;
\node (loss_function) [loss, left of=observation, xshift = -2.5cm,  yshift = 1cm] {Loss Function} ;
\draw [arrow] (in1) -- (experiments);
\draw [arrow] (laser_parameter) -- (in1);
\draw [arrow] (in1) -- (pnode);
\draw [arrow] (pnode) -- (output);
\draw [arrow] (experiments) -- (output);
\draw [arrow] (condition) -- (output);
\draw [arrow] (output) -- node[anchor = west]{Numerical Integration} (prediction);
\draw [arrow] (observation) -- (loss_function);
\draw [arrow] (prediction) -- (loss_function);
\draw [arrow] (loss_function) -| node[xshift = 1.5cm, yshift = 0.5cm, align = center] {Automatic \\ Differentiation}(in1);
\end{tikzpicture}
\end{center}
\caption{Optimization pipeline of the physics-based PNODE framework. Parameters of laser ignition are passed into deep neural networks to predict the heat source and chemical kinetics for our 0D flow model. Predictions of volume-average temperature, pressure, and mass fractions of species over time are obtained through numerical integration. We update the weights in our deep neural networks by leveraging ADCME library.}
\label{fig:optimization}
\end{figure}

The gradient of the loss function $L$ with respect to the weights $\theta$ are computed via automatic differentiation using Automatic Differentiation Library for Computational and Mathematical Engineering (ADCME) \cite{xu2020adcme}, which is designed specifically for high-performance computing and inverse modeling in scientific computing. In our physics-based PNODE, we use a Runge-Kutta fourth-order method to perform numerical integration. We use L-BFGS-B as the optimizer for the weights of the neural networks in ADCME. The overall optimization pipeline is illustrated in \cref{fig:optimization}. 

\section{Numerical experiments} \label{sec:experiments}

We validate our approach on data generated from high-fidelity simulations of a planar jet diffusion based on the Hypersonics Task-Based Research (HTR) solver developed by \cite{di2020htr}. In the rocket combustor, a gaseous \ce{O2} jet, along with \ce{CH4} coflow, is injected into the chamber filled with gaseous \ce{CH4}. The jet is ignited by intense, short-time heating at a specific location. \Cref{fig:combustor} illustrates the setup of the rocket combustor in our 2D high-fidelity simulations. In this work, we consider six parameters in total for laser-induced ignition, which are described in \cref{tab:parameters}. We use a 1 step chemical mechanism, which has 5 species and 1 global reaction, as the chemical scheme in our 0D flow model \cite{franzelli2012large}. To model ignition in high-fidelity simulations, we adjusted our initial values of the Arrhenius reaction parameters before training for our 0D flow model. In particular, for initial values before optimization, we set our pre-exponential factors to be $A_f = 121.75$ and set the activation energies equal to $E_{a} = 1.68 \times 10^7$. Note that to fit simulation data from high-fidelity 2D simulations, we are setting both the pre-exponential factor and the activation energy to be smaller than default values. This is because the volume-averaged temperature during ignition is significantly lower than that of default 0D reactions, and also it increases at a much slower rate after successful ignition. We also choose $\alpha = 1 \times 10^7$ in \cref{eq:loss}.

\subsection{Simulation setup} \label{sec:setup}

\begin{figure}
\begin{tikzpicture}
\centering
\node (at) at (-3, 3) {\includegraphics[width=4cm, height=2.8cm]{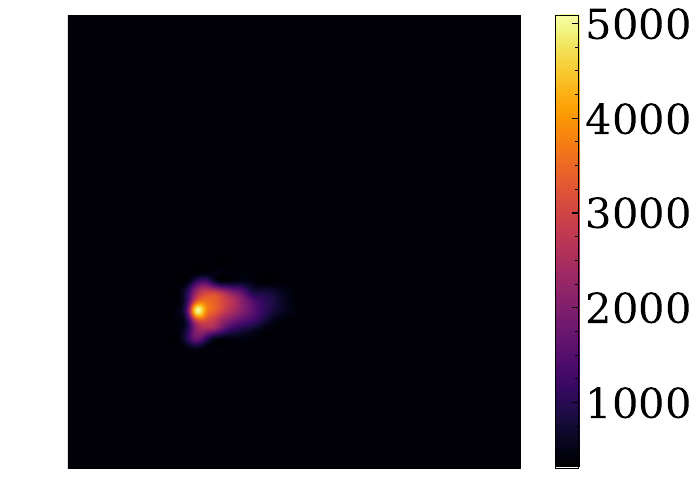}};
\node (et) at (5, 3) {\includegraphics[width=8cm, height=4cm]{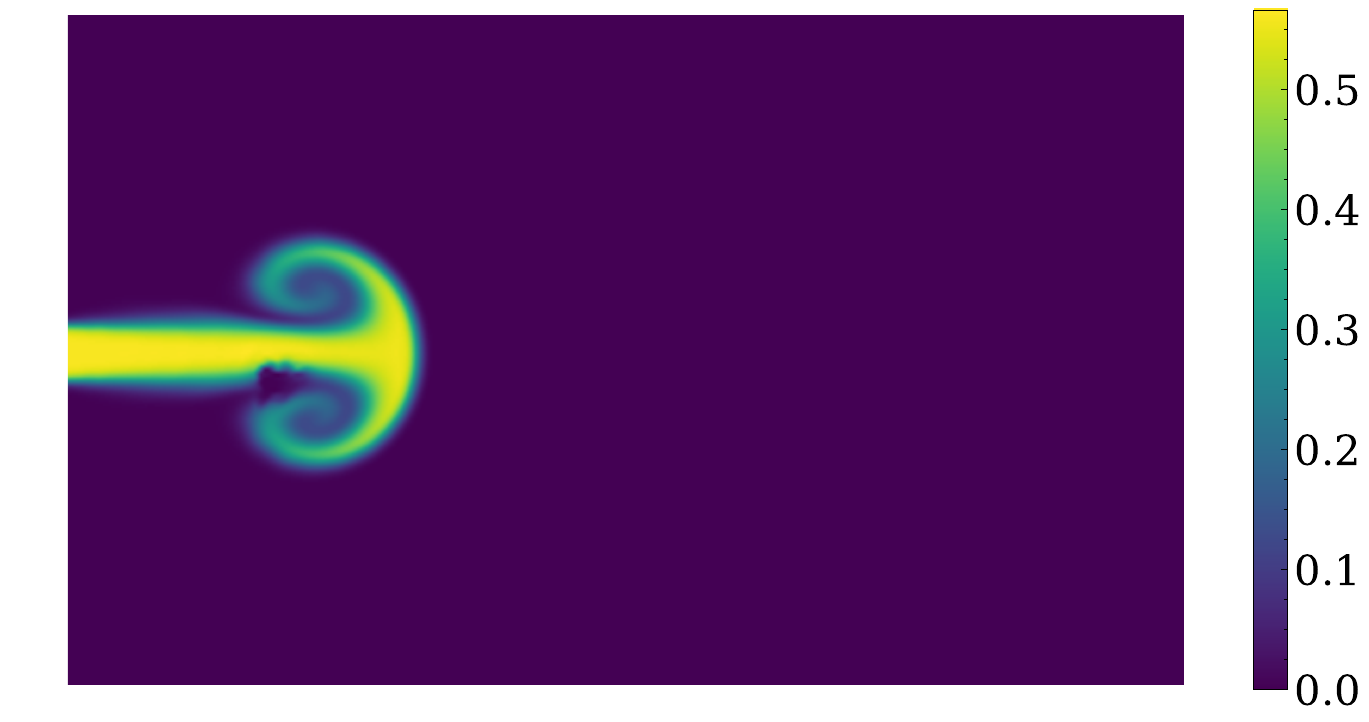}};
\node[draw=red, dashed, minimum width=1.1cm, minimum height=1.2cm, text centered] (bt) at (3,3) {};
\node[yshift = -0.2cm, xshift = -0.5cm] (ct) at (at.north) {};
\node[yshift = 0.3cm, xshift = -0.4cm] (dt) at (at.south) {};
\draw[arrow, dashed, red] (bt.north) -- (ct);
\draw[arrow, dashed, red] (bt.south) -- (dt);
\draw[>=latex, <->, red] (1.5,3.25) -- node[anchor=west]{$D$} (1.5,2.85);
\draw [decorate, decoration = {calligraphic brace}] (7.96,1) -- node[anchor=north]{60.0 D} (1.5,1);
\draw [decorate, decoration = {calligraphic brace}] (9,4.75) -- node[anchor=south, rotate = 270]{20.0 D} (9,1.25);
\draw[arrow] (9.75, 4) -- node[anchor=south] {Outflow} (11, 4);
\draw[arrow] (9.75, 2) -- node[anchor=south] {Outflow} (11, 2);
\draw[arrow] (-0.75, 4) -- node[anchor=south] {\ce{CH4} co-flow} (1.25, 4);
\draw[arrow] (-0.75, 3) -- node[anchor=south] {\ce{O2} jet} (1.25, 3);
\draw[arrow] (-0.75, 2) -- node[anchor=south] {\ce{CH4} co-flow} (1.25, 2);
\node[text = white] at (4.75,4.65) {Wall};
\node[text = white] at (4.75,1.5) {Wall};
\node at (4.75,5.1) {Mass Fraction of Oxygen};
\node at (-3.25,4.6) {Temperature (F)};
\node[draw = red, circle] (rc) at (-3.8, 2.6) {};
\draw (rc) node[anchor=south, text = white, above=5pt]{Ignition};
\draw[dashed, red] (rc) -- node[anchor = south, text = black, below = 9pt] {$x$}(-3.8, 1.7);
\draw[dashed, red] (rc) -- node[anchor = east, text = black, left = 8pt] {$y$} (-4.8, 2.6);
\end{tikzpicture}
\caption{Setup of a 2D laser-induced ignition in a rocket combustor. The snapshot was taken during laser deployment. Pure oxygen is injected at \SI{350}{K} into a two-dimensional chamber, along with a methane co-flow. Here $x$ and $y$ are the two-dimensional coordinates of laser-induced ignition inside the chamber. The length and width of the 2D chamber are shown using scalar $D$ as the unit, which refers to the jet-thickness of the injected fuel.}
\label{fig:combustor}
\end{figure}
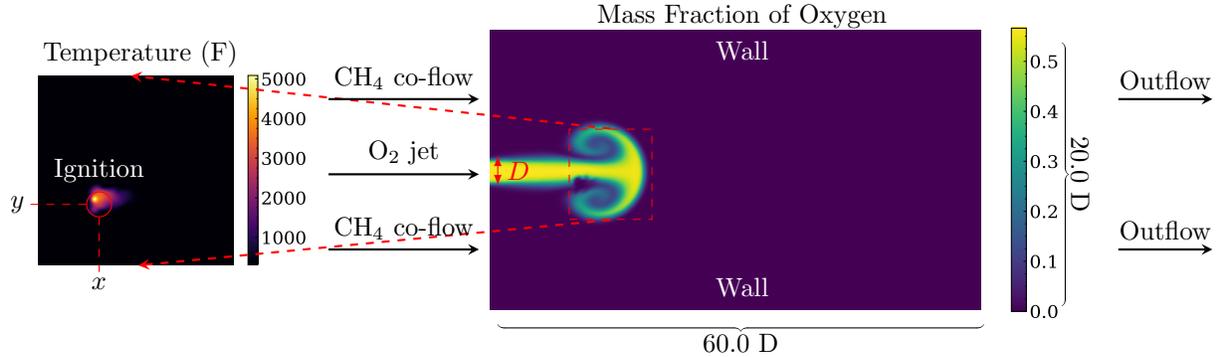

To validate our approach, we used a direct numerical simulation of a two-dimensional planar jet diffusion flame as a high-fidelity reference. Although simplified compared to full-scale rocket combustors, for the present objectives, this configuration bears a sufficient physical resemblance to such a system while requiring relatively low computational resources, and it is taken as a high-fidelity model compared to the 0D model (\cref{sec:0dmodel}). A schematic of the simulation domain is shown in \cref{fig:combustor}. The combustion chamber initially contains 100\% methane at \SI{350}{K} and \SI{50662.5}{Pa}, and pure oxygen is injected at \SI{350}{K} and with Reynolds number $\Rey\equiv Ud/\nu_{\ce{O2}}=400$ and Mach number $\Ma\equiv U_{\ce{O2}}/a_{\ce{O2}}=0.1$, where $U_{\ce{O2}}$ is the injection velocity, $d$ is the jet diameter, $\nu_{\ce{O2}}$ is the kinematic viscosity of the injected oxygen and $a_{\ce{O2}}$ is its speed of sound. A coflow of methane at a lower velocity $U_{\ce{CH4}}/U_{\ce{O2}}=0.001$ accompanies the oxygen jet.

The compressible Navier--Stokes equations for a multicomponent ideal gas are solved with four chemical species (\ce{CH4}, \ce{O2}, \ce{H2O}, \ce{CO2}) and a 1-step irreversible reaction for methane--oxygen combustion \cite{franzelli2012large}: 
\[
    \ce{CH4 + 2 O2 -> 2 H2O + CO2}.
\]
Characteristic boundary conditions are used at inflow and outflow boundaries, and an isothermal no-slip condition is used on the walls. Standard methods for calculating transport and thermodynamic properties are used. The simulations are conducted using HTR \cite{di2020htr}.

Shortly after the injection of methane and oxygen begins, a focused deposition of energy is deployed near the leading tip of the oxygen jet, as shown in \cref{fig:combustor}. This is modeled as an energy source $\dot{Q}_L$ in the governing equation for total energy,
\[
    \dot{Q}_L = B \frac{1}{2\pi\sigma_r^2}
    \exp\Big[-\frac{(x-x_0)^2+(y-y_0)^2}{2\sigma_r^2}\Big]
    \frac{2}{\sigma_t\sqrt{2\pi}}
    \exp\Big[-\frac{4(t-t_0)^2}{2\sigma_t^2}\Big]\,,
\]
where $B$ is the amplitude, $\sigma_r$ is the radius of the energy kernel, $\sigma_t$ is the duration of the energy pulse, $(x_0,y_0)$ is the focal location, and $t_0$ is the time of energy deposition. This produces a kernel of hot gas, seen in the \cref{fig:combustor} inset. Successful ignition depends on the parameters of the energy deposition as well as the local composition and flow conditions as the hot kernel cools and advects with the flow.

\begin{figure}
    \centering
    \includegraphics[scale=0.275]{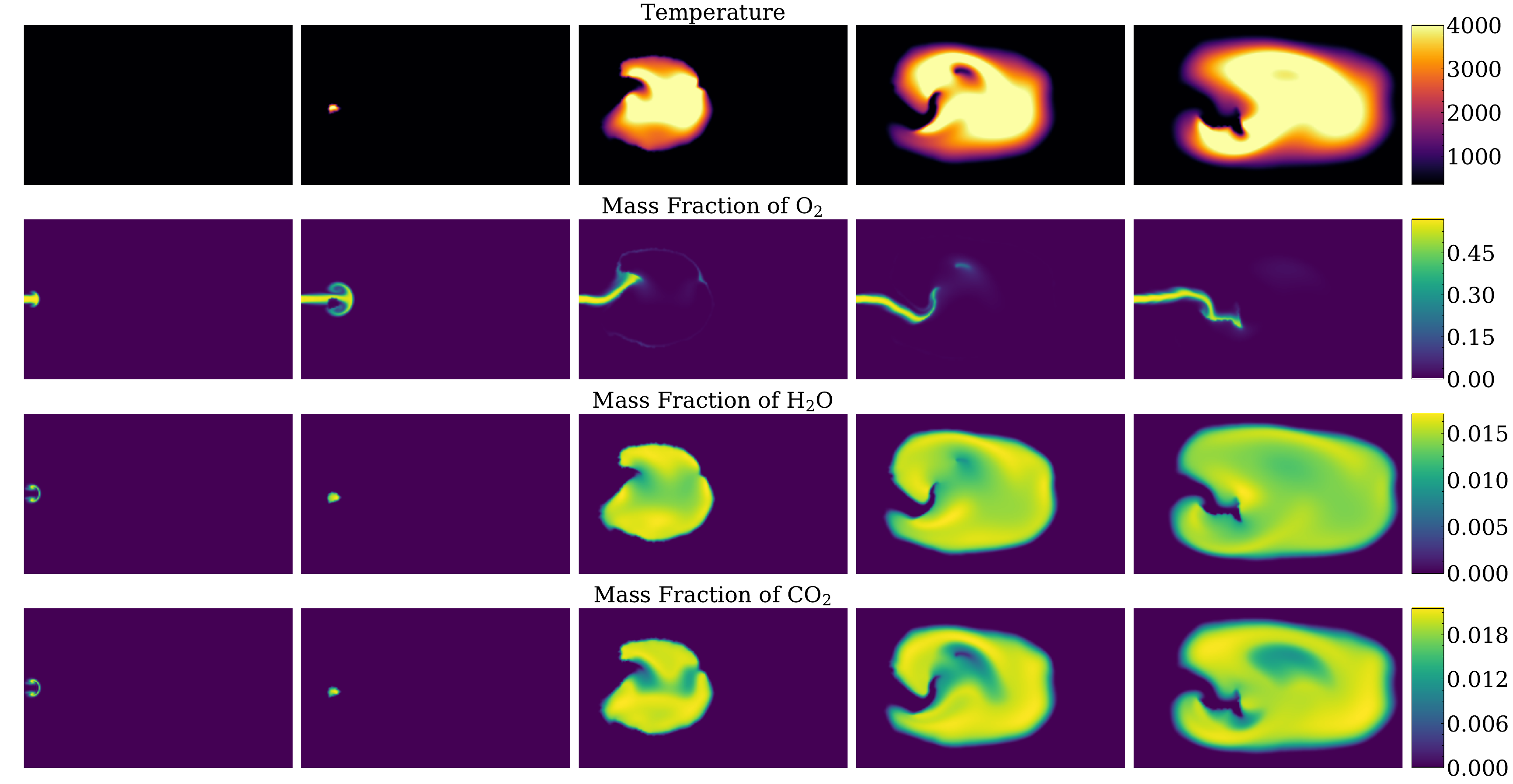}
    \caption{Evolution of the temperature, mass fraction of $\ce{O2}$, $\ce{H2O}$, and $\ce{CO2}$ in the 2D chamber with successful ignition. Snapshots of the temperature and mass fraction of different species are taken every 80 microseconds. Oxygen and methane are constantly injected into the reactor from the left side. A laser is deployed inside the chamber to induce ignition around 120 microseconds after the simulation starts. This triggers combustion mechanisms, gradually increasing the average temperature as more and more methane is ignited in the engine.}
    \label{fig:snapshots}
\end{figure}

\begin{table}[htbp]
\begin{center}
\begin{tabular}{ll,} 
 \toprule
 \hdr{l}{Parameter} & \hdr{l}{Definition} & \hdr{c}{Range} \\
 \midrule
 x & x coordinate of the center of the heat kernel & [0, \, 7.0] \\
 y  & y coordinate of the center of the heat kernel & [0, \, 1.0]\\
 amplitude & amount of energy deposited by the heat kernel & [0, \, 0.08] \\
 radius & spatial radius of the heat kernel & [0, \, 0.5] \\
 duration & duration of the heating & [0, \, 1.0] \\ 
 MaF & Mach number of the co-flowing \ce{CH4} jet & [0, \, 0.02]\\
 \bottomrule
\end{tabular}
\caption{Description of combustion parameters}
\label{tab:parameters}
\end{center}
\end{table}

\subsection{Planar jet diffusion simulation with fixed combustion parameters}

\begin{figure}[htbp]
\centering
  \begin{minipage}[b]{0.48\textwidth}
    \includegraphics[width=\textwidth]{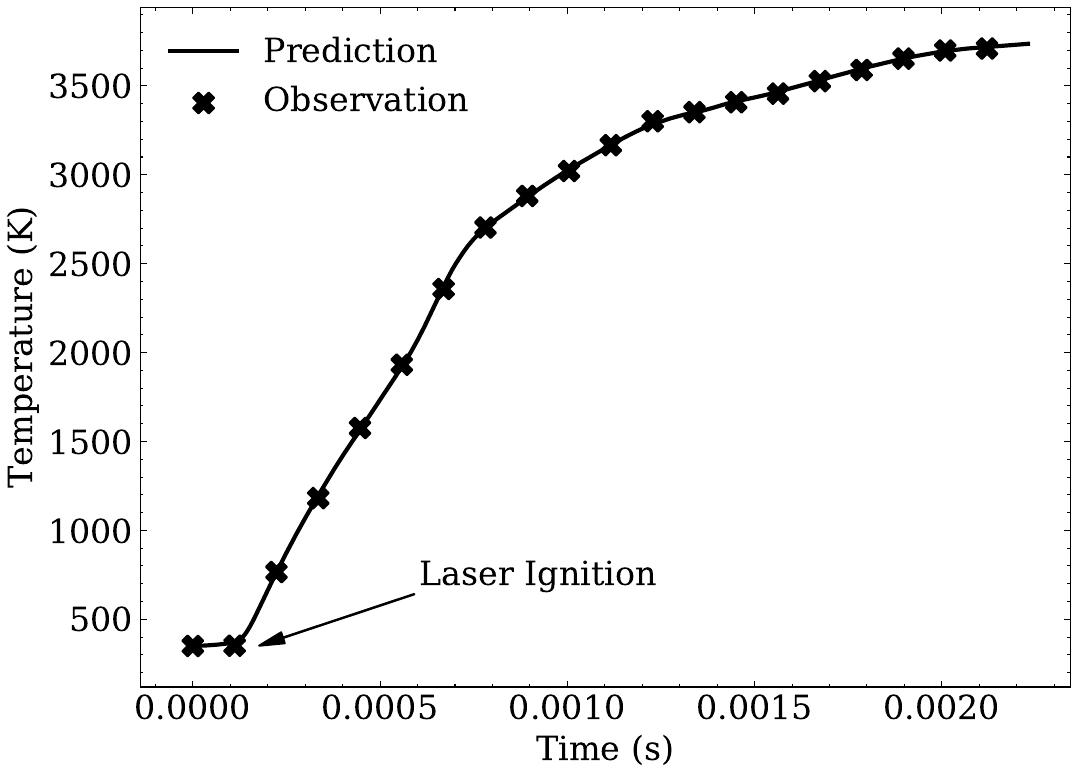}
    \caption{Prediction of average temperature by physics-based PNODE. The temperature initially stays at \SI{350}{K} and then rises gradually to \SI{3500}{K} after successful ignition. The evolution of volume-averaged temperature from PNODE matches closely with observations from simulations.}
    \label{fig:sing_dynamic_temp}
  \end{minipage}
  \hfill
  \begin{minipage}[b]{0.48\textwidth}
    \includegraphics[width=\textwidth]{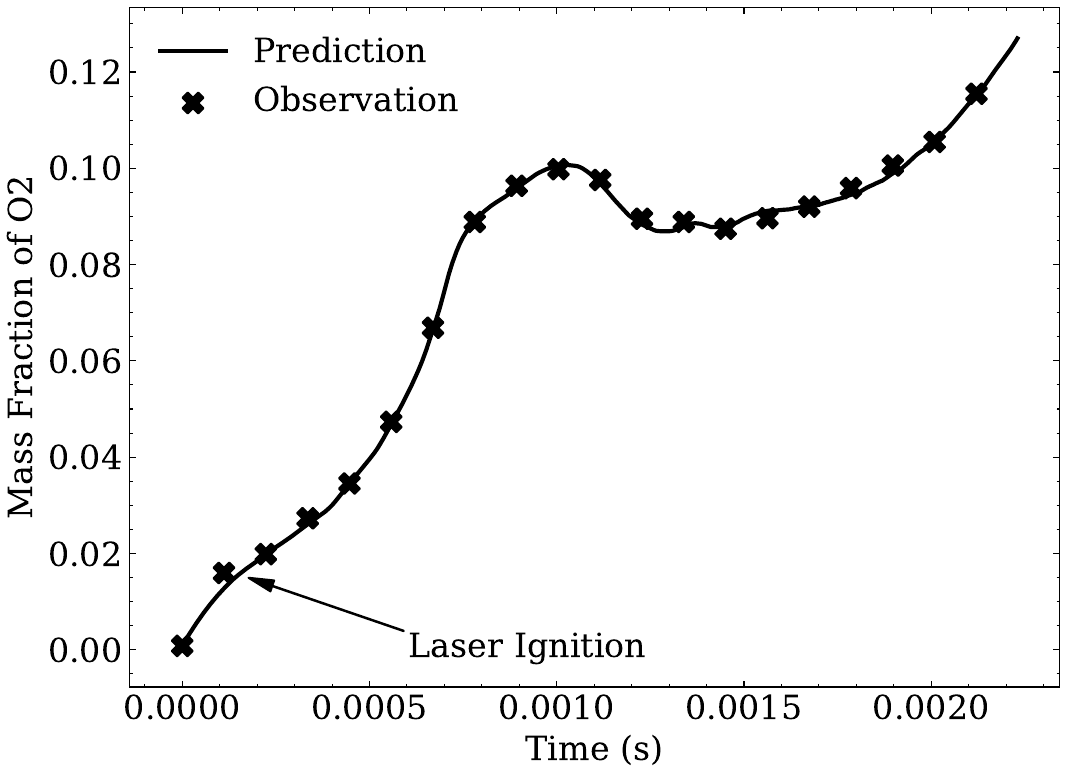}
    \caption{Prediction of mass fraction of \ce{O2} by physics-based PNODE. The mass fraction of oxygen gradually increases from 0 to 12\%, with a temporary decrease during laser ignition that triggers combustion mechanisms. Similarly, the prediction from PNODE matches closely with observations.}
    \label{fig:sing_dynamic_o2}
  \end{minipage}
\label{fig:single_dynamics_1}
\end{figure}

\begin{figure}[htbp]
\centering
  \begin{minipage}[b]{0.48\textwidth}
    \includegraphics[width=\textwidth]{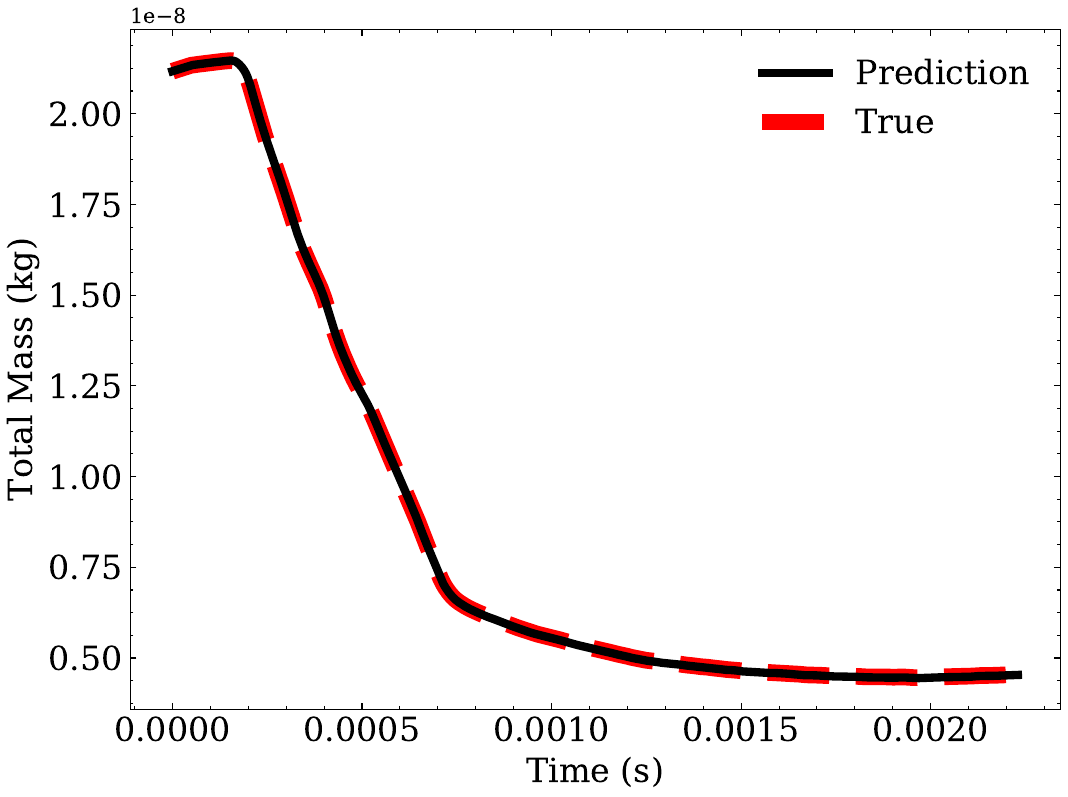}
    \caption{Prediction of total mass by physics-based PNODE. The overall mass inside the reactor constantly decreases due to mass outflow from the outlet. Since our physics-based PNODE enforces the law of mass conservation through \cref{eq:mass_conservation}, the total mass matches perfectly with observations.}
    \label{fig:sing_dynamic_mass}
  \end{minipage}
  \hfill
  \begin{minipage}[b]{0.48\textwidth}
    \includegraphics[width=\textwidth]{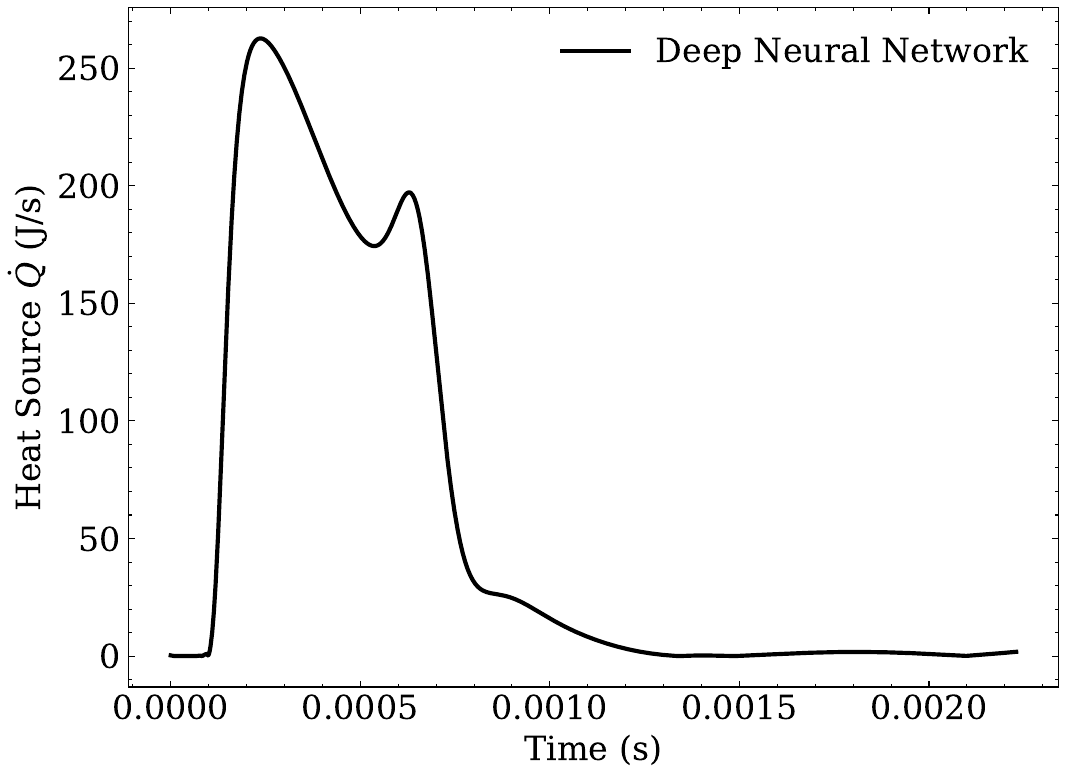}
    \caption{Heat source $\dot{Q}$ predicted by deep neural networks to represent the deposition of laser energy. Our physics-based PNODE calibrates the amount of heat deposited into the reactor so that the 0D model can simulate successful ignition from high-fidelity simulations.}
    \label{fig:sing_dynamic_heat}
  \end{minipage}
\label{fig:single_dynamics_2}
\end{figure}

We first show the performance of our physics-based PNODE model in learning a single trajectory from a planar jet diffusion simulation with successful ignition, which is shown in \cref{fig:snapshots}. \Cref{fig:sing_dynamic_temp,fig:sing_dynamic_o2} show the evolution of temperature and the mass fraction of oxygen over time. In this case, we choose $2 \times 10^{-5} \ \rm s$ as the time step of our numerical integration for our physics-based PNODE and choose a neural network with 2 hidden layers and 50 neurons in each layer as $F_{\theta}$. With 20 observations (\textit{i.e.,} points in simulation chosen to compute the mean squared errors) of the temperature and the mass fraction of oxygen, our physics-based PNODE can predict the evolution of both the average temperature and the mass fraction of oxygen with high accuracy. \Cref{fig:sing_dynamic_mass} shows the prediction of the total mass of the system over time. Note that by enforcing the law of mass conservation in \cref{eq:mass_conservation} with knowledge of the mass inflow and outflow, our physics-based PNODE can recover the change in total mass over time exactly. \Cref{fig:sing_dynamic_heat} shows the predicted heat source from the neural network as a function of time. We observe that the heat source in our 0D flow model represents the laser energy deposited. The predicted temperature from the PNODE continues to rise due to the energy released from chemical reactions after the peak of the heat source function, showing that our physics-based PNODE is able to learn not only the impact of the laser beam but also the chemical reactions from the combustion system. 

\subsection{Planar jet diffusion simulations with varying \texorpdfstring{$y$}{y} coordinate and amplitude of heat kernel}

\begin{figure}[htbp]
\centering
  \begin{minipage}[b]{0.48\textwidth}
    \includegraphics[width=\textwidth]{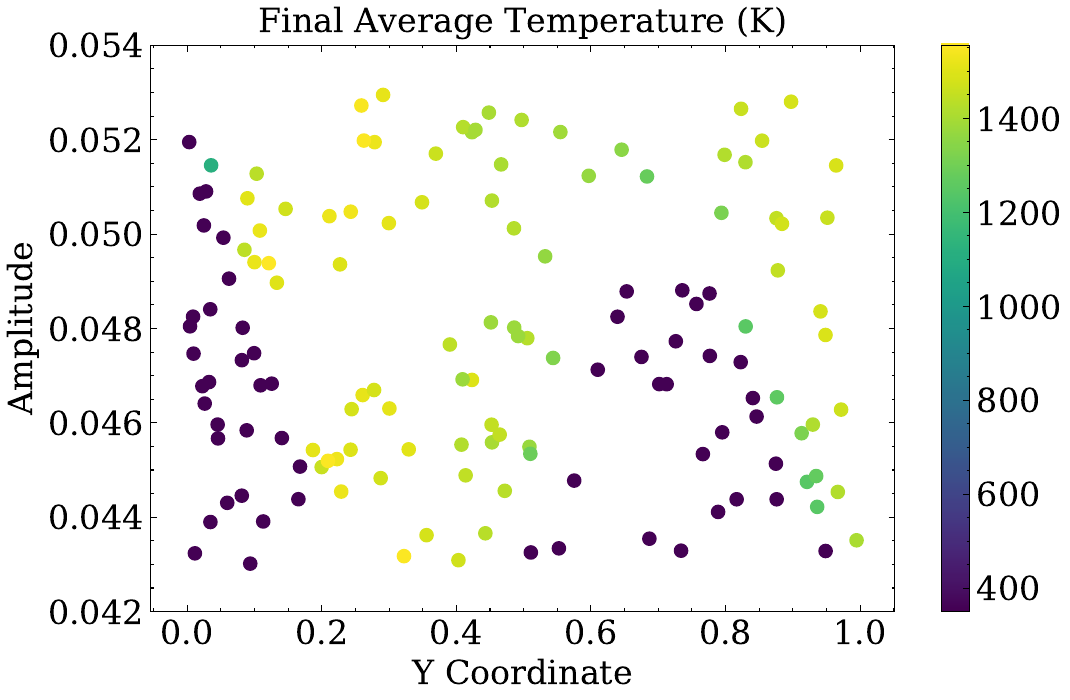}
    \caption{Final average temperature with different $y$ and amplitude of laser beam. We observe that there are sharp transitions near the boundary of successful ignition. The final temperature is equal to \SI{350}{K} in the non-ignition area but increases drastically to around \SI{1400}{K} once the parameters cross the S-shaped boundary as shown in the figure.}
    \label{fig:2d_simulation_final_temp}
  \end{minipage}
  \hfill
  \begin{minipage}[b]{0.48\textwidth}
    \includegraphics[width=\textwidth]{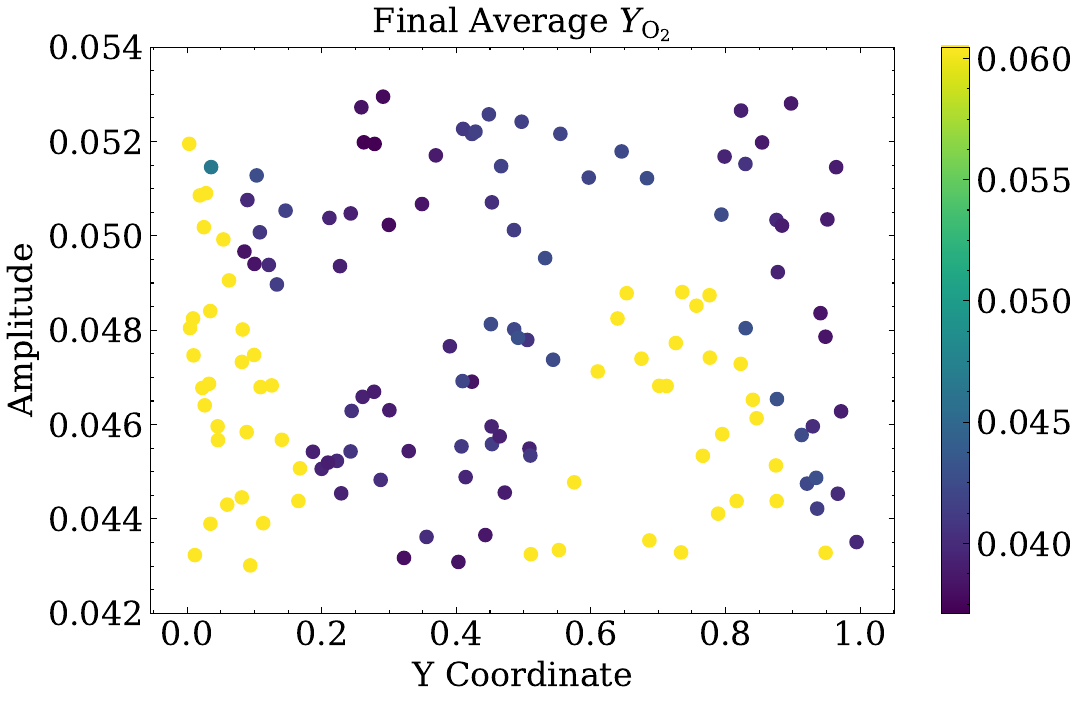}
    \caption{Final mass fraction of \ce{O2} with different $y$ and amplitude of laser beam. Similarly, we observe that the average mass fraction displays a binary nature. The final values are either 0.04 or 0.06 depending on the success of laser ignition. The parameters that lead to ignition can be separated from other parameters by an S-shaped curve.}
    \label{fig:2d_simulation_final_o2}
  \end{minipage}
\label{fig:2d_simulation_1}
\end{figure}

\begin{figure}[htbp]
\centering
  \begin{minipage}[b]{0.48\textwidth}
    \includegraphics[width=\textwidth]{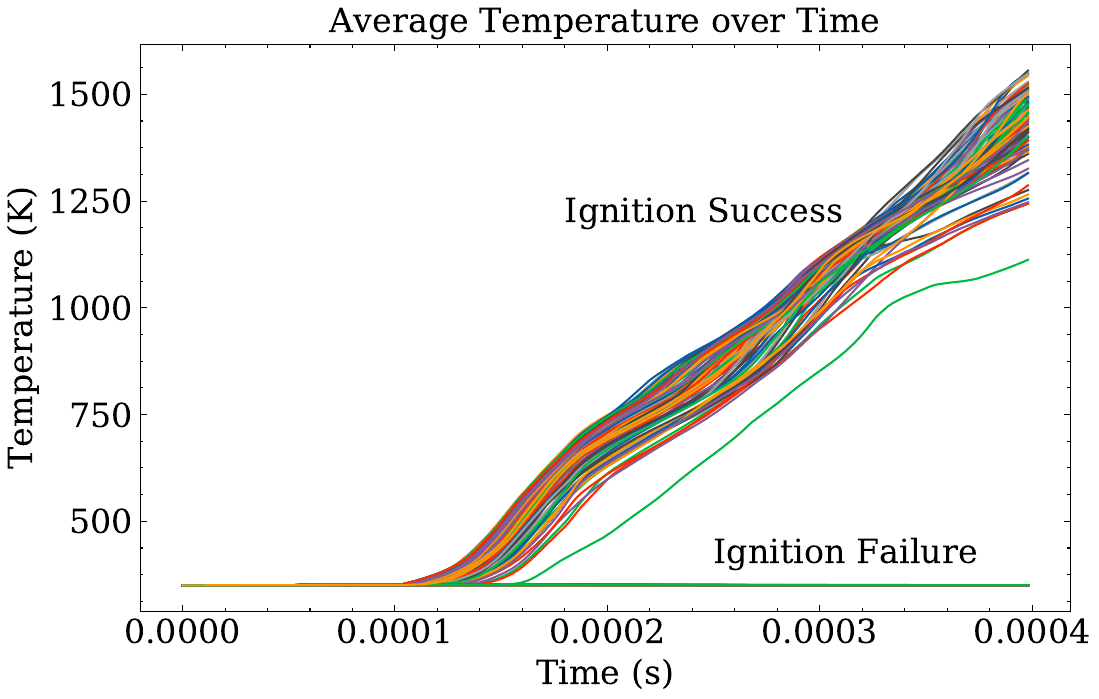}
    \caption{Evolution of average temperature in training data. The average temperature is initially at \SI{350}{K}. With successful ignition, the temperature gradually increases to values above \SI{1000}{K}. We observe that most ignitions happen between 100 and 150 microseconds after the simulation starts.}
    \label{fig:2d_simulation_temp_evolution}
  \end{minipage}
  \hfill
  \begin{minipage}[b]{0.48\textwidth}
    \includegraphics[width=\textwidth]{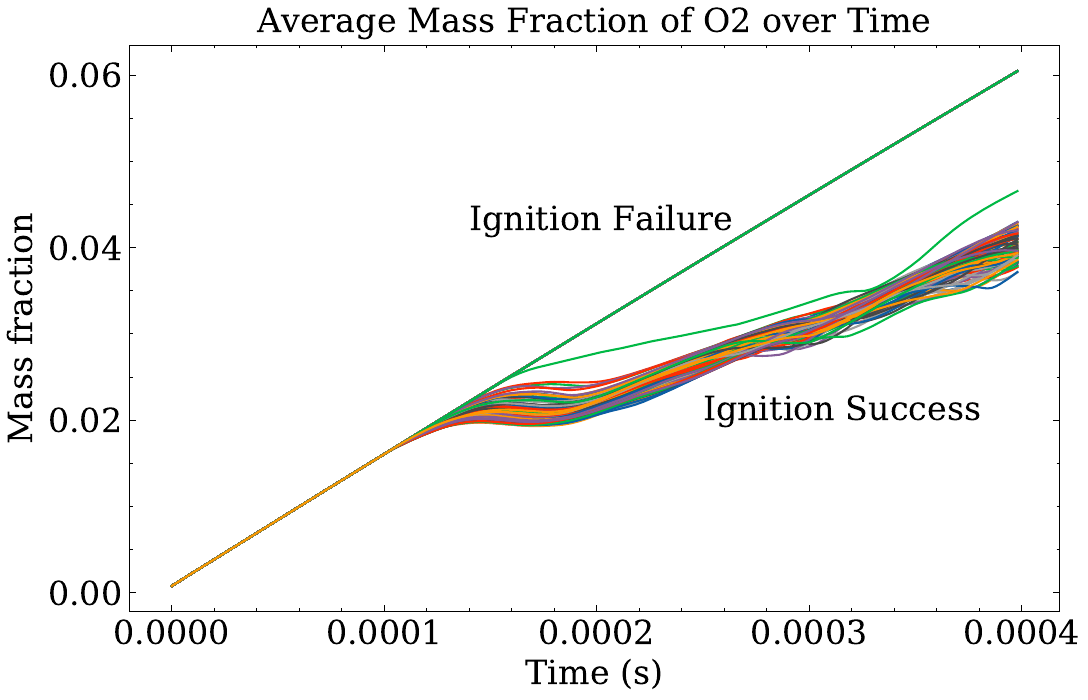}
    \caption{Evolution of mass fraction of \ce{O2} in training data. The mass fraction of \ce{O2} increases linearly first due to the injection of oxygen into the chamber. Successful ignition triggers chemical reactions that consume the oxygen, leading to a decrease in mass fraction of \ce{O2}}
    \label{fig:2d_simulation_o2_evolution}
  \end{minipage}
\label{fig:2d_simulation_2}
\end{figure}

In this section, we consider data generated from planar jet diffusion simulations with only two varying parameters: the $y$ coordinate of the location of the laser in the chamber and the amplitude of the laser beam. 143 data points are generated uniformly at random from selected intervals for the $y$ location and amplitude of the laser. The generated final data points are shown in \cref{fig:2d_simulation_final_temp,fig:2d_simulation_final_o2}, which show the distribution of the temperature and the mass fraction of \ce{O2}, respectively, at the end of the simulation with different $y$ locations and amplitudes. \Cref{fig:2d_simulation_temp_evolution,fig:2d_simulation_o2_evolution} show the evolution of the average temperature and the mass fraction of \ce{O2}, respectively, over time for all simulations in the training data. We observe that there are sharp transitions of temperature near the boundary of ignition success (\textit{i.e.,} the boundary of the subset of laser parameters where the final temperature is above \SI{1000}{K}) in the distribution when we vary two laser parameters. 

\begin{figure}[htbp]
\centering
  \begin{minipage}[b]{0.48\textwidth}
    \includegraphics[width=\textwidth]{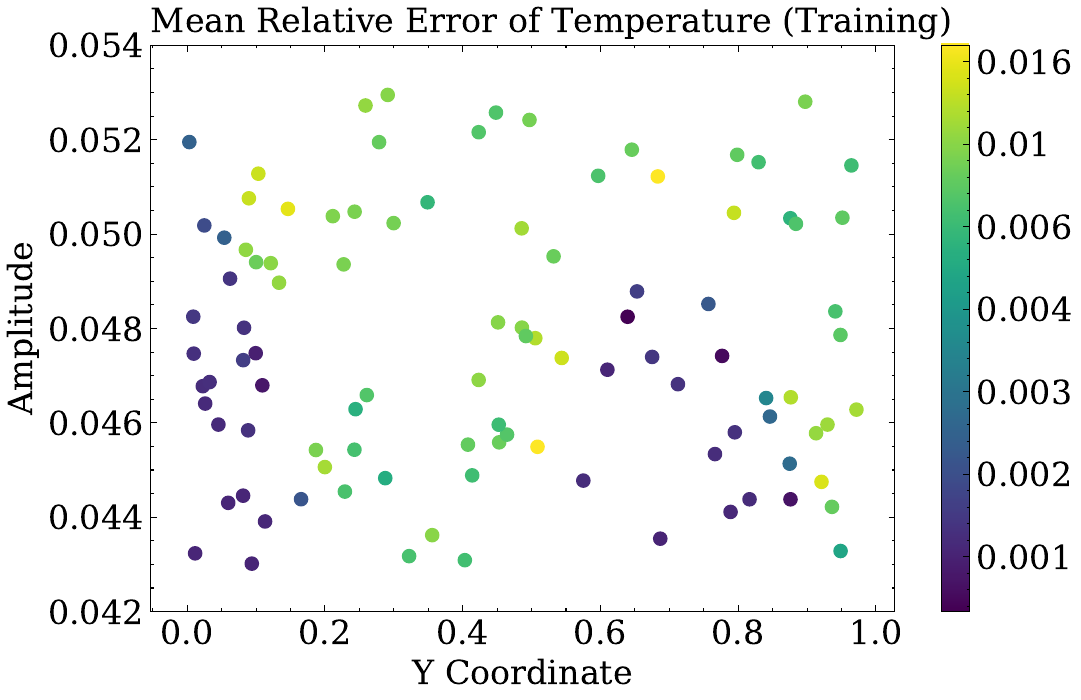}
    \caption{Mean relative error of temperature on training data. We observe that errors for data points in the non-ignition region are around 0.1\%, while errors for the ignition region are around 1.6\%. The observed difference is most likely attributed to the wider range of final temperatures ranging from \SI{1000}{K} to \SI{1400}{K} for successful ignition cases, in contrast to the final temperature of \SI{350}{K} across all training data points for non-ignition cases.}
    \label{fig:2d_simulation_temp_train}
  \end{minipage}
  \hfill
  \begin{minipage}[b]{0.48\textwidth}
    \includegraphics[width=\textwidth]{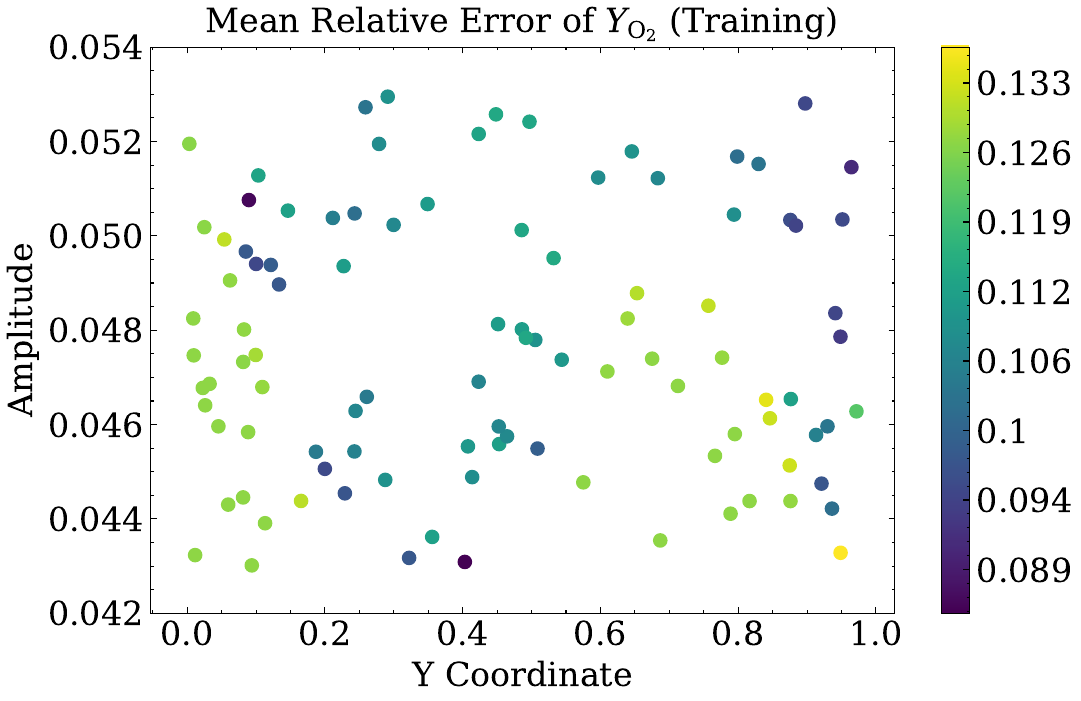}
    \caption{Mean relative error of mass fraction of \ce{O2} on training data. We observe that the error ranges from 8.9\% to 13.3\% across all the training data points. The evolution of the mass fraction of oxygen is more difficult for PNODE to learn, as it is continuously increasing due to the injection of the fuel. The mean relative errors for successful ignition cases are around 11.2\%, which is slightly lower than the values for non-ignition cases.}
    \label{fig:2d_simulation_o2_train}
  \end{minipage}
\label{fig:2d_simulation_3}
\end{figure}

\begin{figure}[htbp]
\centering
  \begin{minipage}[b]{0.48\textwidth}
    \includegraphics[width=\textwidth]{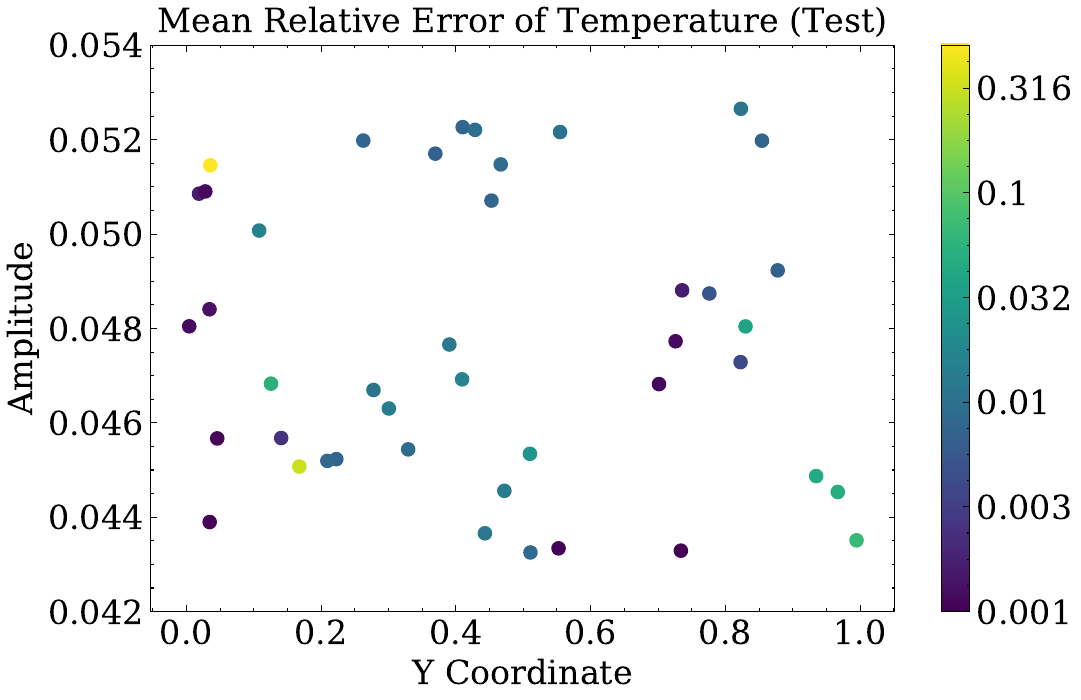}
    \caption{Mean relative error of temperature on test data. Most of the errors from our physics-based PNODE are less than 5\%. Due to sharp transitions of final temperature from \SI{350}{K} to \SI{1400}{K}, failure in predicting ignition success leads to large relative errors. Hence, we observe that there are mean relative errors around 31\% occur near the boundary of the ignition area.}
    \label{fig:2d_simulation_temp_test}
  \end{minipage}
  \hfill
  \begin{minipage}[b]{0.48\textwidth}
    \includegraphics[width=\textwidth]{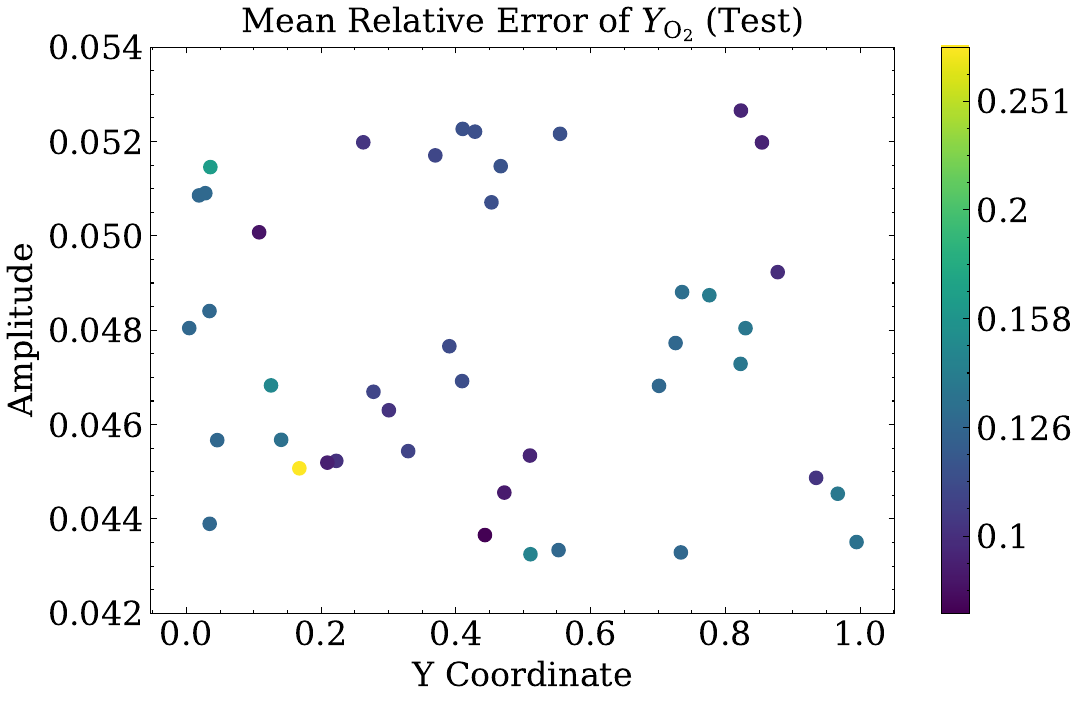}
    \caption{Mean relative error of mass fraction of \ce{O2} on test data. Most of the errors are less than 15\%. Similar to errors in final temperature, there are some errors larger than 20\% near the boundary of the ignition region because of failure in predicting ignition success. The overall errors are also higher than errors in temperature, as it is more difficult for PNODE to predict the mass fraction of oxygen.}
    \label{fig:2d_simulation_o2_test}
  \end{minipage}
\label{fig:2d_simulation_4}
\end{figure}

We randomly selected 100 data points out of 143 samples to train our PNODE-based 0D model with neural networks with the hyperparameters shown in \cref{tab:hyperparameters}. Since we are using a simple feed-forward neural network, we manually optimize the structure of our PNODE based on its performance on 20 samples from a validation dataset randomly chosen from 100 training samples. The performance of our PNODE on seen parameters (training data) is shown in \cref{fig:2d_simulation_temp_train,fig:2d_simulation_o2_train}, which show the mean relative error of temperature and mass fraction of oxygen over time. We observe that the prediction of our neural  ODE model has high accuracy on all training data points with a mean relative error of temperature lower than 1.6\% and mean relative error for the mass fraction of oxygen lower than 13.3\%. We also test the performance of our PNODE-based 0D model with the remaining 43 unseen samples from our data set. The results are shown in \cref{fig:2d_simulation_temp_test,fig:2d_simulation_o2_test}. Our PNODE-based model is able to predict ignition success or failure accurately for all points far away from the boundary of the ignition area. Since there are sharp transitions from a temperature of \SI{350}{K} to a temperature higher than \SI{1000}{K} near the boundary of the area of successful ignition, false prediction of either ignition or non-ignition will both lead to large errors in mean relative error of temperature and mass fraction of oxygen. Hence, we observe that there are some errors by our physics-based PNODE for some of the test data points close to the boundary of the ignition area. Nevertheless, our PNODE-based model is still able to capture the sharp transition from the non-ignition region to the ignition region, which is often difficult to learn with conventional interpolation methods. 

\begin{table}[htbp]
\begin{center}
\begin{tabular}{ll} 
 \toprule
 Hyperparameter & Value \\ \midrule
 Number of hidden layers  & 2 \\
 Number of neurons in each layer  &  300\\
 Activation function & tanh \\
 Optimization algorithm & L-BFGS-B \\
 \bottomrule
\end{tabular}
\caption{Hyperparameters for PNODE}
\label{tab:hyperparameters}
\end{center}
\end{table}

\subsection{Planar jet diffusion simulations with six varying combustion parameters}

We now consider planar jet diffusion simulations with all six varying combustion parameters. We generate training data by sampling within selected intervals for each of the six parameters uniformly at random as specified in \cref{tab:parameters}. To illustrate both the expressiveness and robustness of our PNODE, we use the same hyper-parameters in \cref{tab:hyperparameters} and train our PNODE with data sets of two different sizes: one with 100 samples and one with 4,000 samples. Both data sets are generated by random uniform sampling from selected intervals for each of the six parameters in \cref{tab:parameters}. 

We test the performance of our physics-based PNODE on test data that represent cross sections of selected two dimensions of our six-dimensional parameter space. More precisely, after choosing two parameters out of six parameters for validation, we collect test data on a 30 by 30 grid for those two parameters while fixing the value of the remaining four parameters. In particular, we choose the following three pairs of parameters for test data: 
\begin{itemize}
    \item radius and amplitude
    \item duration and amplitude
    \item $y$ coordinate and amplitude
\end{itemize}

We also compare the results of our PNODE with kernel ridge regression and neural networks. 20\% of the samples in each data set are used as validation data. To illustrate the improvement from our physics-based PNODE, we deliberately choose the same network structure for the neural network as the ones embedded in our 0D flow model. For kernel ridge regression, we use a radial basis function with $\alpha = 2$ and $\gamma = 1$ from the scikit-learn package, which is manually optimized based on its performance on the validation set. The kernel ridge regression and classical neural network interpolate the temperature at the end of simulations in the six-dimensional parameter space. That is, they take the combustion parameter $\eta$ as input and predict the final temperature $T$ as output, which is a significantly simpler task than predicting the evolution of the system, including temperature, pressure, and mass fractions of species. We tested the performance of the three models on additional 100 test samples based on the final temperature. The mean relative errors of the PNODE, kernel ridge regression, and neural network are shown in \cref{tab:benchmark}. The physics-based PNODE provides the most accurate predictions with both 100 training samples and 4,000 training samples.

\begin{table}[htbp]
\begin{center}
\begin{tabular}{ldd} 
 \toprule
 & \hdr{c}{One hundred samples} & \hdr{c}{Four thousand samples} \\
 \midrule
 Physics-based PNODE  & \bd{40.36\%}  & \bd{28.53\%} \\
 Neural Network  & 88.76\%  & 61.57\%\\
 Kernel Ridge Regression  & 46.06\%  & 33.95\% \\
 \bottomrule
\end{tabular}
\caption{Mean relative error of prediction on the final average temperature by three models in the test data}
\label{tab:benchmark}
\end{center}
\end{table}

\begin{figure}[htbp]
\centering
\begin{tabular}{ll}
\includegraphics[scale=0.45]{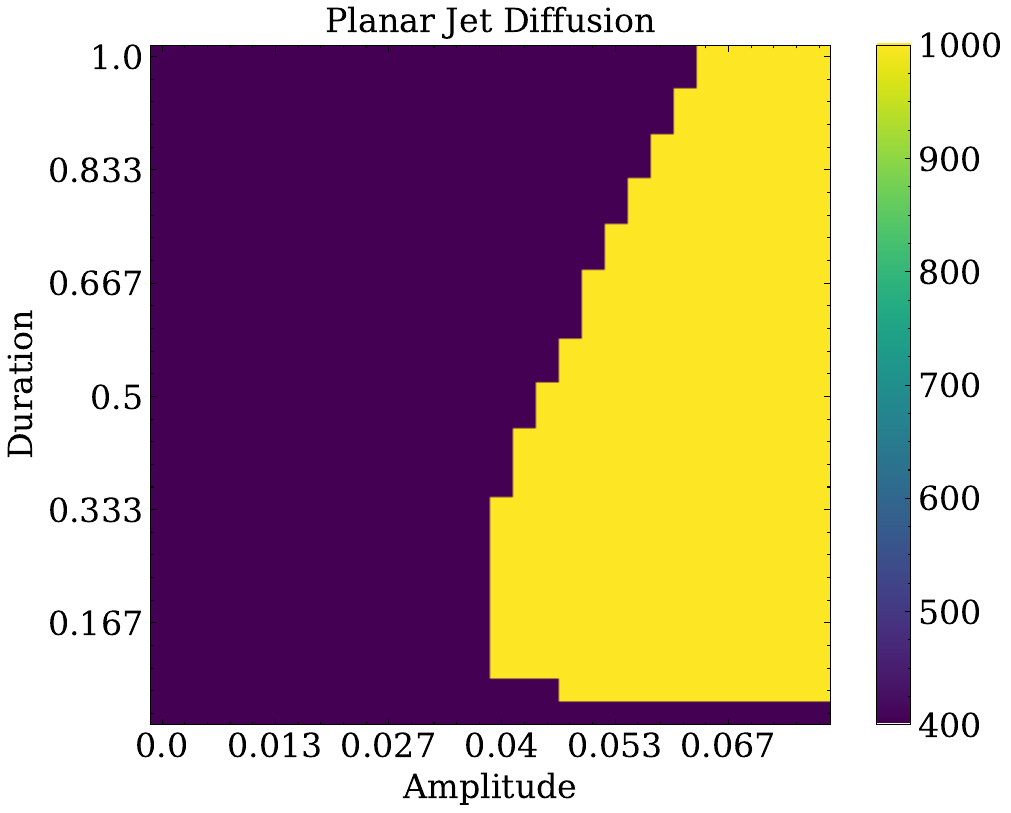}
&
\includegraphics[scale=0.45]{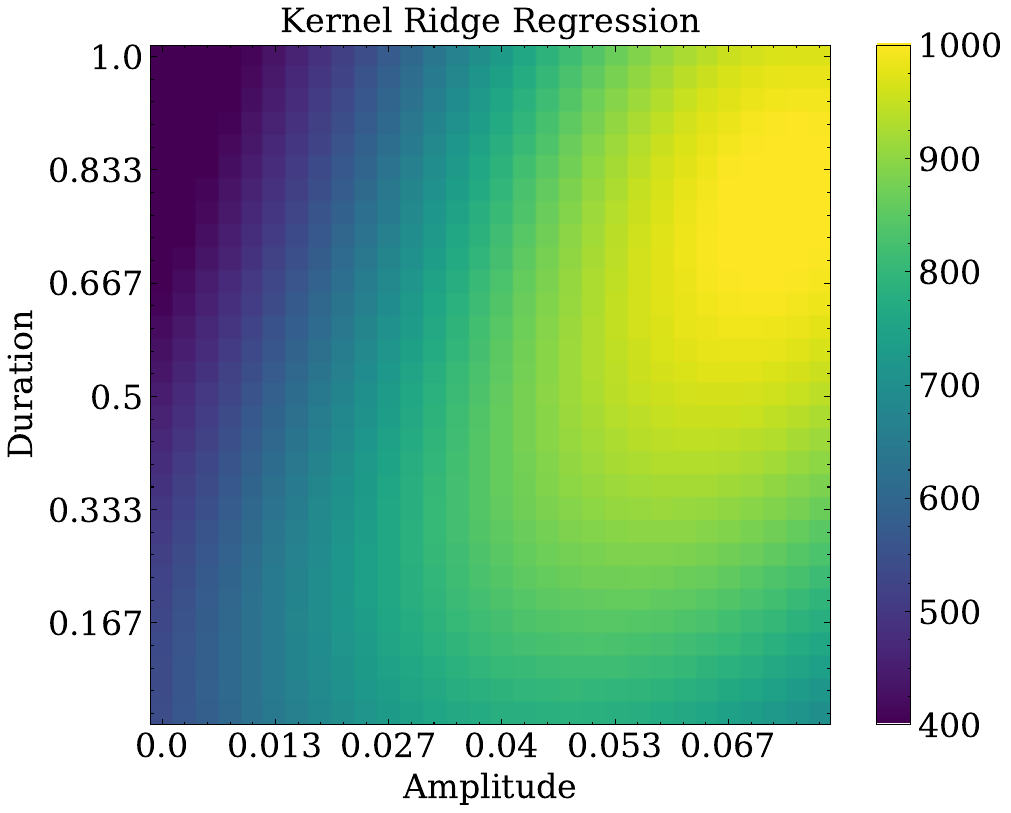}
\\
\includegraphics[scale=0.45]{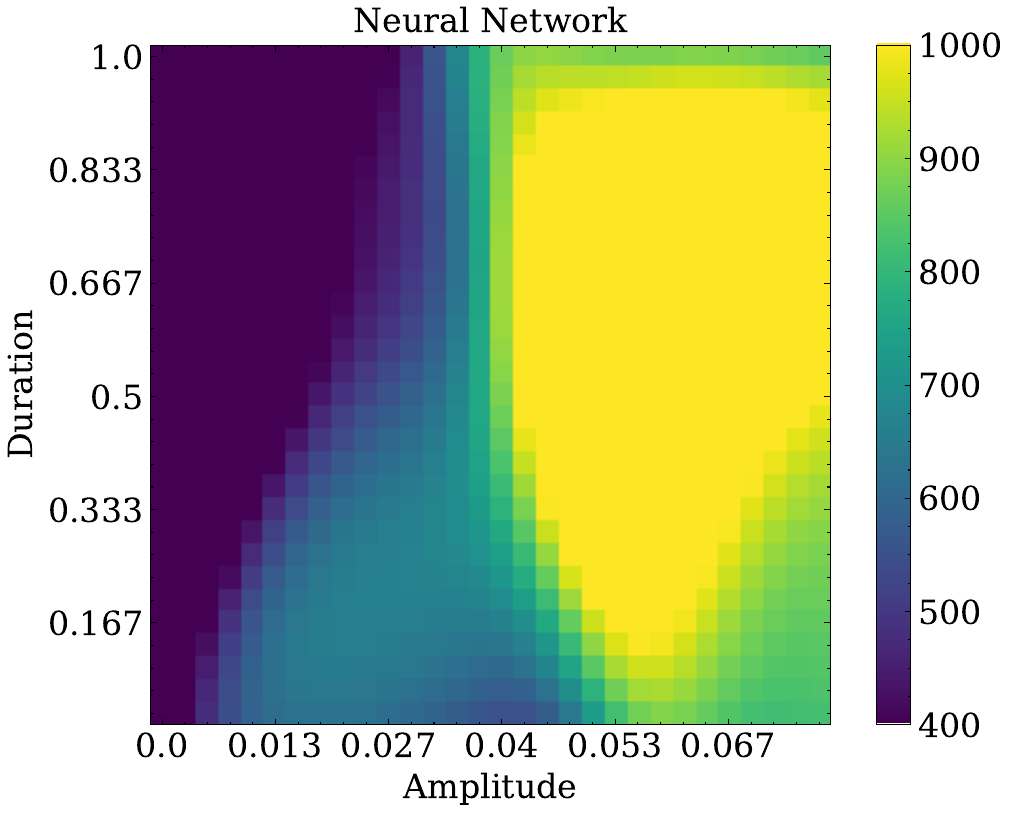}
&
\includegraphics[scale=0.45]{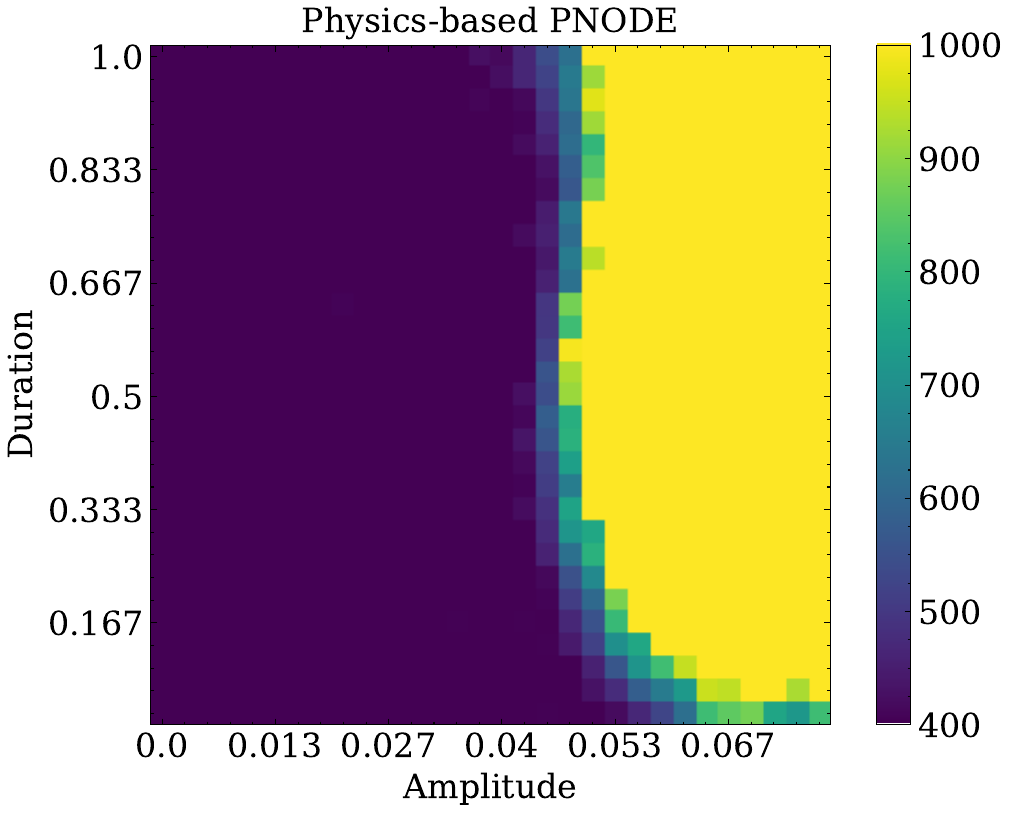}
\end{tabular}
\caption{Prediction of the final temperature by the three models with 100 training samples on simulations with varying amplitude and duration of the laser beam. Top left: ground truth by Planar Jet Diffusion Simulations. Top right: prediction by kernel ridge regression. Bottom left: prediction by neural networks. Bottom right: prediction by physics-based PNODE. We observe that, with only 100 training data points, our physics-based PNODE provides the most accurate prediction of final temperature among all three methods, with a sharp boundary between ignition and non-ignition regions. On the contrary, kernel ridge regression, and neural networks predict the final temperature of many data points to be between \SI{400}{K} and \SI{1000}{K}, which is inconsistent with our high-fidelity simulations.}
\label{fig:amplitude_duration}
\end{figure}

\begin{figure}[htbp]
\centering
\begin{tabular}{ll}
\includegraphics[scale=0.45]{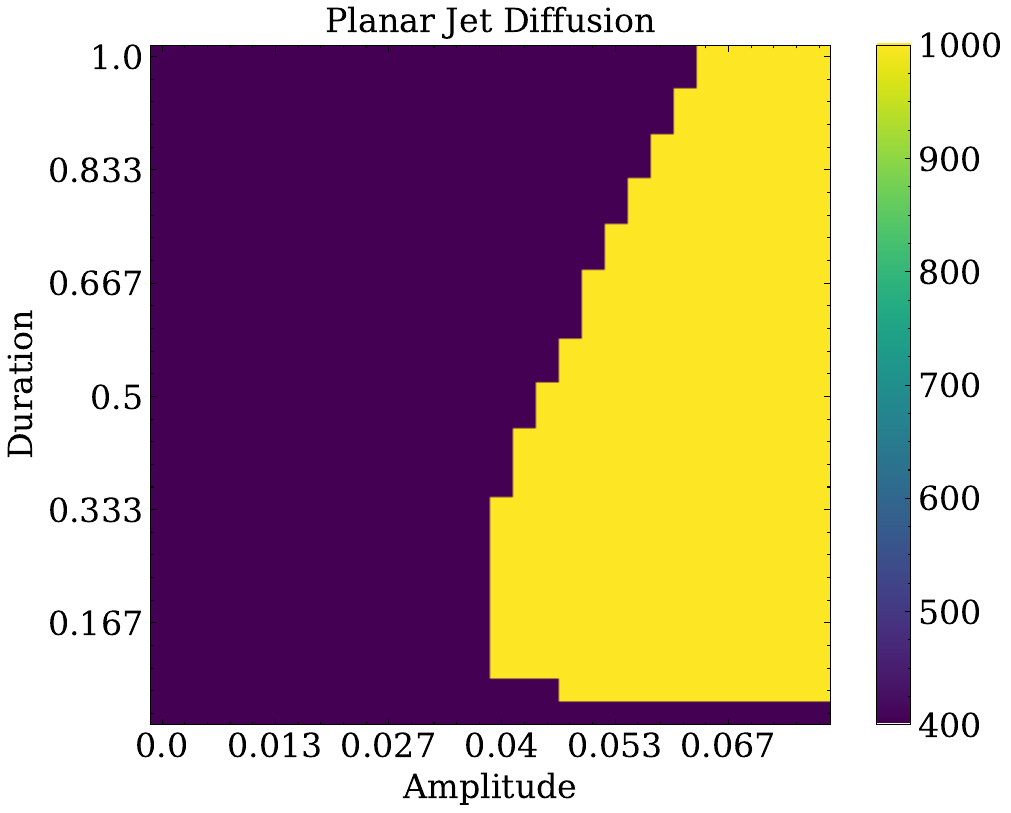}
&
\includegraphics[scale=0.45]{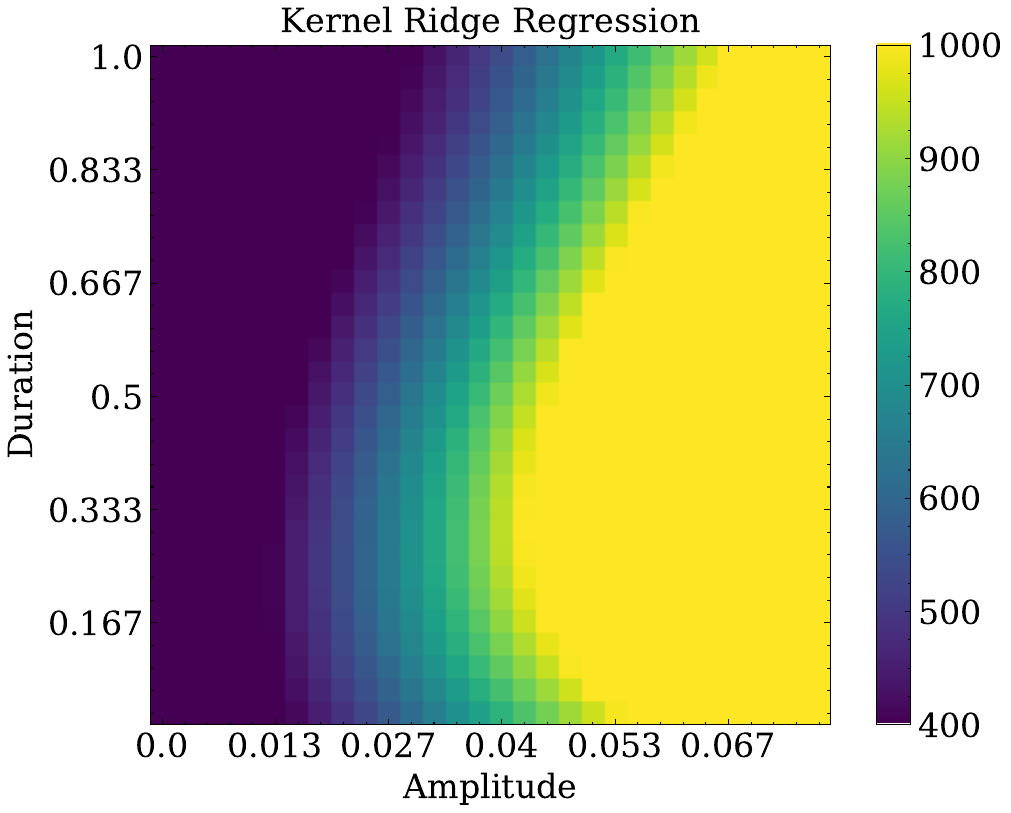}
\\
\includegraphics[scale=0.45]{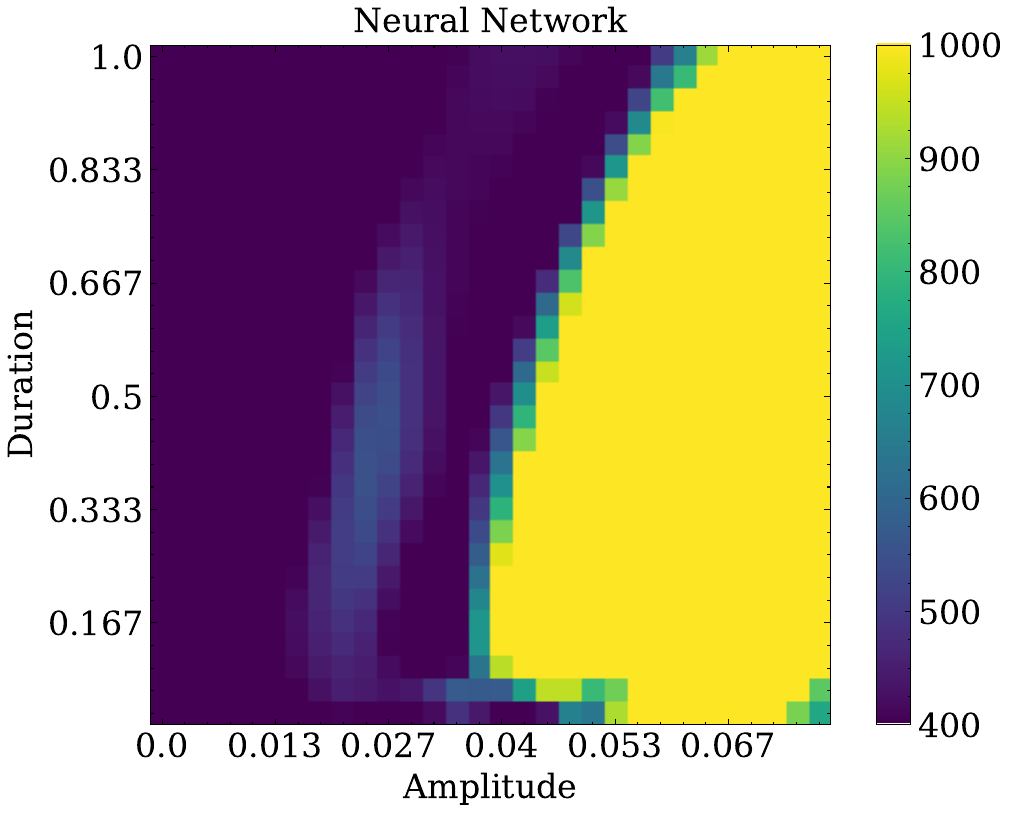}
&
\includegraphics[scale=0.45]{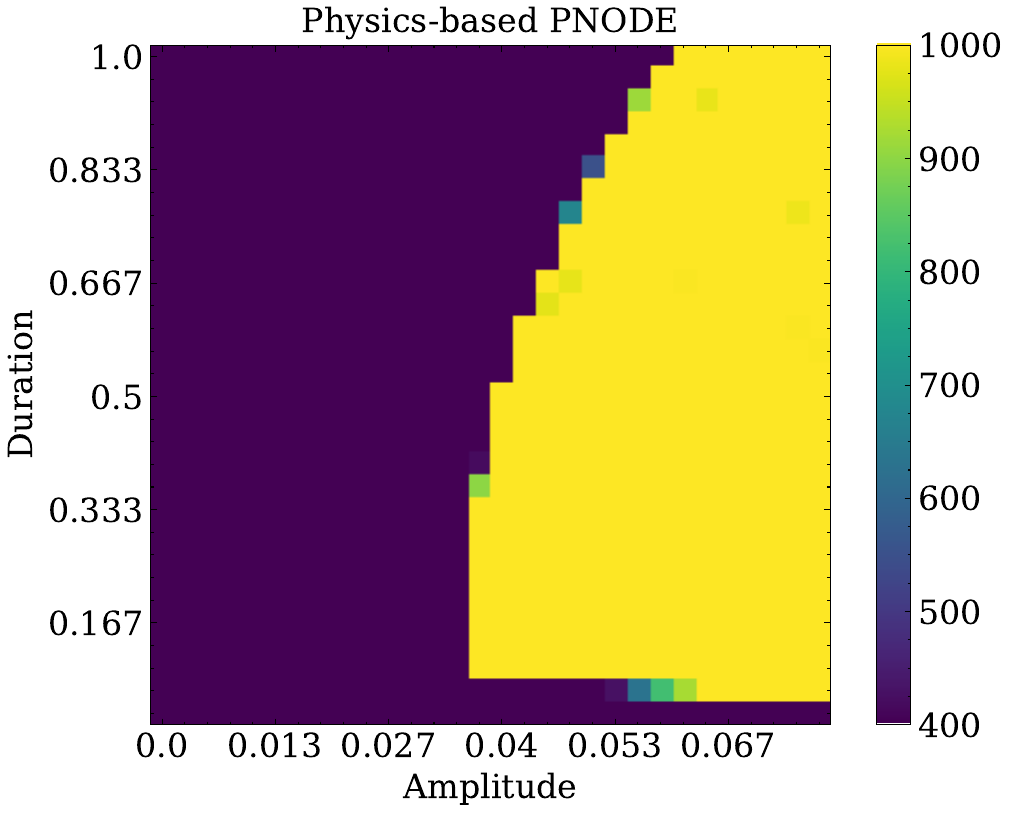}
\end{tabular}
\caption{Prediction of the final temperature by the three models with 4,000 training samples on simulations with varying amplitude and duration of the laser beam. Top left: ground truth by Planar Jet Diffusion Simulations. Top right: prediction by kernel ridge regression. Bottom left: prediction by neural networks. Bottom right: prediction by physics-based PNODE. We observe that, with only 100 training data points, our physics-based PNODE provides the most accurate prediction of final temperature among all three methods, with a sharp boundary between ignition and non-ignition regions. On the contrary, kernel ridge regression, and neural networks predict the final temperature of many data points to be between \SI{400}{K} and \SI{1000}{K}, which is inconsistent with our high-fidelity simulations.}
\label{fig:amplitude_duration_4000}
\end{figure}

\begin{figure}[htbp]
\centering
\begin{tabular}{ll}
\includegraphics[scale=0.45]{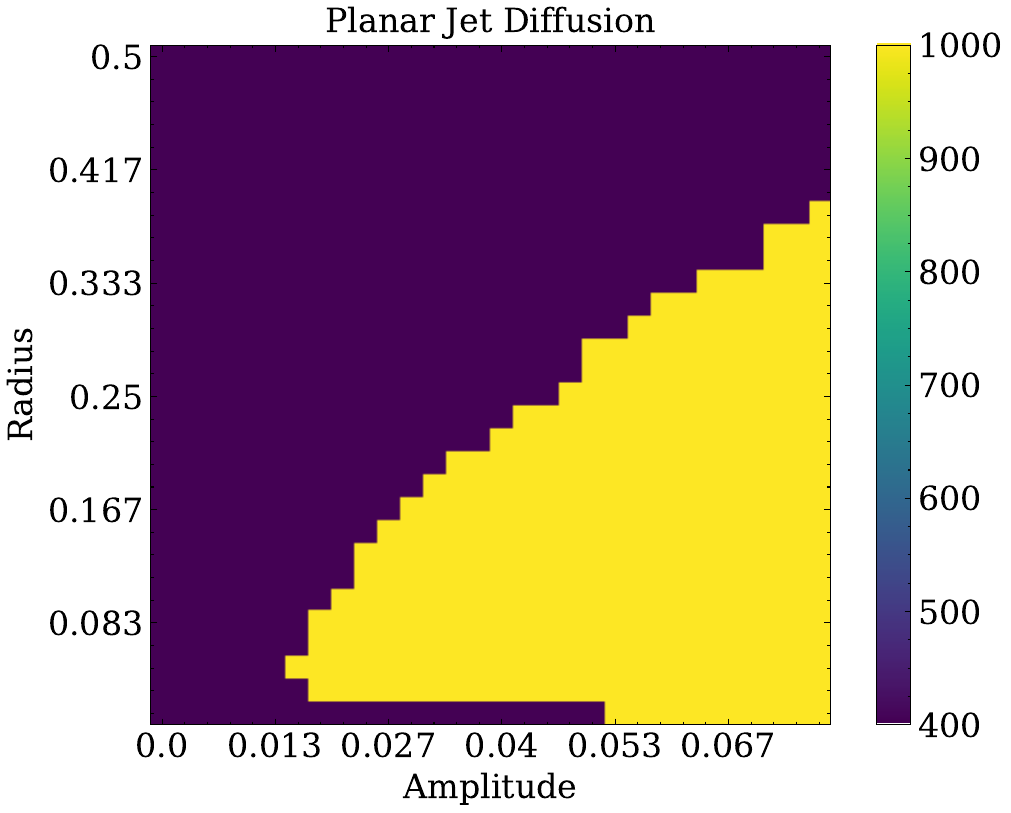}
&
\includegraphics[scale=0.45]{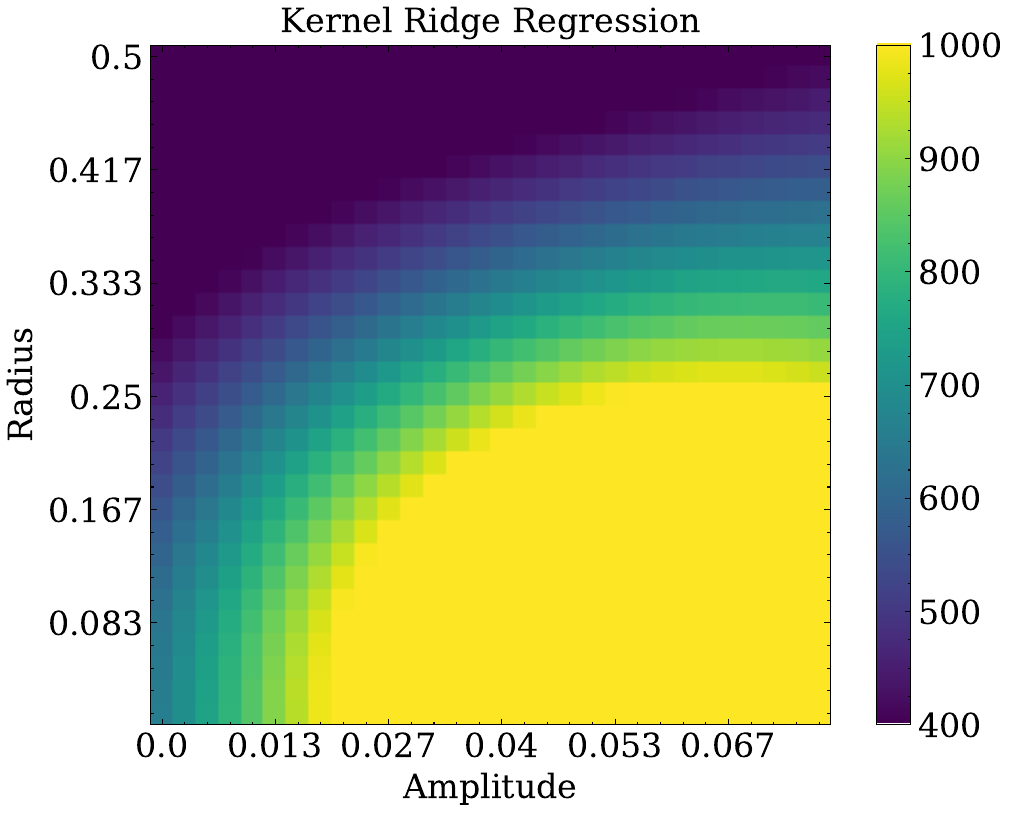}
\\
\includegraphics[scale=0.45]{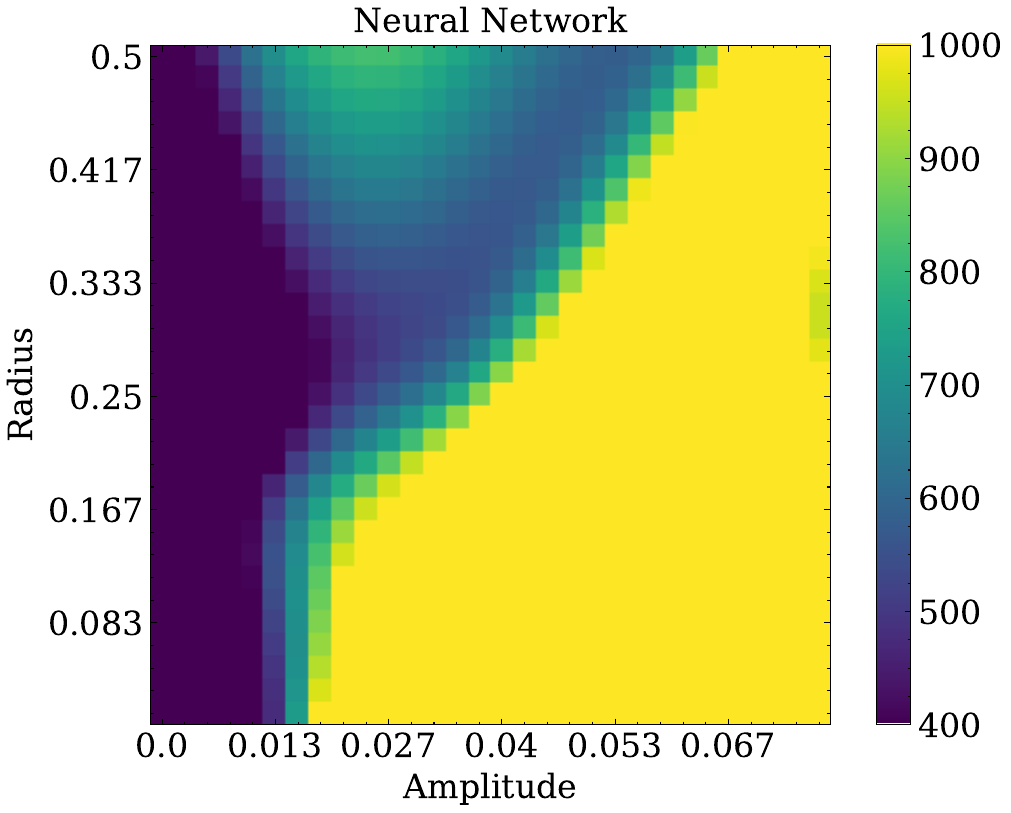}
&
\includegraphics[scale=0.45]{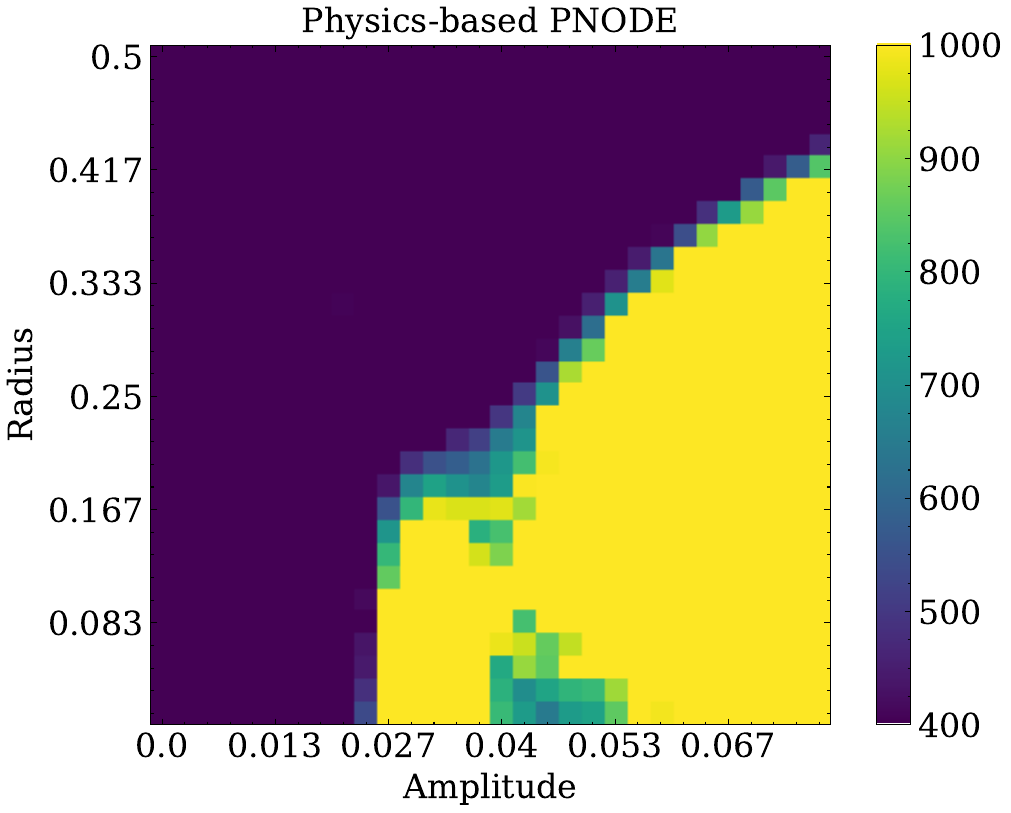}
\end{tabular}
\caption{Prediction of the final temperature by the three models with 100 training samples on simulations with varying amplitude and radius of the laser beam. Top left: ground truth by Planar Jet Diffusion Simulations. Top right: prediction by kernel ridge regression. Bottom left: prediction by neural networks. Bottom right: prediction by physics-based PNODE. We observe that, with only 100 training data points, our physics-based PNODE provides the most accurate prediction of final temperature among all three methods, with a sharp boundary between ignition and non-ignition regions. On the contrary, kernel ridge regression, and neural networks predict the final temperature of many data points to be between \SI{400}{K} and \SI{1000}{K}, which is inconsistent with our high-fidelity simulations.}
\label{fig:amplitude_radius}
\end{figure}

\begin{figure}[htbp]
\centering
\begin{tabular}{ll}
\includegraphics[scale=0.45]{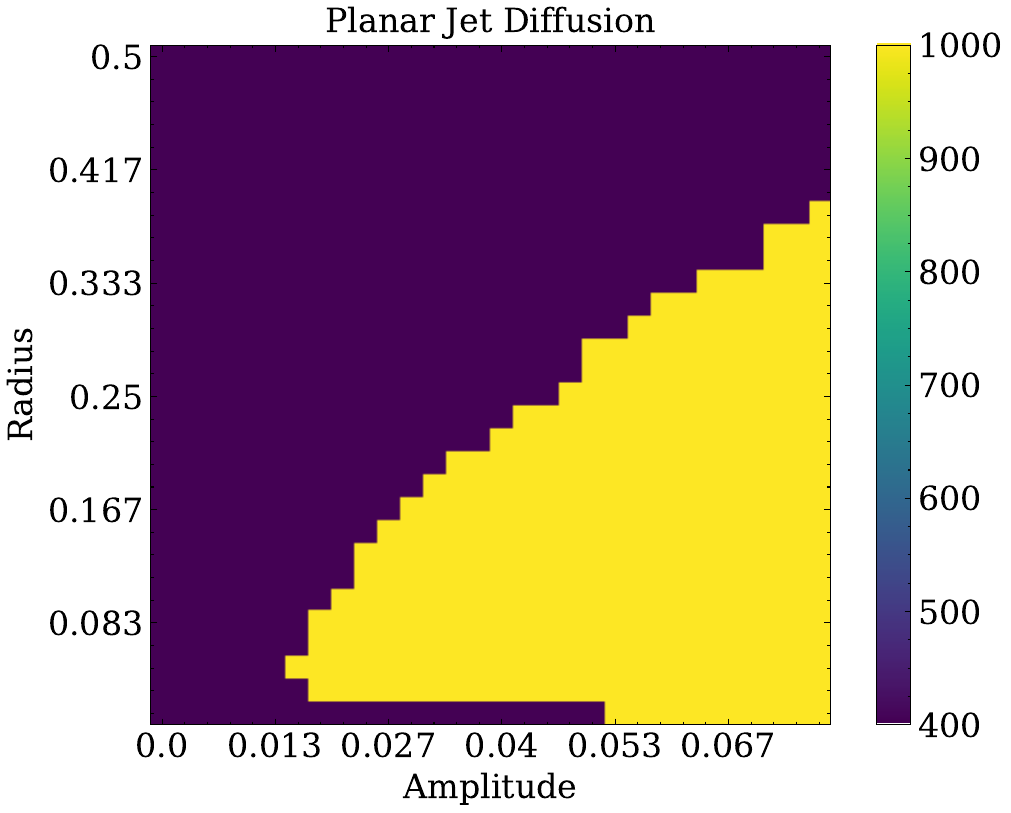}
&
\includegraphics[scale=0.45]{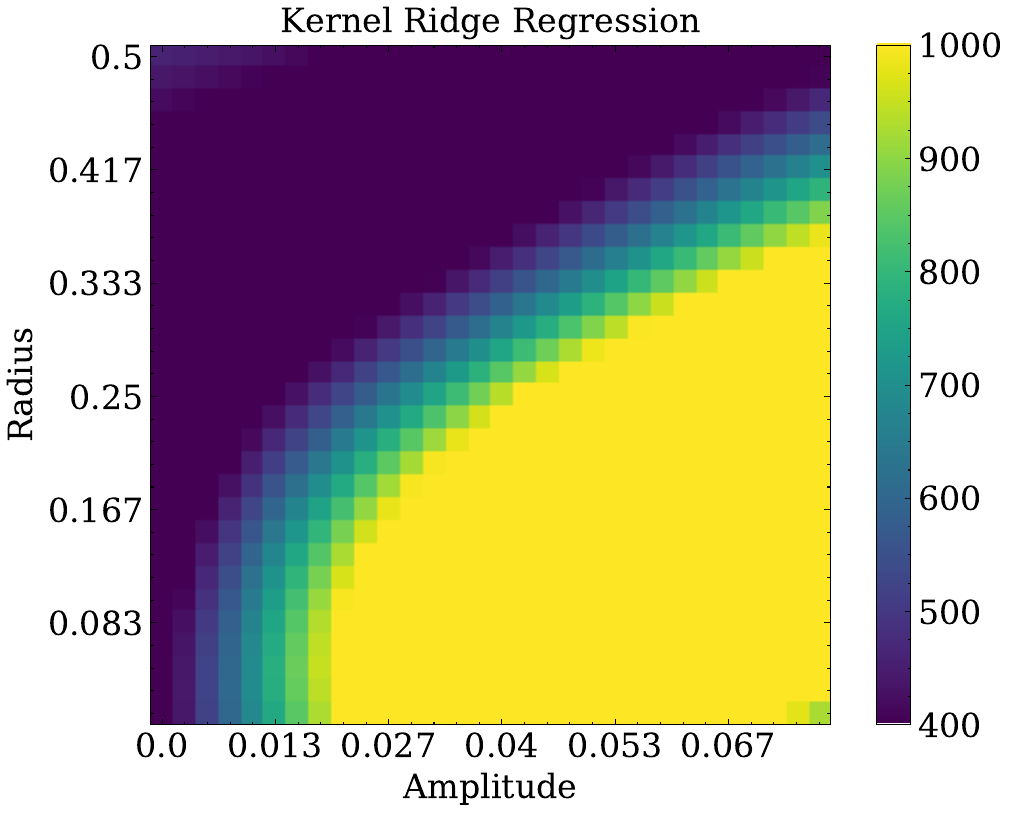}
\\
\includegraphics[scale=0.45]{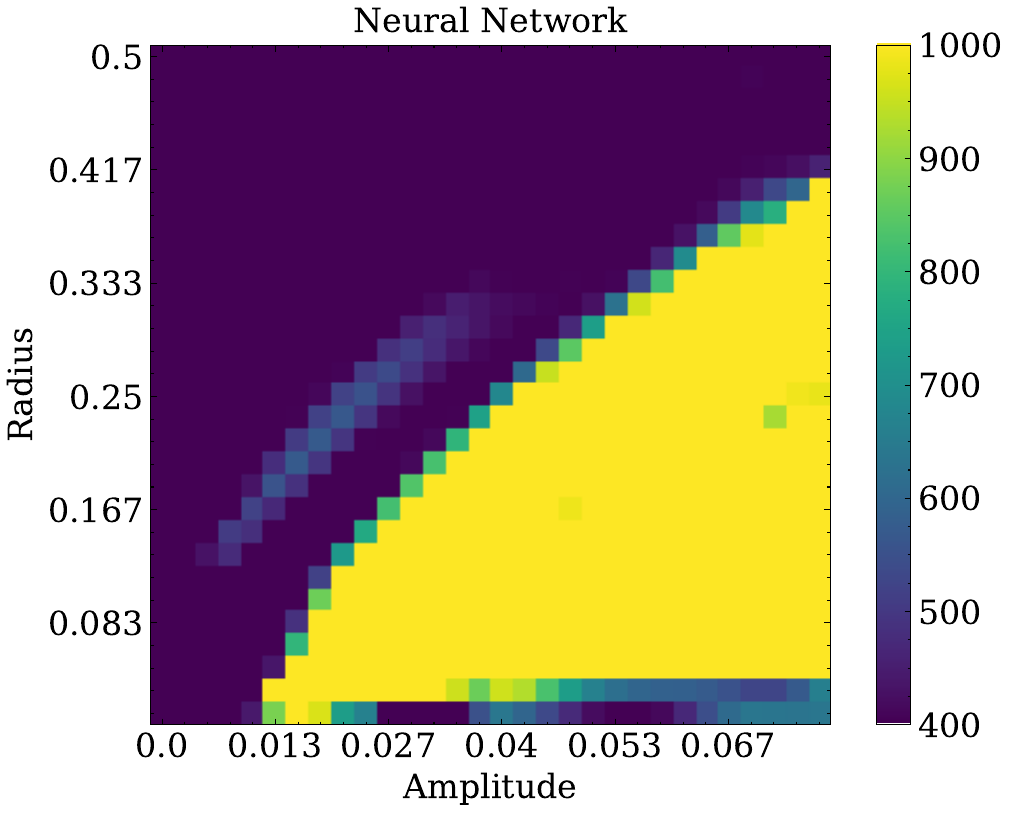}
&
\includegraphics[scale=0.45]{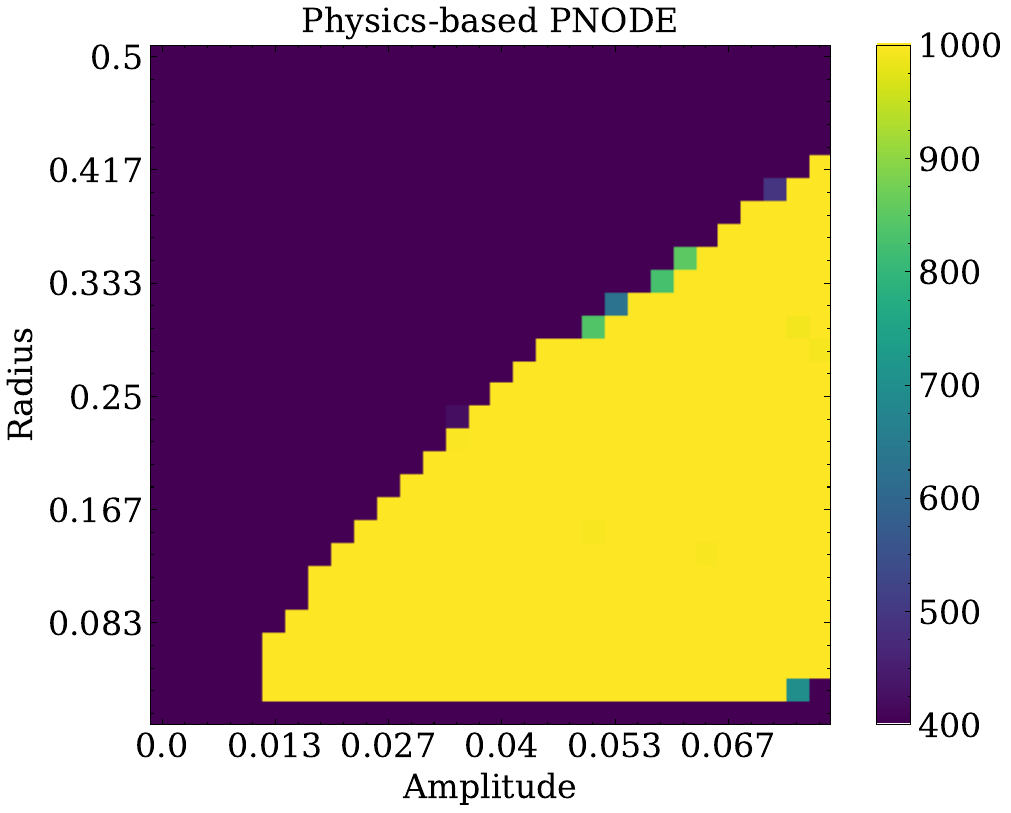}
\end{tabular}
\caption{Prediction of the final temperature by the three models with 4,000 training samples on simulations with varying amplitude and radius of the laser beam. Top left: ground truth by Planar Jet Diffusion Simulations. Top right: prediction by kernel ridge regression. Bottom left: prediction by neural networks. Bottom right: prediction by physics-based PNODE. We observe that, with only 100 training data points, our physics-based PNODE provides the most accurate prediction of final temperature among all three methods, with a sharp boundary between ignition and non-ignition regions. On the contrary, kernel ridge regression, and neural networks predict the final temperature of many data points to be between \SI{400}{K} and \SI{1000}{K}, which is inconsistent with our high-fidelity simulations.}
\label{fig:amplitude_radius_4000}
\end{figure}

\begin{figure}[htbp]
\centering
\begin{tabular}{ll}
\includegraphics[scale=0.45]{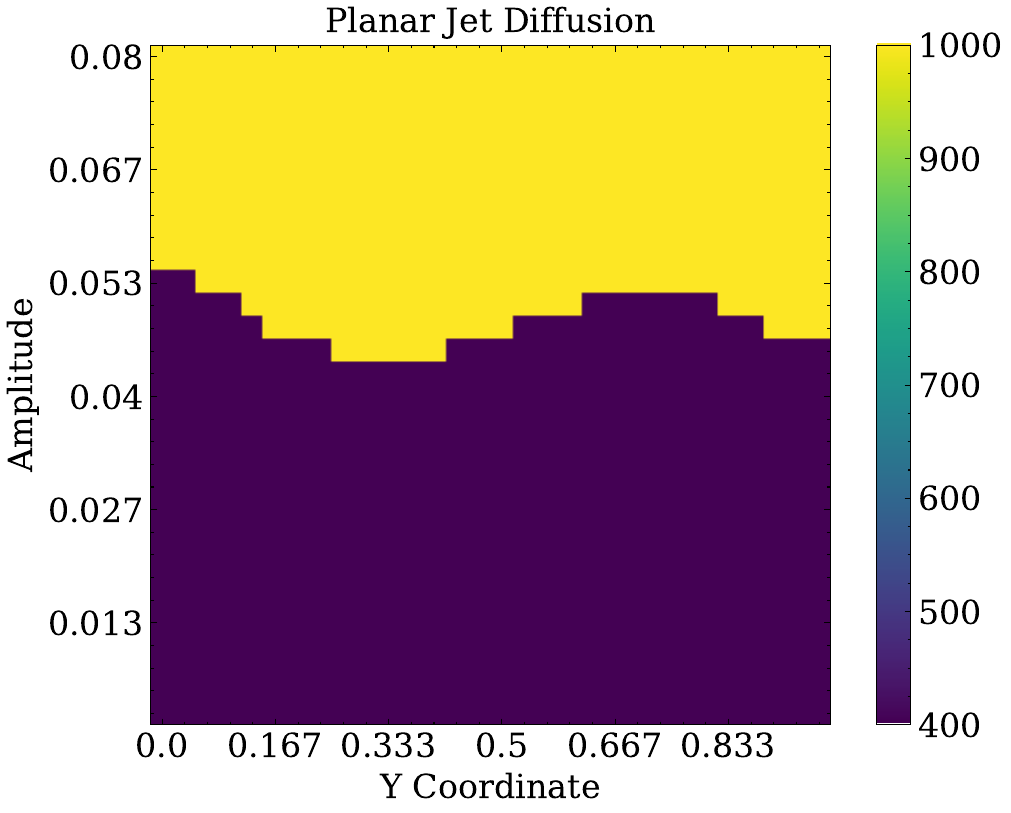}
&
\includegraphics[scale=0.45]{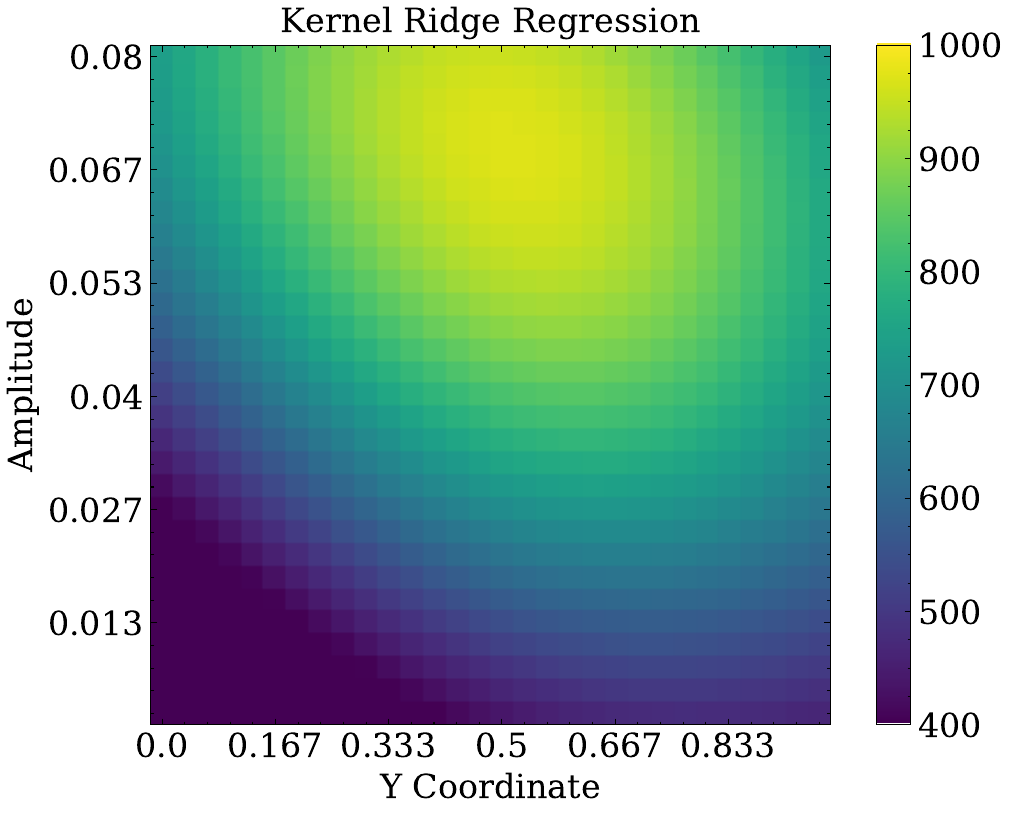}
\\
\includegraphics[scale=0.45]{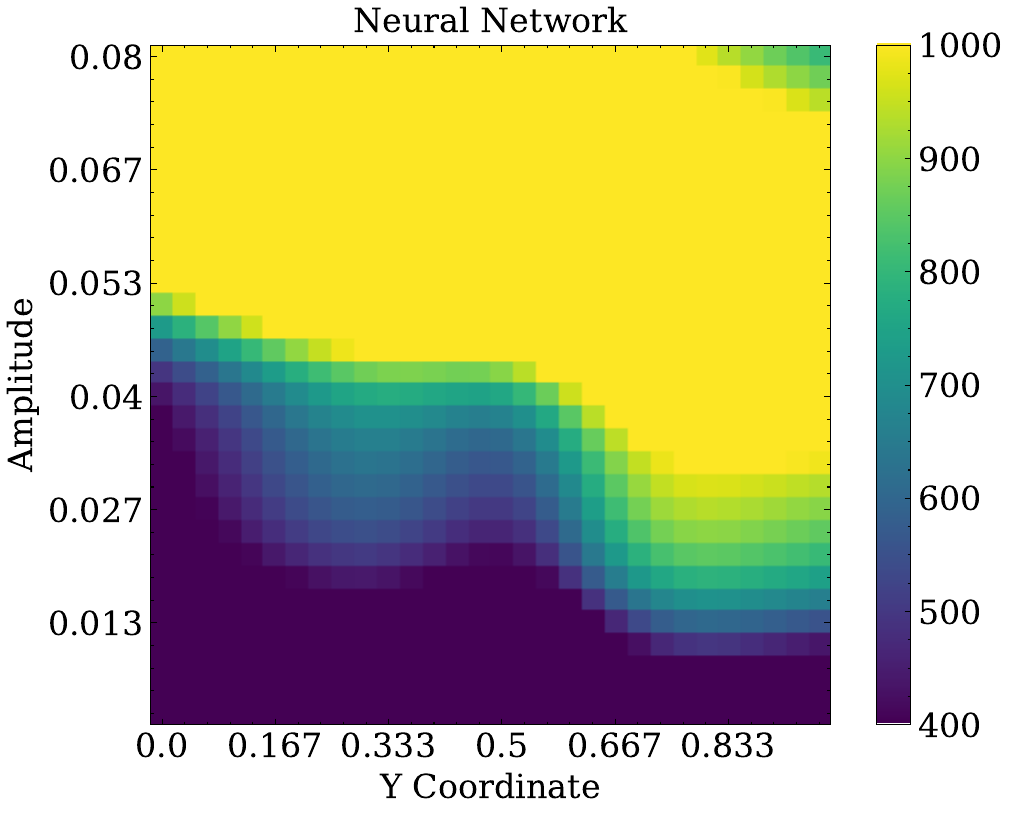}
&
\includegraphics[scale=0.45]{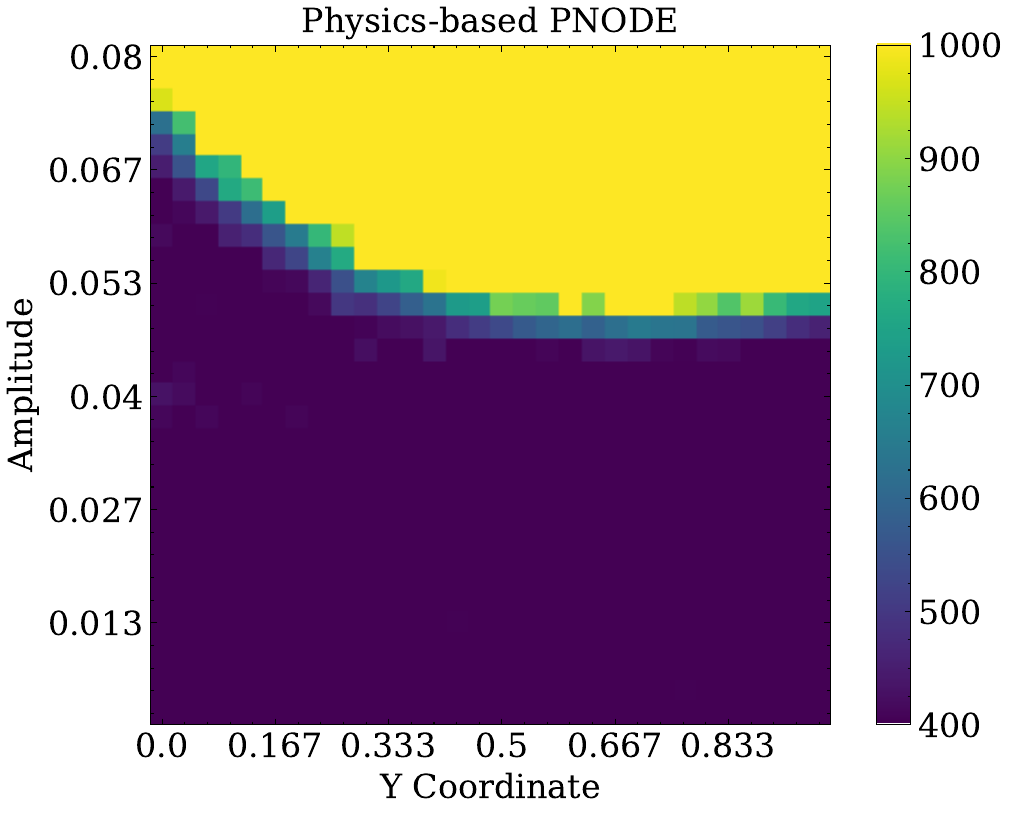}
\end{tabular}
\caption{Prediction of the final temperature by the three models with 100 training samples on simulations with varying amplitude and y coordinate of the laser beam. Top left: ground truth by Planar Jet Diffusion Simulations. Top right: prediction by kernel ridge regression. Bottom left: prediction by neural networks. Bottom right: prediction by physics-based PNODE. With 4,000 data points, all three methods improve their predictions on volume-average final temperature across all data points. We observe that neural networks and kernel ridge regression still predict linear transitions across the boundary of successful ignition, while physics-based PNODE can predict an almost vertical jump from non-ignition to ignition, which is consistent with our high-fidelity simulations.}
\label{fig:y_amplitude}
\end{figure}

\begin{figure}[htbp]
\centering
\begin{tabular}{ll}
\includegraphics[scale=0.45]{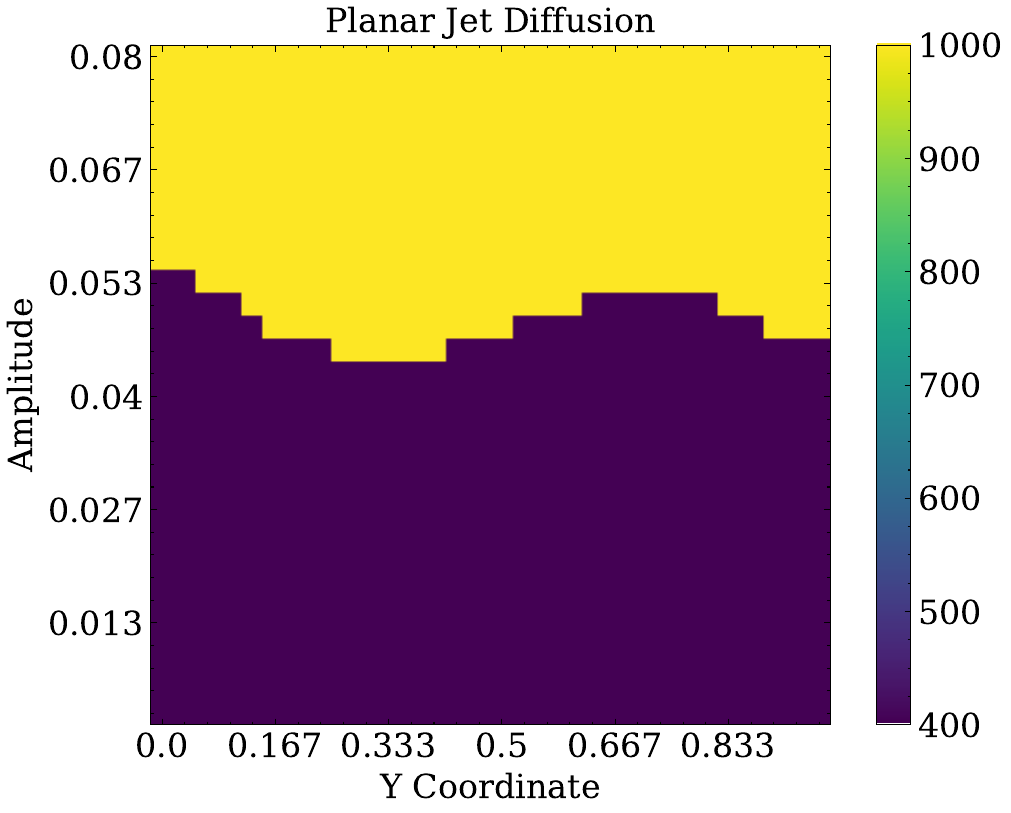}
&
\includegraphics[scale=0.45]{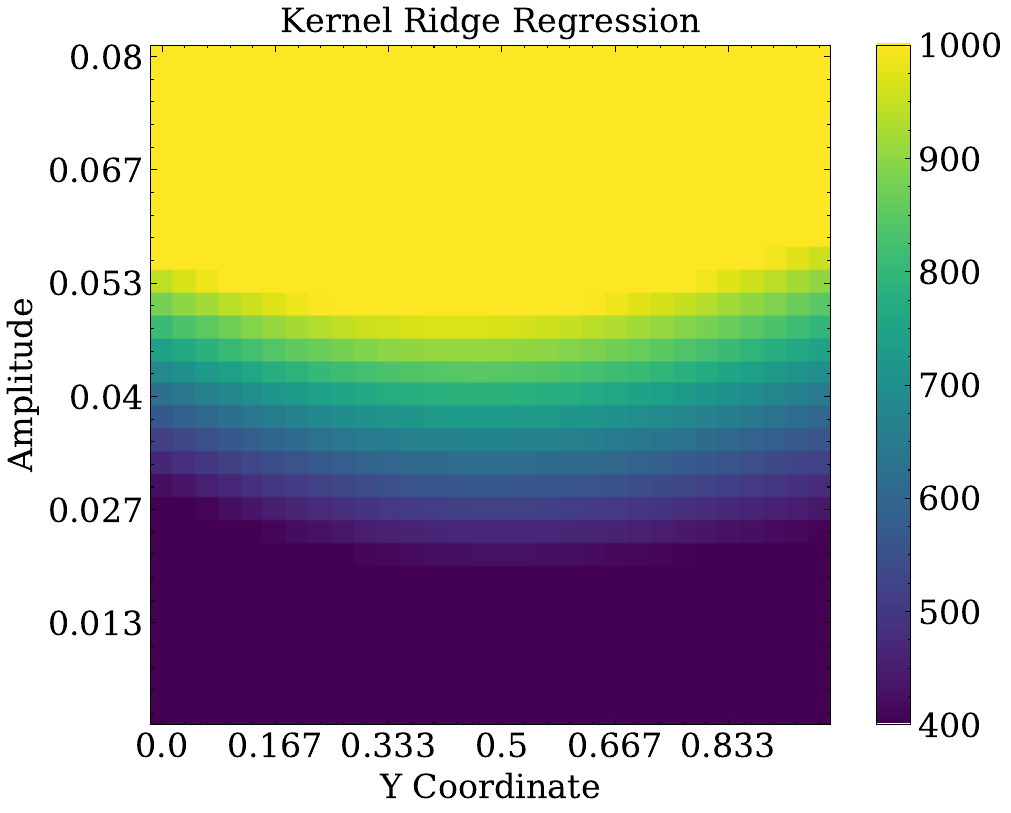}
\\
\includegraphics[scale=0.45]{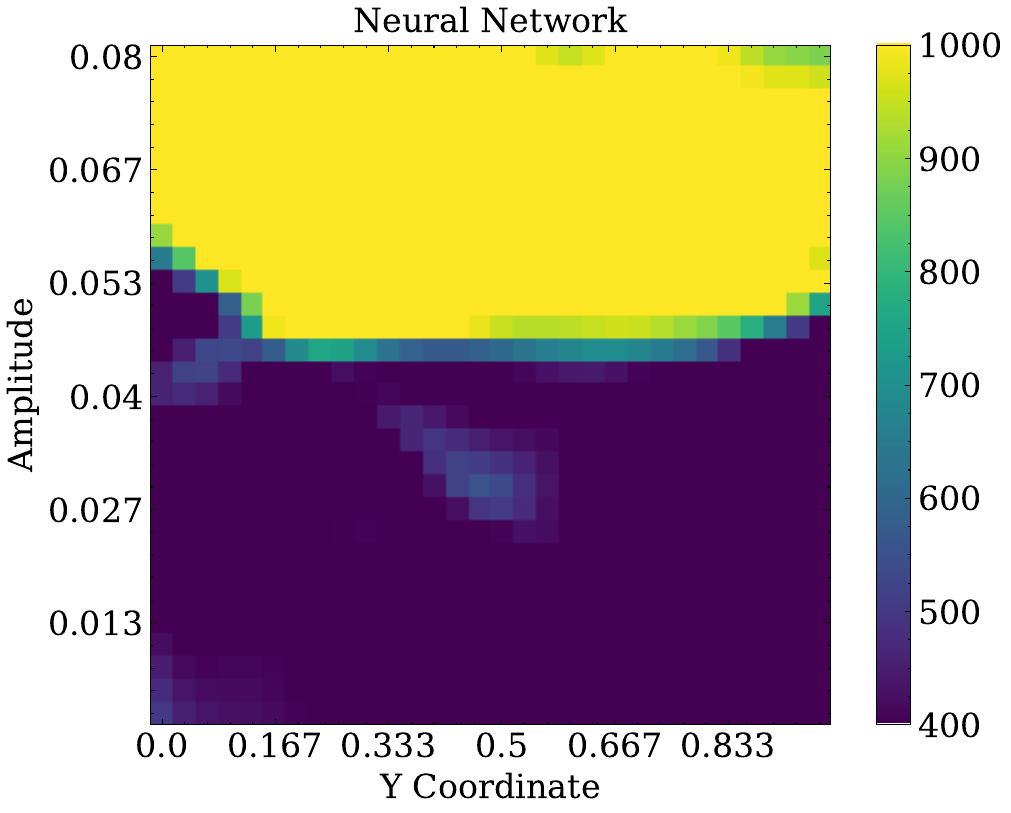}
&
\includegraphics[scale=0.45]{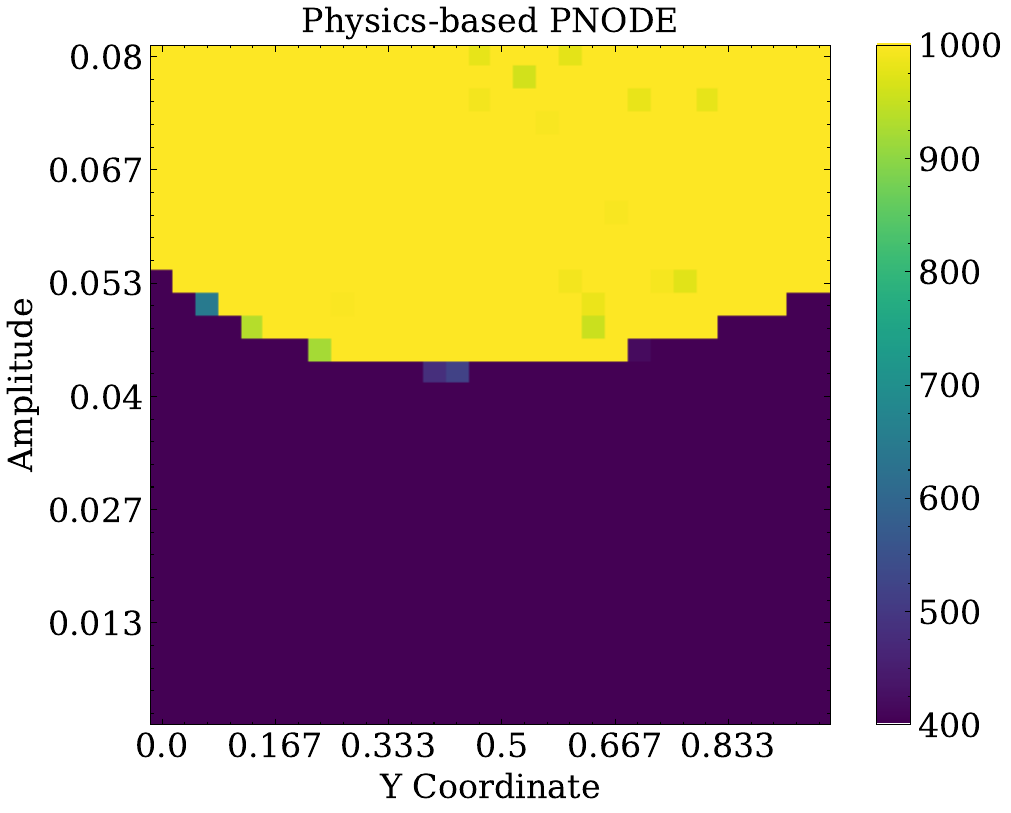}
\end{tabular}
\caption{Prediction of the final temperature by the three models with 4,000 training samples on simulations with varying amplitude and y coordinate of the laser beam. Top left: ground truth by Planar Jet Diffusion Simulations. Top right: prediction by kernel ridge regression. Bottom left: prediction by neural networks. Bottom right: prediction by physics-based PNODE. With 4,000 data points, all three methods improve their predictions on volume-average final temperature across all data points. We observe that neural networks and kernel ridge regression still predict linear transitions across the boundary of successful ignition, while physics-based PNODE can predict an almost vertical jump from non-ignition to ignition, which is consistent with our high-fidelity simulations.}
\label{fig:y_amplitude_4000}
\end{figure}

The prediction of the three models for the cross-section data with 100 training samples is shown in \cref{fig:amplitude_duration,fig:amplitude_radius,fig:y_amplitude}. The prediction with 4,000 training samples is shown in \cref{fig:amplitude_duration_4000,fig:amplitude_radius_4000,fig:y_amplitude_4000}. With only 100 training data points to interpolate in six-dimensional parameter space, both neural networks and kernel ridge regression predict solutions that transition smoothly from non-ignition to ignition in all three cross sections, while our PNODE can capture the sharp transitions near the boundary of the ignition area and also provides an accurate prediction for final temperature for most cases with successful ignition. With the advantage of physics-based PNODE, our 0D flow model can capture the nonlinearity and complexity of the ignition map for planar jet diffusion simulations, despite the limited size of training data. With 4,000 training samples, all three models improve their predictions of the ignition area as shown in three cross sections. However, the neural network and kernel ridge regression still predict linear transitions across the boundary of successful ignition, while physics-based PNODE can predict an almost vertical jump from non-ignition to ignition, which is consistent with our ground-truth data. This further shows that physics-based PNODE, with sufficient training data, is capable of matching the accuracy of high-fidelity simulations in predicting the region of successful ignition, even when learning combustion from a high-dimensional parameter space.

In this study, we aim to distinguish the performance of physics-based PNODE, neural networks, and kernel ridge regression in predicting final temperatures in combustion systems. To achieve this, we applied Density-based Spatial Clustering of Applications with Noise (DBSCAN) on 300 additional test data points and analyzed the error distribution of the three methods \cite{ester1996density}. Our test data was separated into two data sets: ignited and non-ignited. We calculate the distances between data points based on six combustion parameters and divided the points into three categories: core, boundary, and noise points.

\begin{figure}[htbp]
\centering
\begin{tabular}{lll}
\includegraphics[scale=0.35]{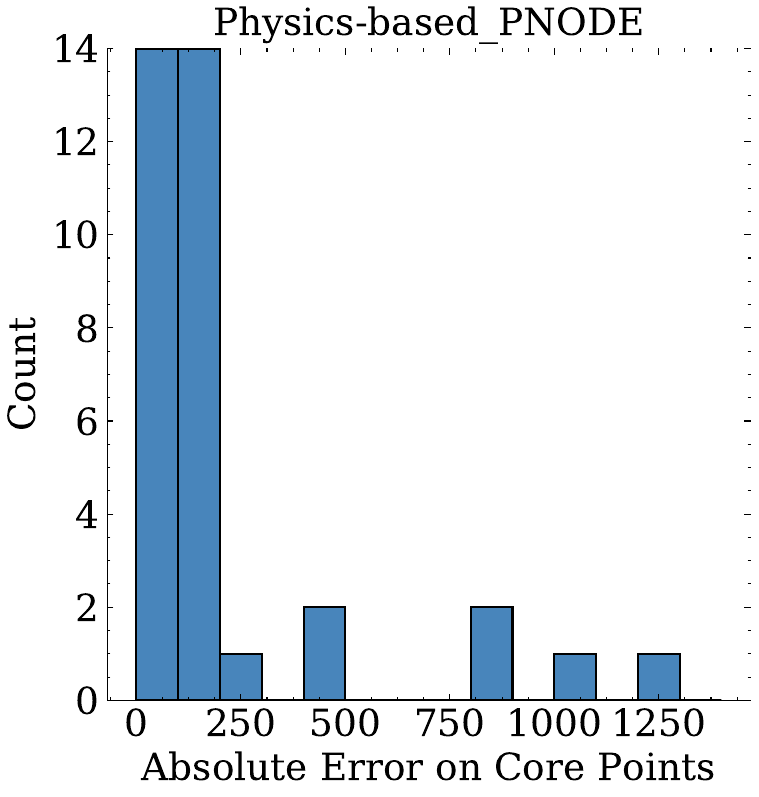}
&
\includegraphics[scale=0.35]{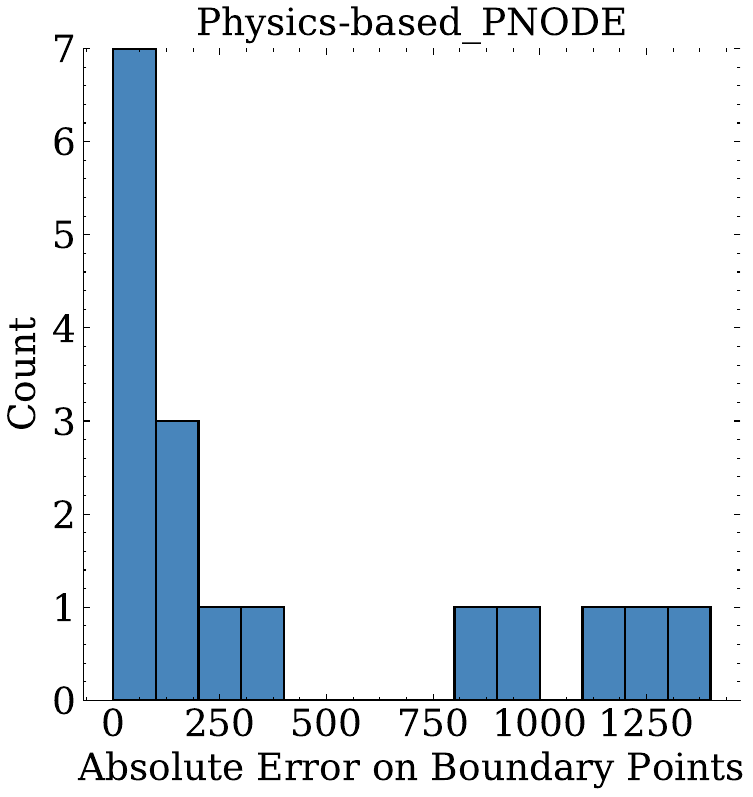}
&
\includegraphics[scale=0.35]{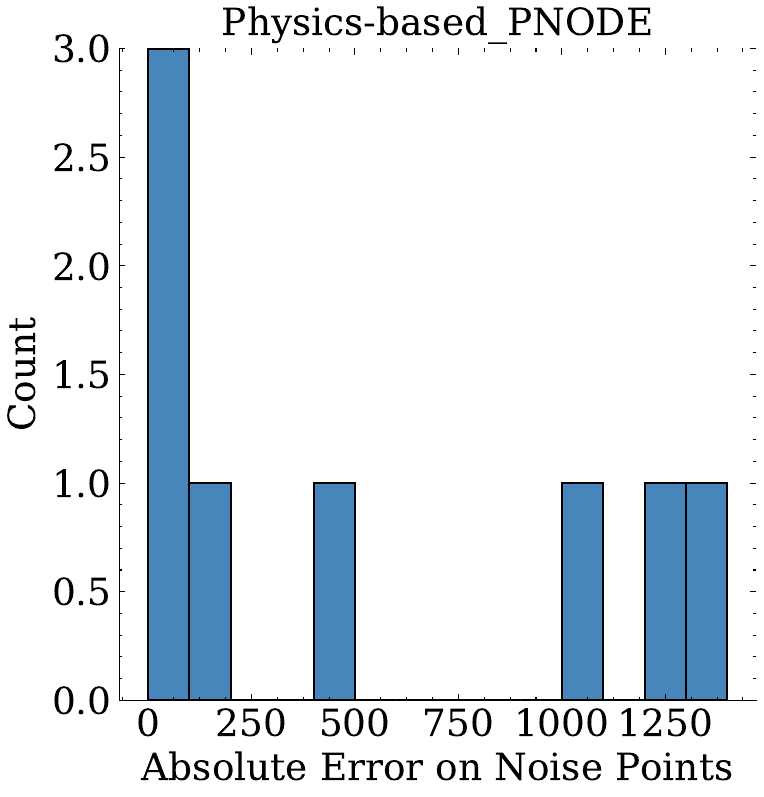}
\\
\includegraphics[scale=0.35]{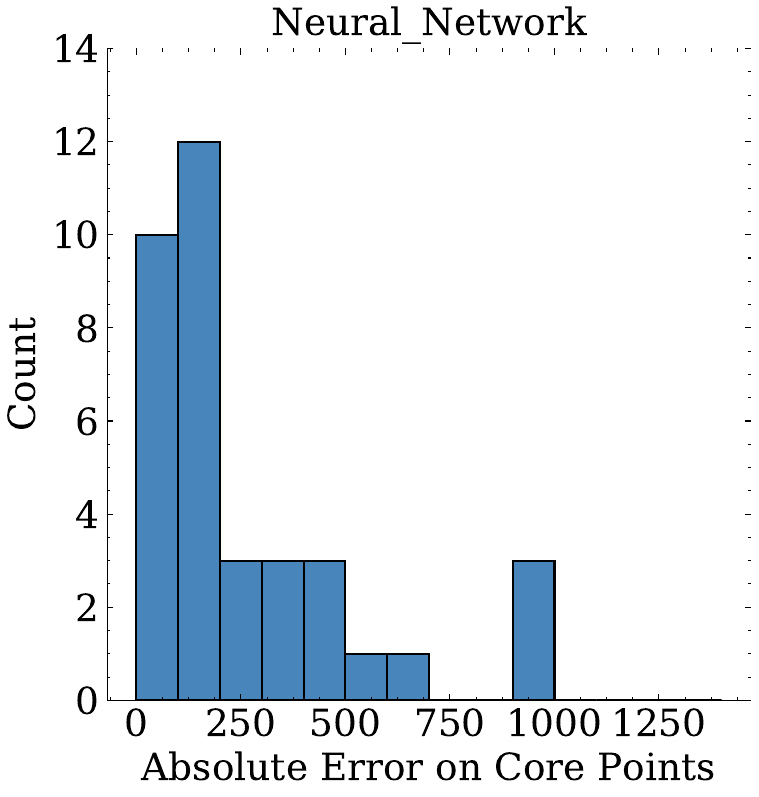}
&
\includegraphics[scale=0.35]{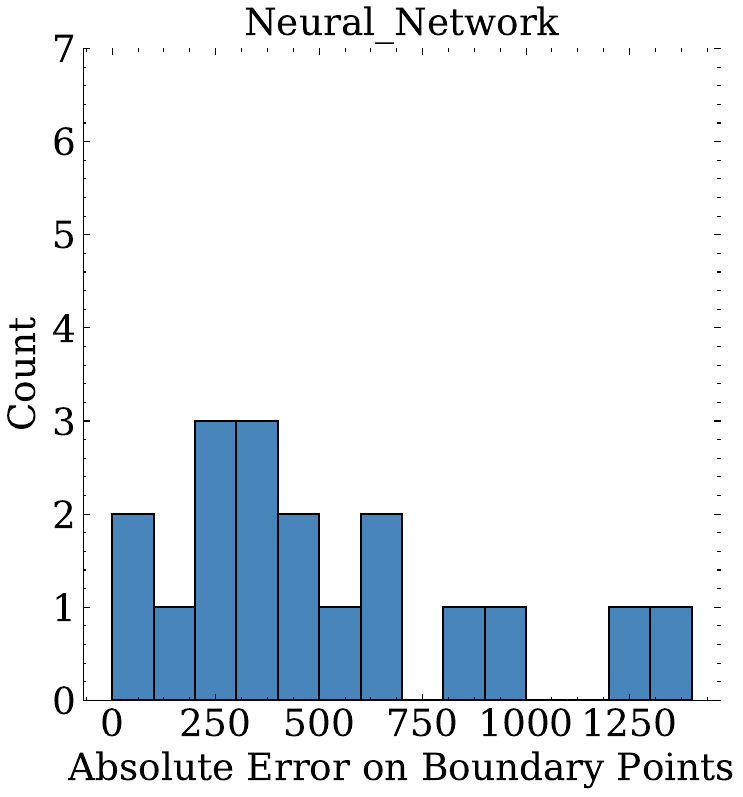}
&
\includegraphics[scale=0.35]{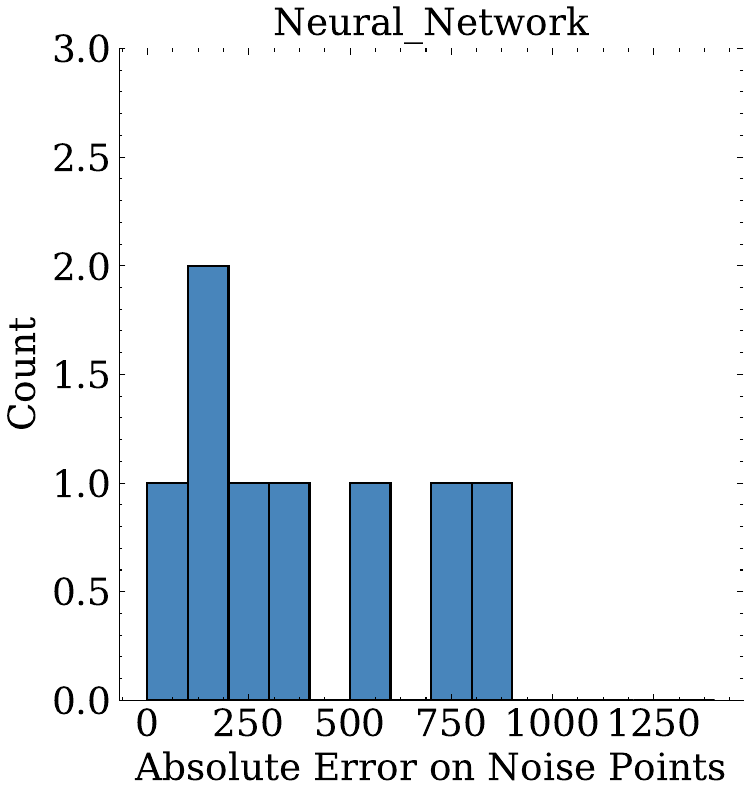}
\\
\includegraphics[scale=0.35]{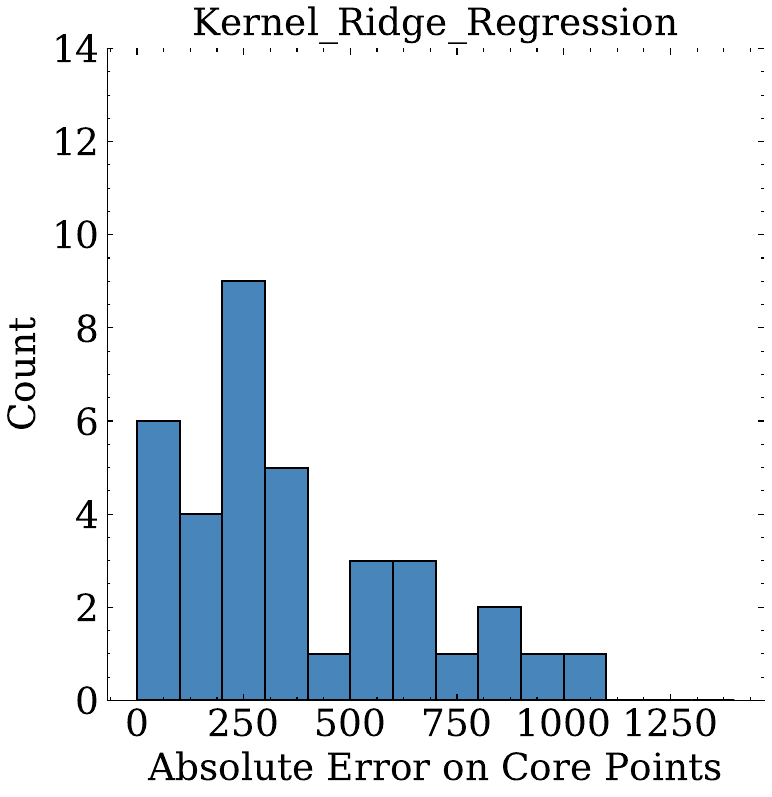}
&
\includegraphics[scale=0.35]{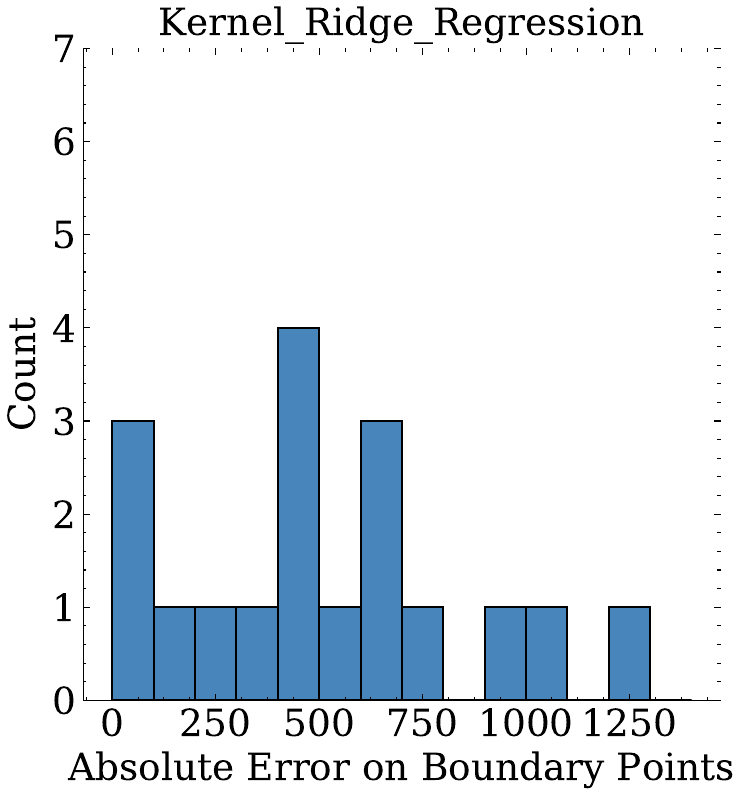}
&
\includegraphics[scale=0.35]{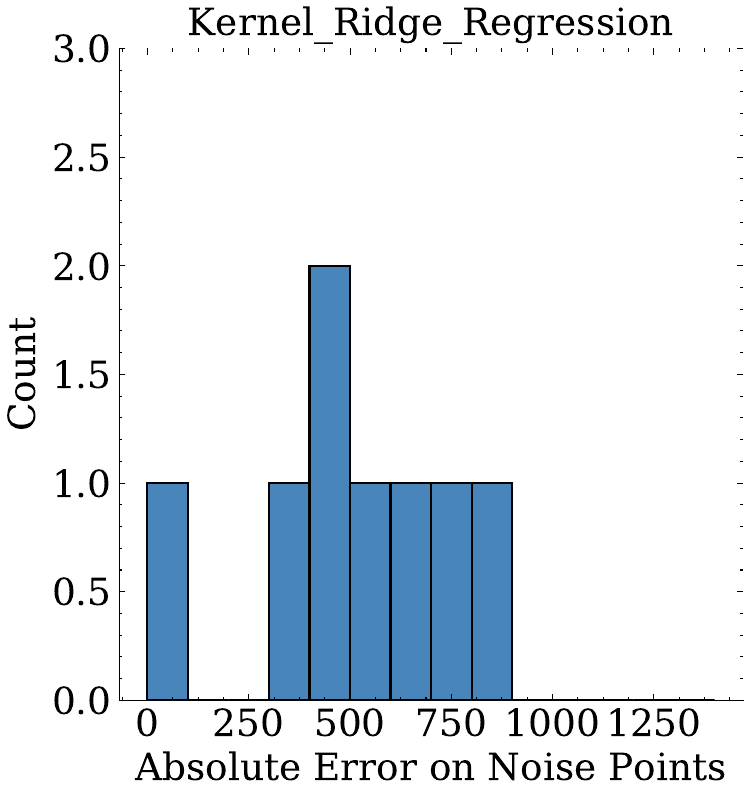}
\end{tabular}
\caption{Error distribution of physics-based PNODE, neural networks, and kernel ridge regression on ignited data sets. Ignited data points are separated into three categories using Density-based Spatial Clustering of Applications with Noise (DBSCAN): core, boundary, and noise points. Core points have many neighboring ignited data points. Boundary points, located near the margin of the ignited cluster, have at least one neighboring ignited data point. Noise points are far away from the core points and have no neighboring ignited data points. Our results indicate that the neural networks and kernel ridge regression provide less accurate predictions than physics-based PNODE, especially on boundary points, as it is more challenging to distinguish ignited cases from non-ignited cases near the boundary of successful ignition.}
\label{fig:DBSCAN_ignited}
\end{figure}

\begin{figure}[htbp]
\centering
\begin{tabular}{lll}
\includegraphics[scale=0.35]{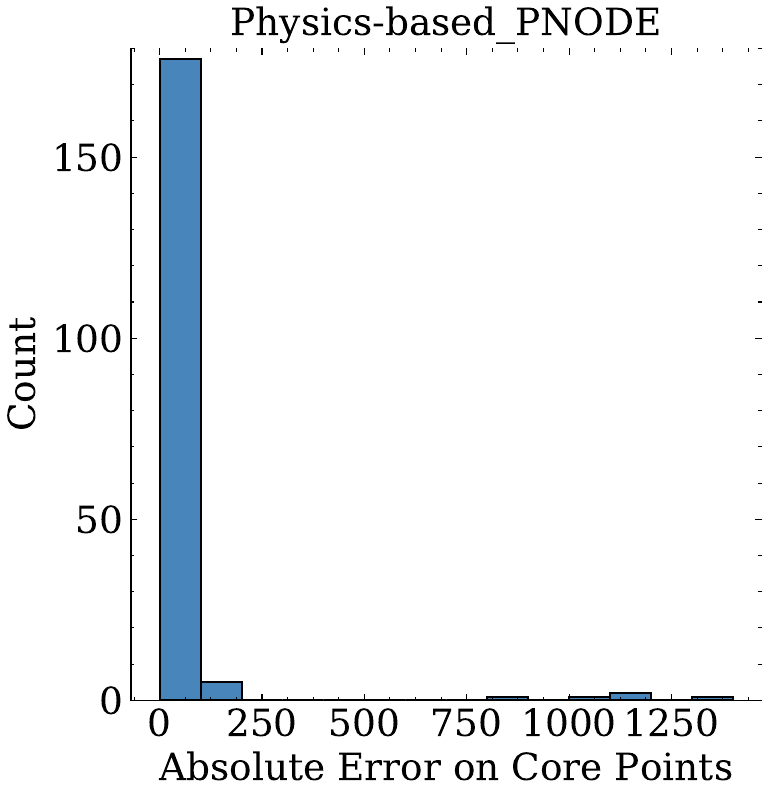}
&
\includegraphics[scale=0.35]{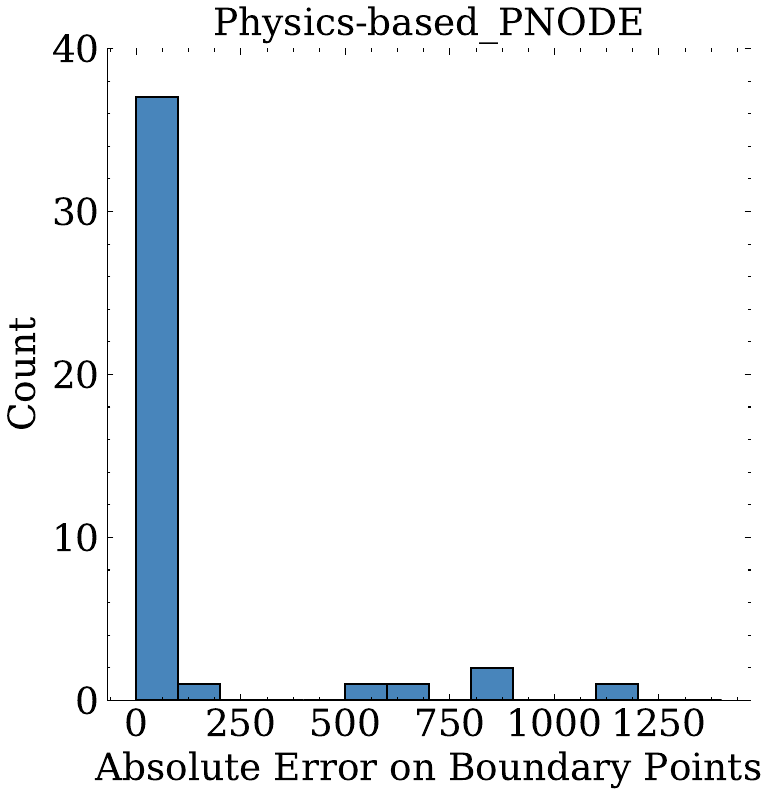}
&
\includegraphics[scale=0.35]{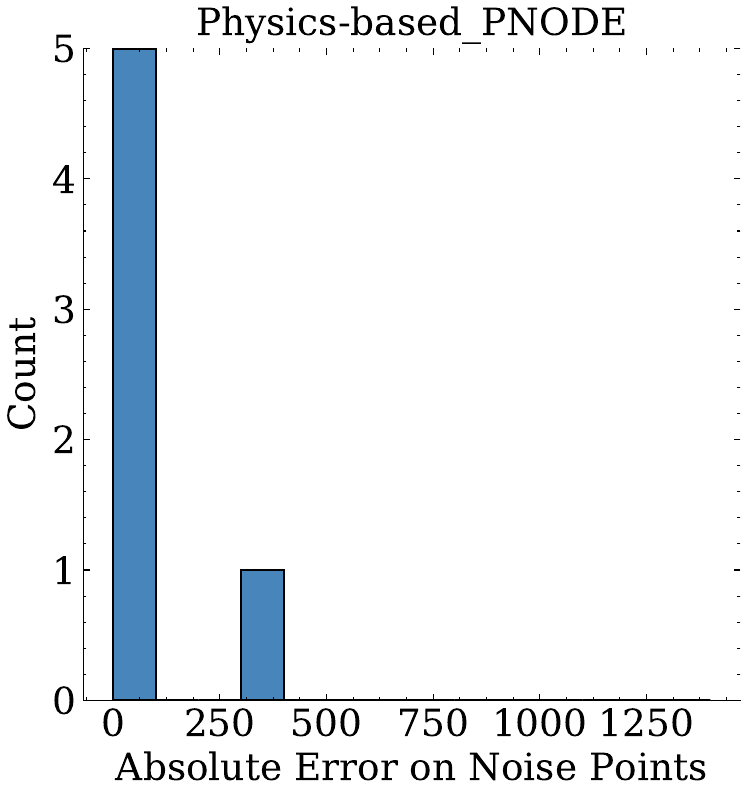}
\\
\includegraphics[scale=0.35]{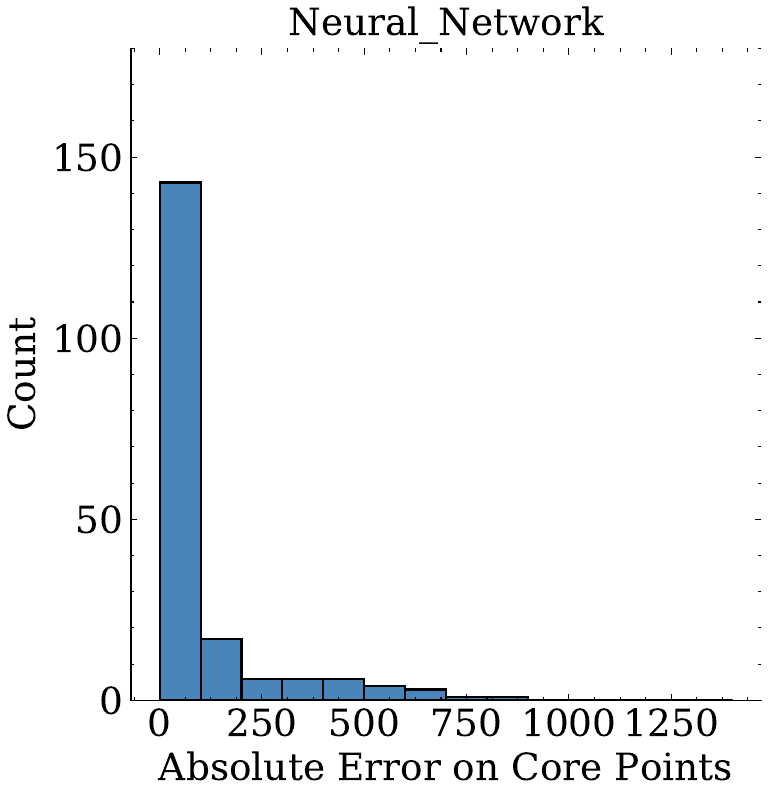}
&
\includegraphics[scale=0.35]{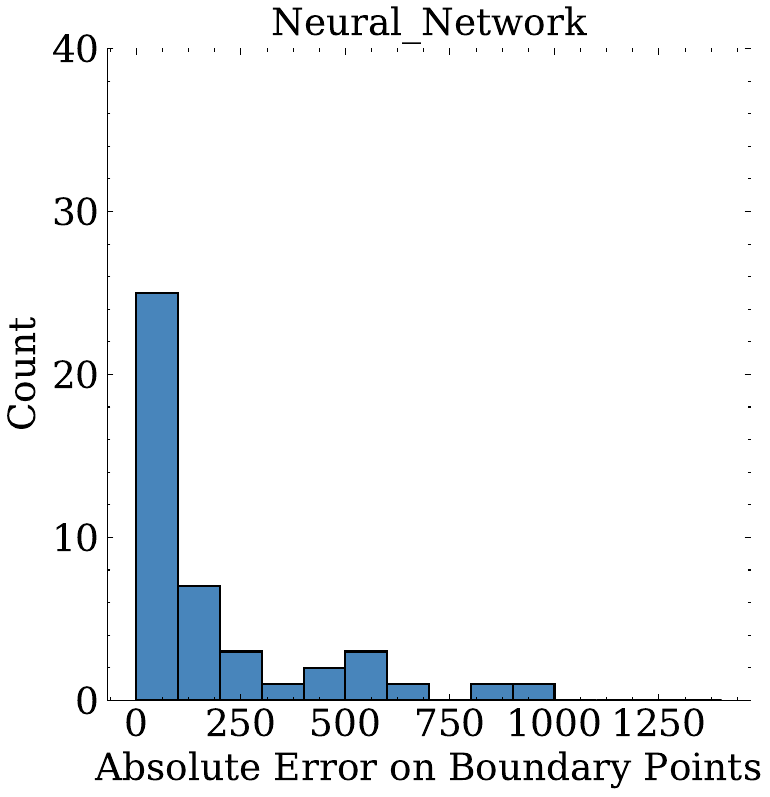}
&
\includegraphics[scale=0.35]{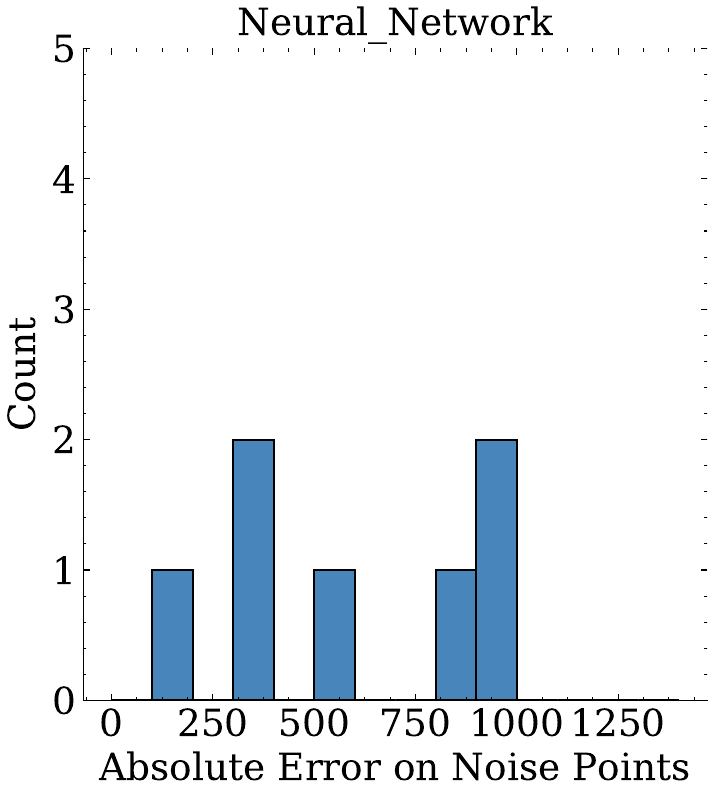}
\\
\includegraphics[scale=0.35]{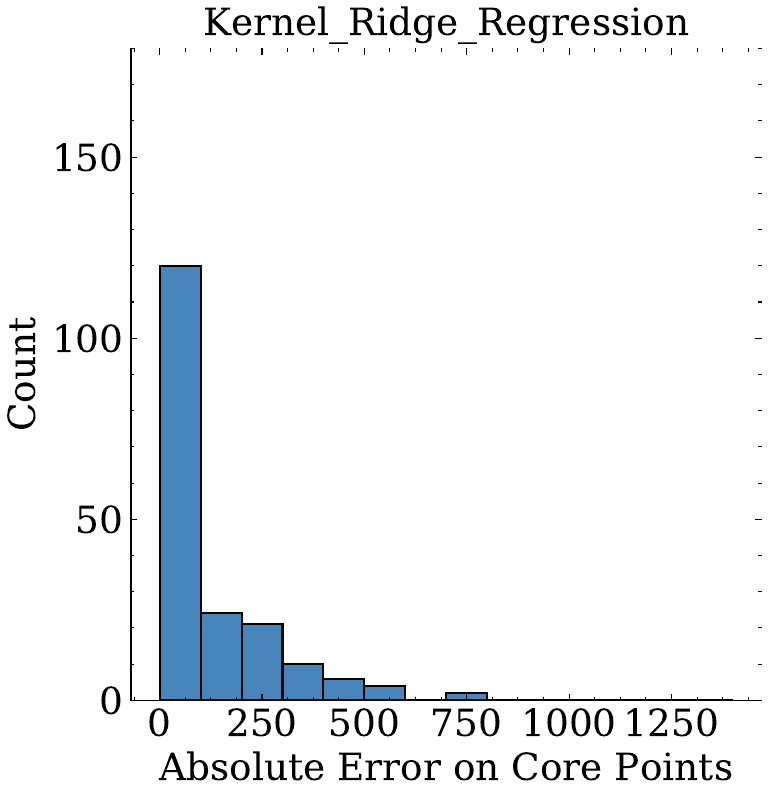}
&
\includegraphics[scale=0.35]{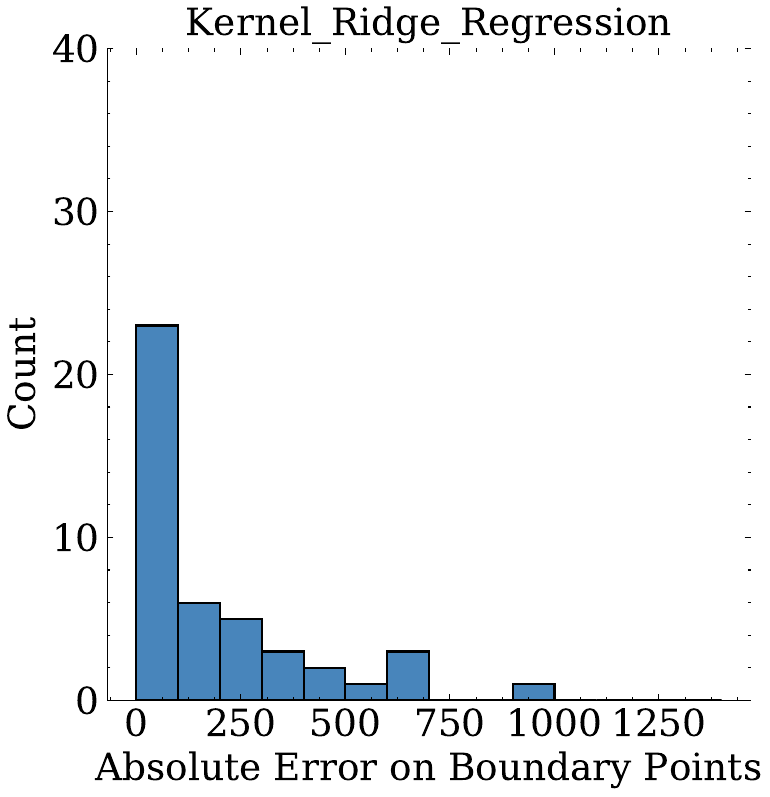}
&
\includegraphics[scale=0.35]{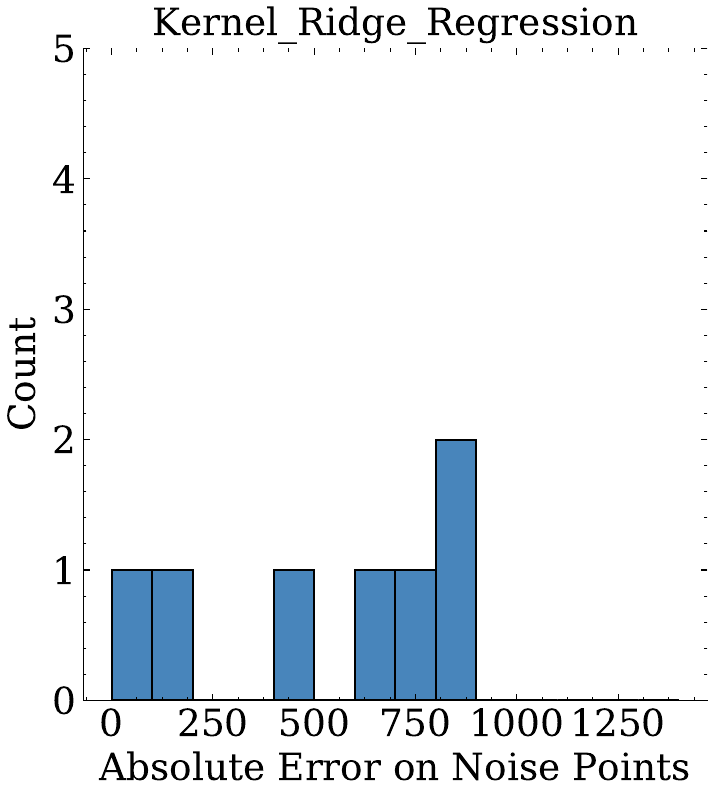}
\end{tabular}
\caption{Error distribution of physics-based PNODE, neural networks, and kernel ridge regression on non-ignited data sets. Non-ignited data points are separated into three categories using Density-based Spatial Clustering of Applications with Noise (DBSCAN): core, boundary, and noise points. Core points have many neighboring non-ignited data points. Boundary points, located near the margin of the non-ignited cluster, have at least one neighboring non-ignited data point. Noise points are far away from the core points and have no neighboring non-ignited data points.  Our results indicate that neural networks and kernel ridge regression provide less accurate predictions than physics-based PNODE, especially on boundary points, as it is more challenging to distinguish ignited cases from non-ignited cases near the boundary of successful ignition.}
\label{fig:DBSCAN_nonignited}
\end{figure}

Core points have many close neighbors in the same data set, making them easier to predict than other points. Boundary points, on the other hand, are located at the margin of two clusters, most of which are often close to both the ignition and non-ignition data sets. Lastly, we have noise data points that are far away from both data sets. The error distribution of the three models on different types of data points for the ignited case and non-ignited case are shown in \cref{fig:DBSCAN_ignited,fig:DBSCAN_nonignited}. Our results indicate that physics-based PNODE provides consistently accurate predictions of final temperature for both core and boundary points, with most absolute errors less than \SI{200}{K}.

On the contrary, the performance of neural networks and kernel ridge regression was significantly worse on boundary points, as it is more challenging to distinguish ignited cases from non-ignited cases near the boundary. All three methods have low accuracy for noise points, as they are difficult to predict due to a lack of nearby data points. Our analysis showed that a large number of errors from neural networks and kernel ridge regression fall within the 400 to 1000 range, indicating a smooth transition from ignition to non-ignition regions in terms of final temperature. This is inconsistent with the actual ignition process, as final temperatures from high-fidelity simulations or experiments typically display a binary nature with sharp transitions from non-ignition to ignition regions.

\section{Conclusion}

In this work, we proposed a novel hybrid model that combines a 0D reacting flow model and deep neural networks based on physics-based PNODE. By embedding deep neural networks into the 0D model as the heat source function and Arrhenius reaction parameters, we have developed a reduced-order surrogate model that enables high-fidelity Computational Fluid Dynamics simulations of combustion processes. Furthermore, our PNODE-based model is capable of providing, with a limited number of training samples, physically constrained solutions that describe a highly complex parameter space for combustion systems. We validated our approach with high-fidelity Planar Jet Diffusion simulations using the HTR solver with six varying combustion parameters. The performance of our PNODE-based 0D model is compared with that of two widely used approaches, namely kernel ridge regression and classical neural networks. Our results demonstrate that our PNODE-based 0D model can predict sharp transitions near the boundary of successful ignition, as well as the evolving chemistry of the combustion system, even with a limited number of data points.

\printbibliography

\end{document}